\documentclass{article}

\usepackage{microtype}
\usepackage{graphicx}
\usepackage{subcaption}
\usepackage{booktabs} % for professional tables
\usepackage[ruled,vlined]{algorithm2e} % for \While, \ForEach, etc.
\usepackage{hyperref}
\usepackage[preprint]{icml2026}

\usepackage{amsmath}
\usepackage{amssymb}
\usepackage{mathtools}
\usepackage{amsthm}

\usepackage[capitalize,noabbrev]{cleveref}

\theoremstyle{plain}
\newtheorem{theorem}{Theorem}[section]

\theoremstyle{definition}

\theoremstyle{remark}

\newcommand{\name}[0]{\textit{AugServe}}
\newcommand*\circled[1]{\tikz[baseline=(char.base)]{ \node[shape=circle,draw,inner sep=0.4pt,color=black,fill=white,minimum size=0.3cm] (char) {#1};}}
\usepackage{tabularx}
\usepackage{makecell}
\usepackage{booktabs}
\usepackage{ragged2e}   % 让 X 列左对齐更好看
\usepackage{threeparttable} % 可选：加脚注

\usepackage{booktabs,tabularx,threeparttable,makecell,ragged2e}
\usepackage{tikz}
\usepackage{amsmath}
\usepackage{subcaption} % 用于子图环境
\usepackage{graphicx}
\usepackage{caption}
\usepackage{array}
\usepackage{multirow}
\usepackage{hyperref}
\usepackage{enumitem}

\usepackage{calc} % 用于 \heightof 命令
\icmltitlerunning{\textit{AugServe}: Adaptive Request Scheduling for Augmented Large Language Model Inference Serving}
\begin{document}

\twocolumn[
  \icmltitle{\textit{AugServe}: Adaptive Request Scheduling  for \\ Augmented Large Language Model Inference Serving}



  \icmlsetsymbol{equal}{*}

  \begin{icmlauthorlist}
    \icmlauthor{Ying Wang}{sch}
    \icmlauthor{Zhen Jin}{sch}
    \icmlauthor{Zhenqian Chen}{sch}
    \icmlauthor{Jiexiong Xu}{sch}
    \icmlauthor{Wenhai Lin}{comp}
    \icmlauthor{Yiquan Chen}{comp}
    \icmlauthor{Wenzhi Chen}{sch}
  \end{icmlauthorlist}

  \icmlaffiliation{sch}{College of Computer Science and Technology, Zhejiang University, Hangzhou, China}
  \icmlaffiliation{comp}{Alibaba Group, Hangzhou, China}

  \icmlcorrespondingauthor{Jiexiong Xu}{jasonxu@zju.edu.cn}
  \icmlcorrespondingauthor{Wenzhi Chen}{chenwz@zju.edu.cn}

  \icmlkeywords{Machine Learning, ICML}

  \vskip 0.3in
]



\printAffiliationsAndNotice{}  % no special notice (required even if empty)

\begin{abstract}
  % Augmented large language models (LLMs) that invoke external calls are increasingly prevalent in inference serving, introducing execution pauses and request variability that challenge inference efficiency under Service-Level Objectives (SLOs).
% Existing systems rely on request scheduling that is unaware of external-call-induced execution dynamics and typically employ static or coarse-grained batch-level token budgets, resulting in severe Head-of-Line (HoL) blocking and degraded effective throughput.
% We present \name{}, an efficient augmented LLMs inference serving framework that reduces queueing latency and substantially improves effective throughput under external-call-augmented workloads.
% \name{} combines state-aware request scheduling with the dynamic batch-level token budget to adapt to heterogeneous requests and their dynamic execution states induced by external calls.
% Experimental results show that \name{} achieves 6.5$\times$ and 4.7$\times$ higher effective throughput than vLLM and INFERCEPT, respectively.

Augmented large language models (LLMs) that invoke external calls are increasingly prevalent in inference serving.
However, such augmentations pose significant challenges to inference efficiency under strict Service-Level Objectives (SLOs). Existing inference systems are agnostic to the dynamic execution behaviors induced by external calls and rely on fixed batch-level token budget, which leads to severe Head-of-Line (HoL) blocking and substantially reduced effective throughput.
We present \name{}, an efficient augmented LLM inference serving framework
that mitigates request queuing latency and improves effective throughput under
external-call-augmented workloads. \name{} integrates state-aware request scheduling with dynamic batch-level token budgets to adapt to heterogeneous requests and their dynamically changing execution states. Experimental results show that \name{} achieves 6.5$\times$ and 4.7$\times$ higher effective throughput than vLLM and INFERCEPT, respectively.
  
\end{abstract}
% \vspace{-10pt}
\section{Introduction}
Augmented Large Language Models (LLMs) have rapidly emerged as a promising paradigm \cite{infercept,toolkengpt} for modern LLM inference serving.
% In contrast to
Compared with traditional text-only LLMs, which rely on fixed pretrained parameters and lack real-time knowledge \cite{toolformer,abouttime}, augmented LLMs extend their capabilities by invoking external tools (e.g., web APIs, databases, or specialized models) during inference \cite{ nips24advancing,nature2025study,toollearning, acl24gear}. This approach enables augmented LLMs to perform more complex tasks such as arithmetic computation \cite{acl24good,toolbox}, real-time information retrieval \cite{dragin,abouttime}, and web interactions \cite{webrl,webpiot}.
% (e.g., web APIs, databases, or specialized models) during inference \cite{ nips24advancing,nature2025study,toollearning, acl24gear, chatcot}
% , gradually positioning them as a core component of web services.

\begin{figure}[t]
    \centering 
    \includegraphics[width=1.0\linewidth]{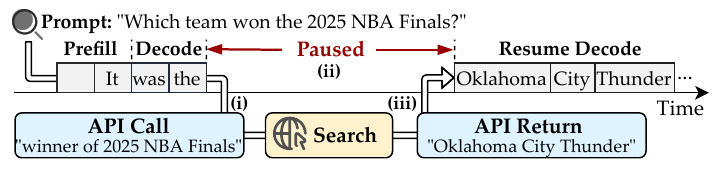}
    % \vspace{-20pt}
    \caption{Augmented LLM inference process.} 
    % \vspace{-18pt}
    \label{fig:aug_llm_infer}
\end{figure}

Augmented LLM inference service systems are becoming the key infrastructure for AI-centric cloud computing, with inference efficiency directly impacting user experience. \autoref{fig:aug_llm_infer} illustrates the workflow of the augmented LLM inference service \cite{infercept, asynclm}:
\textbf{(i)} During inference, the augmented LLM identifies the need for real-time information and triggers the corresponding tool calls. 
\textbf{(ii)} The inference process is paused while awaiting the response from the external augmentation module. 
\textbf{(iii)} Upon the response being returned, the serving system appends it to the sequence generated and resumes normal generation.
Ideally, inference systems must simultaneously deliver high throughput and low latency.
In this context, \textbf{Service-Level Objectives (SLOs)} serve as strict latency boundaries (e.g., requiring Time-to-First-Token (TTFT) below a fixed threshold) ~\cite{aptserve,distserve,fastserve,splitwise}. Accordingly, the system's efficiency is best characterized by \textbf{effective throughput} (or \textbf{goodput}), defined as the volume of requests processed per unit time that successfully satisfy these SLO requirements ~\cite{revisiting_slo, optimizing_goodput, nsdi23_shepherd_goodput}.

State-of-the-art inference systems focus on improving inference performance.
vLLM \cite{vllm} has emerged as the de facto standard for efficient LLM serving.
However, in augmented LLM inference, vLLM treats external calls as request termination and discards the request's context (i.e., Key-Value (KV) cache) ~\cite{infercept}. 
When the call returns, the system must recompute the KV cache, incurring substantial computation overhead and processing latency.
To address this issue, INFERCEPT~\cite{infercept} dynamically manages context based on call duration and context length, selecting among three KV-cache handling policies: discarding it, preserving it in GPU memory, or swapping it to the host memory. This design reduces resource waste and significantly improves the efficiency of augmented LLM inference.

% 然而现有的这些推理框架在提升增强LLM推理的有效吞吐量目标下仍然面临两个严重的挑战：(1) 排队延迟过高导致更多请求严重超过SLOs . 现有的推理框架如vLLM和InferCept通常都采用FCFS调度，容易导致长请求阻塞短请求，造成head-of-line阻塞，从而引发高延迟和吞吐损失。如图2a所示，在负载4.0下在H800GPU上用InferCept部署OPT-13B，百分之五十的请求排队时间超过343s，严重超过SLOs 。
% 同时图b显示了在负载超过2.0req/s后，InferCept每秒完成的满足SLO的请求数量下降严重(这里我们设置SLO为TTFT<1,per-token latency <10×一轮迭代时间)。这种严格按照到达时间进行排序的调度方式，严重阻碍了推理系统在高负载下的有效吞吐量提升。
% （2）现有系统采用静态token 批处理上限，无法适应动态负载与硬件条件，限制吞吐量的提升。
% 设置过小导致每轮处理请求很少吞吐量很低，设置过大引发资源争用和过度重计算，这些限制使得现有框架难以在满足 SLOs 的同时进一步提升吞吐量。
% 因此需要一个有效的调度策略以及适应负载变化的动态token批处理方案来打破这些限制，有效实现性能目标

% 然而，现有推理框架在提升增强 LLM 推理的有效吞吐量方面仍面临两大挑战：
% (1) 排队延迟过高,break the SLOs。在现有服务系统中，更多的请求排队延迟和整体完成时间严重超过SLOs ，将将导致有效吞吐量显著降低。例如 vLLM 和 INFERCEPT 通常采用 FCFS 调度，长请求会阻塞短请求，导致严重的 head-of-line (HoL) 阻塞，从而带来高延迟和吞吐下降。增强LLM推理中外部调用期间长上下文管理和返回长度的不确定性会进一步加剧HoL问题，导致排队延迟更严重。如图 2展示了在 H800 GPU 上运行InferCept with OPT-13B服务混合增强LLM工作负载\cite{infercept}，请求的排队延迟分布和有效吞吐量随负载变化情况。如图所示，负载为4.0req/s时有 50% 的请求排队时间超过 343s，远超 SLOs 要求(例如一般SLO要求TTFT<1s)；图 2b 进一步显示，在负载超过 2.0 req/s 后，InferCept 每秒完成的满足 SLOs 的请求数急剧下降（此处 SLOs 设为 TTFT < 1，per-token latency < 10× 单轮迭代时间）。这种严格按照到达时间进行排序的调度方式，严重阻碍了增强LLM推理服务在高负载下的有效吞吐量提升。
% (2) 静态 token 批处理上限。在增强 LLM 中，请求的输入、输出和 API 返回长度差异巨大。固定的 token 上限无法适应动态负载和硬件条件：设置过小会导致每轮处理请求过少、吞吐偏低；设置过大会引发资源争用和过度重计算。
% 综上，现有框架难以在满足 SLOs 的同时进一步提升吞吐量，因此需要一种能够结合 增强 LLM 特性 的调度策略和动态 token 批处理方案。
% However, existing inference systems still face two challenges in improving the effective throughput of augmented LLM inference:
However, these systems still face two challenges in improving the goodput of augmented LLM inference:
\textbf{C1: Inadequate scheduling leads to Head-of-Line (HoL) blocking and SLO violations.}
In augmented LLM inference, existing systems (e.g., vLLM and INFERCEPT) typically adopt First-Come-First-Served (FCFS) scheduling,
batching requests without accounting for external calls. 
When long requests trigger external calls and pause execution, their context (KV cache) may be preserved in GPU memory, swapped out to host memory, or discarded, all of which can block subsequent short requests. 
This results in severe HoL blocking, causing queuing delays that exceed SLOs and sharply degrade goodput.
Some work attempts to mitigate these delays using approximate Shortest-Job-First (SJF) scheduling based on request length ~\cite{s3,nips24_rank,fastserve}.
However, they still ignore the execution heterogeneity introduced by external calls and remain suboptimal for augmented LLM inference.
\textbf{C2: Fixed batch-level token budget restricts throughput under external calls.} 
In augmented LLM inference, external calls introduce paused requests whose contexts may occupy GPU memory, complicating batch capacity selection.
A static batch-level token budget cannot adapt to this dynamic memory availability.
A small budget limits per-iteration concurrency and reduces throughput,
while an overly large budget induces resource contention and frequent eviction of paused contexts, incurring recomputation overhead.
% If the budget is too small, only a few requests can be processed per iteration, reducing throughput. If the budget is too large, it induces resource contention and triggers frequent eviction of paused request contexts, incurring additional recomputation overhead.
Moreover, existing approaches \cite{batchllm} adjust the budget based only on free GPU memory, ignoring reclaimable memory under different context-handling policies, leading to suboptimal budget decisions.
In this paper, we propose \name{}, an augmented LLM inference serving framework
that jointly rethinks request ordering and batch capacity.
Our key insight is that augmented LLM inference introduces significant execution heterogeneity, where requests with different external calls and context-handling policies exhibit distinct resource demands across multiple stages.
% Our key insight is that, in augmented LLM inference, external calls and context-handling decisions jointly introduce heterogeneous execution behavior across a request’s lifecycle, spanning multiple execution rounds and stages with distinct resource demands.
% with dynamically varying resource demands and execution efficiency.
% Our key insight is that augmented LLM inference exhibits highly heterogeneous execution behavior across a request’s lifecycle, spanning multiple rounds and stages, where external calls and context-handling decisions jointly shape each stage’s latency and resource characteristics.
% Our key insight is that augmented LLM inference exhibits highly heterogeneous execution behavior across a request’s lifecycle, spanning multiple rounds and stages. External calls and context-handling decisions jointly shape each stage’s latency and resource characteristics, leading to dynamic and state-dependent impacts on queueing delay and effective throughput.
% Our key insight is that augmented LLM inference exhibits highly heterogeneous and state-dependent execution behavior across a request’s lifecycle, where latency and throughput are jointly shaped by external call behaviors and context-handling decisions.
% Our key insight is that, augmented inference exhibits highly heterogeneous execution behavior across a request’s inference lifecycle, where latency and throughput are jointly shaped by external call behaviors and context-handling decisions. 
Guided by this insight, \name{} adopts a unified, state-aware design that optimizes request scheduling and batch-level capacity adaptation to maximize goodput under dynamic augmented LLM workloads.
% Guided by this insight, \name{} introduces an \textbf{adaptively state-aware scheduling strategy (C1)} that incorporated external call features and runtime feedback to optimize request ordering,
% and a \textbf{dynamic batch-level token budget mechanism (C2)} that adjusts system capacity in response to paused requests and memory pressure. Together, these designs achieves superior effective throughput under dynamic workloads.

For \textbf{state-aware scheduling (C1)}, \name{} models request scheduling across the full inference lifecycle,  explicitly capturing multi-stage execution and cross-round state evolution induced by external calls.
% explicitly accounting for external-call behavior and context-handling outcomes.
Scheduling priorities are constructed in a state-aware manner that adapts to each request’s execution state, context-handling outcome, and observed runtime feedback.
% By prioritizing requests with higher execution efficiency in lower expected resource cost,
By prioritizing requests with higher execution efficiency given their current execution stages and resource footprint,
% By favoring requests with higher space-time efficiency, 
\name{} alleviates HoL blocking in augmented LLM inference, significantly reducing queuing delays and improving goodput.
% The scheduler favors requests with higher space-time efficiency, alleviating head-of-line blocking under external calls, significantly reducing cumulative queuing delays and enhancing throughput.

% Specifically, a lightweight prediction model estimates call duration and output length at arrival, while priorities are dynamically refined using runtime feedback such as realized return lengths and context-handling outcomes.
% As a result, requests with high service efficiency are batch-processed first, significantly reducing cumulative queuing delays and enhancing throughput.
% By being aware of request execution state and dynamically updating priorities using runtime feedback, \name{} effectively alleviates HoL blocking under heterogeneous workloads.

% For dynamic token-level batching，我们提出结合发生外部调用的请求情况、硬件内存信息、实时负载情况和容量使用情况，动态设置批处理token上限，保持在不同负载下的高吞吐。同时我们设置区间约束来防止极端情况的发生，提升推理效率的同时保证系统的稳定性
For \textbf{dynamic batch-level token budget (C2)}, 
\name{} adapts batch capacity based on available GPU memory and reclaimable memory from paused requests under different context-handling policies. Besides, \name{} enforces bounded budget adjustments to ensure robustness.
We implemented \name{} with vLLM and evaluated it against both vLLM and INFERCEPT across multiple LLMs and GPU platforms. Experimental results show that \name{} consistently outperforms both baselines in latency and effective throughput. In particular, \name{} achieves a geometric mean of 6.5$\times$ and 4.7$\times$ higher effective throughput than vLLM and INFERCEPT, respectively, while reducing TTFT 95.6\% and 96.0\% on average.
In summary, our contributions are as follows:
\begin{itemize}[leftmargin=2em, itemsep=0pt, topsep=0pt, parsep=0pt, partopsep=0pt]

% \item We introduce \name{}, an efficient inference serving framework for augmented LLMs, which improves request queuing latency and effective throughput under dynamic workloads. (\S\ref{sec:design})
\item We present \name{}, an augmented LLM inference serving
framework that efficiently improves effective throughput (\S\ref{sec:design}).
% an efficient inference serving framework for augmented LLMs that mitigates HoL blocking and improves effective throughput. (\S\ref{sec:design})
\item We propose an adaptively state-aware request scheduling strategy that optimizes request ordering based on request characteristics, external-call-induced execution states, and runtime feedback. (\S\ref{sec:predict}, \S\ref{sec:scheduling})

% \item We propose an adaptively state-aware request scheduling strategy that optimizes request ordering based on request characteristics, execution state over the inference lifecycle, and runtime feedback.  (\S\ref{sec:budget}, \S\ref{sec:scheduling})
% \item We design a two-stage adaptive request scheduling strategy that adaptively optimizes request serving order based on request features and system state. (\S\ref{sec:two-scheduling})

% \item We develop a dynamic batch-level token budgeting mechanism that adjusts serving capacity in response to real-time load and memory pressure, improving throughput while preserving system stability. (\S\ref{sec:token-batch})
% \item We develop a dynamic batch-level budget mechanism adapting to free and preemptible memory. (\S\ref{sec:budget})
% \item We develop a dynamic batch-level token budget mechanism adapting to memory availability from running and paused requests. (\S\ref{sec:budget})
% \item We develop a dynamic batch-level token budget mechanism that adapts to free and reclaimable GPU memory from paused requests. (\S\ref{sec:budget})
\item We develop a dynamic batch-level token budget mechanism adapting to free and reclaimable memory. (\S\ref{sec:budget})

% \item We conduct extensive evaluations across multiple models and hardware platforms, demonstrating significant improvements in latency and effective throughput over existing systems. (\S\ref{sec:eval})
\item We conduct extensive evaluations to validate the effectiveness of \name{} (\S\ref{sec:eval}).
% \item We develop a dynamic token-level batching mechanism that adjusts the batch-level token budget adapting to real-time load. (\S\ref{sec:token-batch})
% and system conditions. (\S\ref{sec:token-batch})

% to confirm that \name{} outperforms baselines in inference performance (\S\ref{sec:eval}).
% the superiority of \name{} in inference performance over baselines.
% that \name{} outperforms baselines in inference performance (\S\ref{sec:eval}).
% \item We conduct extensive evaluations to confirm  that \name{} outperforms vLLM and INFERCEPT in inference performance (\S\ref{sec:eval}).
% the superiority of \name{} in inference performance over vLLM and INFERCEPT. (\S\ref{sec:eval})
% improving effective throughput and reducing queueing latency over vLLM and INFERCEPT. (\S\ref{sec:eval})
% SLO-driven performance over vLLM and INFERCEPT. (\S\ref{sec:eval})

\end{itemize}

\section{Background}
% \vspace{-3pt}
% This section discusses augmented LLMs and existing inference systems.
We review augmented LLMs and existing inference systems.

% \vspace{-8pt}
\subsection{Augmented Large Language Models}
% \vspace{-4pt}
% 随着LLM的快速发展，其纯文本生成能力和固定模型权重参数逐渐暴露出幻觉和缺乏实时知识等局限，难以满足不断扩展的应用需求。为克服这些缺陷，增强型LLM通过在推理过程中集成外部工具和资源(如API调用、数据库查询、外部模型等)，显著扩展LLM处理多样任务的能力，支持算术计算和实时web交互等功能。

% With the rapid development of LLMs,
% As LLMs rapidly evolve, their text-only capabilities and fixed model parameters have exposed limitations, such as hallucinations and lack of real-time knowledge \cite{toolformer,abouttime, nature24detecting}, making them inadequate for growing application demands. To address these issues, 
% 增强LLM在推理过程中整合了外部工具和资源（例如远程API、数据库查询、外部模型）和实时信息检索等复杂任务中展现出卓越的性能。此外，随着函数调用的普及和模型上下文协议（MCP的标准化，工具调用正日益成为现代推理流程中不可或缺的组成部。因此，增强型LLM推理系统正逐渐成为下一代云平台的核心基础设施。
Augmented LLMs integrate external tools (e.g., remote APIs, databases, external models) during inference \cite{ nips24advancing,toolformer, acl24gear, nature2025study}, demonstrating superior performance in complex tasks such as arithmetic computation \cite{toolkengpt,acl24good,toolbox} and real-time information retrieval \cite{dragin,abouttime}. 
Furthermore, with the rise of tool-using agents \cite{icml25BFCL,toolmaker} and the standardized tool interactions via the Model Context Protocol (MCP) \cite{mcp}, tool invocation has become a ubiquitous component of inference pipelines \cite{mcpzero, alm2023survey}. Consequently, augmented LLM inference systems are emerging as the core infrastructure for next-generation cloud platforms.

\subsection{Existing LLM Inference Systems}
% \vspace{-3pt}
% LLM推理服务已经成为现代数据中心重要的工作负载。许多推理系统被设计出来旨在提升整体的推理性能。
% 为了更高效地处理动态到达的各种序列长度不同的请求，Orca提出用迭代级调度的方式替代原先的请求级调度，以每一轮forward迭代处理为调度粒度，每每一轮forward前选择组成batch的请求，允许优先完成的短请求提前退出，以及新到达的请求可以立即加入下一轮batch。Orca充分利用GPU的并行处理能力，进一步提高了整体系统的吞吐量。
% 迭代级批处理已经广泛应用在现有的推理框架中
% LLM inference services have become a key workload in modern data centers, driving the development of various systems designed to improve overall performance. 
LLM inference has become a dominant workload in modern data centers, motivating the design of advanced systems to improve overall efficiency.
% \textbf{Iteration-level scheduling.} 
% To more efficiently handle dynamically arriving requests with varying sequence lengths, Orca \cite{orca} replaces request-level scheduling with iteration-level scheduling. Treating each forward iteration as the scheduling unit, Orca selects requests to form a batch before each iteration, allows finished short requests to exit early, and admits new requests to join the ongoing batch. This design fully leverages GPU parallelism and improves system throughput,and it has been widely adopted in existing inference frameworks.
To handle varying request sequence lengths, Orca \cite{orca} introduces iteration-level scheduling, which has become the de facto standard in state-of-the-art inference engines.
% which maximizes GPU parallelism and has become the de facto standard in state-of-the-art inference engines.
% This mechanism maximizes GPU parallelism and has become the de facto standard in state-of-the-art inference engines.
Meanwhile, to improve GPU memory utilization, vLLM \cite{vllm} proposes PagedAttention to eliminate memory fragmentation. 
% , enabling non-contiguous KV cache allocation to eliminate fragmentation.
Additionally, some research explores offloading KV cache to CPU or SSD \cite{flashgen, tightllm, flexgen} to alleviate GPU memory bottlenecks.
However, in augmented LLM inference, most prior works simply discard the context (KV cache) during external calls. INFERCEPT~\cite{infercept} improves over these approaches by dynamically selecting among \texttt{Preserve}, \texttt{Discard}, or \texttt{Swap} context-handling policies based on the external call duration and context length:
\begin{itemize}[leftmargin=2em, itemsep=0pt, topsep=0pt, parsep=0pt, partopsep=0pt]
    \item \texttt{Preserve}: The KV cache remains in GPU memory, and decoding resumes once the response returns.
    \item \texttt{Discard}: The KV cache is discarded, and recomputation is required after the response returns. 
    \item \texttt{Swap}: The KV cache is swapped to CPU memory and restored to GPU memory once the response returns.
\end{itemize}
This adaptive design avoids inefficient reliance on a single policy, reducing memory waste and inference latency.
% By balancing memory footprint and latency, this adaptive design avoids inefficient reliance on a single policy, reducing memory waste and inference latency.
% Since each policy entails different trade-offs between memory footprint and recomputation latency, this adaptive design avoids inefficient reliance on a single policy, thereby minimizing memory waste and reducing inference latency.
% Since each policy results in different memory waste, this adaptive design avoids inefficiently relying on a single policy and further reduces memory waste and inference latency.
% Since each policy entails distinct trade-offs between memory footprint and recomputation latency, this adaptive design avoids rigid reliance on a single policy, thereby minimizing memory waste and reducing inference latency

% Given that each of the above handling policy requires different trade-offs between memory footprint and recalculation latency, this adaptive design avoids inefficient reliance on a single policy, thereby minimizing memory waste and improving inference performance.

% 然而这些工作在增强LLM的推理服务中有效吞吐量仍然不高，在调度请求顺序上仍然采用FCFS的处理策略，以及静态的批处理大小设置，限制了增强LLM推理系统性能的进一步提升。
% Nevertheless, even with these advances, the effective throughput of augmented LLM inference services remains unsatisfactory. Existing systems still rely on FCFS scheduling and static batch tokens configurations, which fail to adapt to diverse and fluctuating workloads. These limitations reveal the need for adaptive scheduling in augmented LLM inference, motivating the design of our approach.

% \vspace{-5pt}
\section{Motivation}
% \vspace{-3pt}
\label{sec:mov}

% \subsection{Opportunities}
% \textbf{System goals}: Maximise throughput with the SLO.
% Augmented LLM inference introduces unique challenges beyond those of traditional LLM serving. 满足服务延迟目标下的有效吞吐量成为当前用户体验的性能瓶颈,existing approaches based on FCFS scheduling and static batch sizes fail to provide sufficient throughput or latency guarantees. 在这一章我们将对导致性能瓶颈的原因和潜在提高性能的机会进行深入分析。
% Augmented LLM inference introduces challenges that surpass those of LLM serving systems. 
% We focus on maximizing effective throughput under SLO constraints in augmented LLM inference services, as it directly impacts user experience.
% We aim to maximize effective throughput under SLOs in augmented LLM inference services, as this directly affects user experience. 
% We aim to maximize effective throughput under SLOs in augmented LLM inference services to improve user experience.
% We aim to maximize effective throughput under SLOs in augmented LLM inference services.
% However, existing systems rely on FCFS scheduling and static batch-level token budgets, which often struggle to meet these constraints, motivating the design of \name{}.

Maximizing goodput under SLOs is challenging in augmented LLM inference, where requests are heterogeneous and involve multi-round external calls. 
This section analyzes the key limitations of existing approaches.
% We aim to maximize effective throughput under SLO constraints in augmented LLM inference services.
% However, existing systems rely on external-call-unaware scheduling and static batch-level token budgets, which fundamentally limit performance, motivating the design of \name{}.

% However, existing systems relying on FCFS scheduling and static batch-level token budgets, which often struggle to meet these constraints. This motivates the design of \name{}.
% Existing systems relying on FCFS scheduling and static batch-level token budgets often struggle to meet these constraints, 
% motivating a systematic empirical investigation of key challenges and potential opportunities.

% Existing inference systems, which rely on FCFS scheduling and static batch-level token budgets, often struggle to satisfy these requirements. 
% However, existing inference systems that rely on FCFS scheduling and static batch-level token budget struggle to meet this requirement. 
% This motivates a systematic empirical investigation of current challenges and potential opportunities.

% fail to deliver satisfactory performance.
% to meet these demands. 

% In this section, 

% \vspace{-8pt}
% \subsection{Rigid FCFS Scheduling Policy}
% \subsection{Challenges in Request Scheduling for Augmented LLMs}
\subsection{Limitations of Existing Scheduling Strategies}
% \vspace{-3pt}
\label{sec:mov-fcfs}

% \begin{figure}[t]
%     \centering
%     \begin{minipage}[b]{0.45\linewidth}
%     \vspace*{0pt}  % 添加这行
%         \centering
%         \includegraphics[width=1.0\linewidth,height=2.8cm,keepaspectratio]{fig/mov/h800_slo_goodput.pdf}
%         % \vspace{-20pt}
%         \subcaption{   Goodput (higher better).}
%         \label{fig:fcfs_slo}
%     \end{minipage}\hfill
%     \begin{minipage}[b]{0.52\linewidth}
%     \vspace*{0pt}  % 添加这行
%         \centering
%         \includegraphics[width=1.0\linewidth,height=2.8cm,keepaspectratio]{fig/mov/h800_fcfs_ttft_replace.pdf}
%         % \vspace{-10pt}
%         \subcaption{   TTFT (lower better).}
%         \label{fig:fcfs-queue}
%     \end{minipage}
    
%     % \vspace{-5pt}
%     % \caption{   Effective throughput (req/s) and TTFT (s) with FCFS and random scheduling under different request rates.}
%     \caption{   Random scheduling outperforms FCFS: higher goodput (effective throughput) and lower TTFT.}
%     % \vspace{-8pt}
%     \label{fig:random}
% \end{figure}

\begin{figure}[t]
    \centering 
    \includegraphics[width=0.7\linewidth]{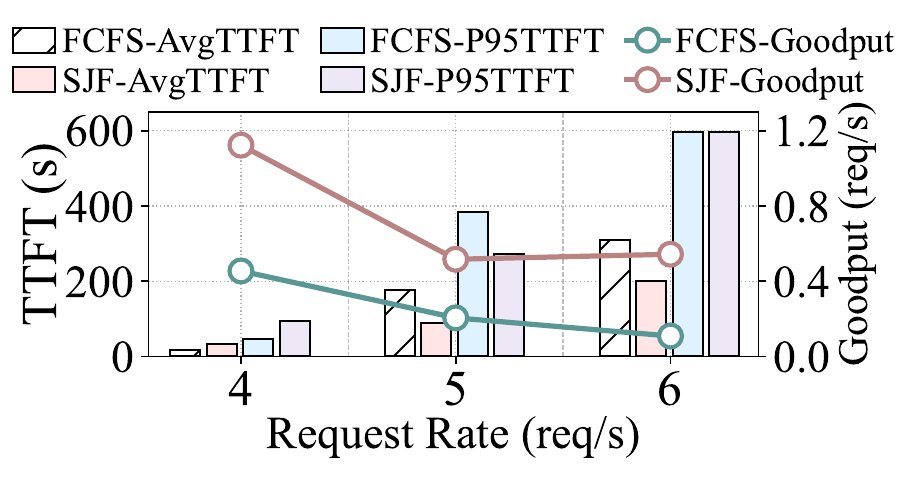}
    % \vspace{-9pt}
    \caption{
    % As the load increases, TTFT rises (driven by queueing latency), causing goodput (effective throughput) to drop.}
    % Effective throughput (req/s) and TTFT (s) with FCFS and SJF scheduling under different request rates.
  % SJF scheduling outperforms FCFS in both goodput (effective throughput) and TTFT, but remains suboptimal under high load.
  SJF outperforms FCFS in goodput (effective throughput) and TTFT, but remains suboptimal under high load.}

    % \vspace{-18pt}
    \label{fig:sjf}
\end{figure}

% 现有的推理服务系统采用FCFS的调度方式，只根据请求的到达时间在选择形成一个batch的请求，忽略请求间的输入、输出差异，但是这可能会导致先到达的长请求一直阻塞后面的短请求，增加后面请求的排队延迟，造成严重的Head-Of-Line(HOL)阻塞，从而导致整体性能的下降。在增强LLM推理中，对api返回长度无感知的情况将会进一步加剧内存资源浪费和整体延迟增加，如果返回大量token的长请求排在前面，会加剧HOL问题。
\textbf{Challenge 1: Existing inference scheduling strategies struggle in dynamic augmented LLM serving.}
% Existing scheduling leads to systemic inefficiency in dynamic augmented LLM serving workloads.}
% \textbf{Challenge 1: External-call-unaware scheduling leads to systemic inefficiency in augmented LLM inference.}
% 传统推理系统用基于到达顺序的FCFS，但是在增强型场景中，当长请求发生调用推理暂停，但是上下文仍占用大量GPU内存，会导致新的短请求被阻塞，引起严重的HoL问题。然后给一个实验表明仅仅是用random打乱请求排序，都能带来比fcfs更低的ttft和更高的有效吞吐量。但是random是粗粒度的随机打乱，性能提升也很有限。
% Current inference systems typically adopt FCFS scheduling and batch requests by arrival order, without considering external tool calls. In augmented LLM inference, when a long request triggers an external tool call, its execution is paused, while its context may be preserved in GPU memory, swapped out to host memory, or discarded and later recomputed after the call returns. In all cases, subsequent short requests can be blocked, causing severe HoL blocking, increasing queuing latency, and degrading effective throughput. 
Current inference systems typically adopt FCFS scheduling and batch requests by arrival order, without considering external calls. In augmented LLM inference, pausing requests for external calls (whether context is preserved, swapped, or discarded) blocks subsequent short queries. This induces severe HoL blocking, increasing latency and degrading goodput.
To alleviate HoL blocking, prior work explores approximate SJF scheduling based on request length~\cite{s3,nips24_rank,fastserve}.
While these approaches outperform FCFS (\autoref{fig:sjf}), they do not account for external calls and still fail to meet SLOs in augmented LLM inference, 
with TTFT often exceeding the SLO (e.g., 1s), high tail latency, and degraded goodput at high load.
Recent systems \cite{lamps} propose memory-based SJF heuristics that approximate job sizes using memory cost. 
However, these approaches make one-shot scheduling decisions at each execution round,
treating requests as newly arriving jobs after external calls. 
% treating request after an external call as a new request re-enters the system.
Consequently, they do not model dynamic and cross-round execution stages in augmented LLM workloads (\autoref{fig:request}), ignoring resumption costs from cumulative context and external call return lengths.

\begin{figure*}[t]
    % Centering the entire figure
    \centering
    % Using minipage to split the figure into two parts
    \begin{minipage}[b]{0.40\textwidth}
        \centering
        % \includegraphics[height=3.0cm]{fig/mov/out_and_api_length.pdf}
        % % \vspace{-5pt} % Add some space between image and subcaption
        % \caption{Input/Output/External\_call\_return length \\ distribution in INFERCEPT and ToolBench datasets.} % Subcaption for the left group
        %  \label{fig:api_length}
         \centering
        \includegraphics[height=3.0cm]{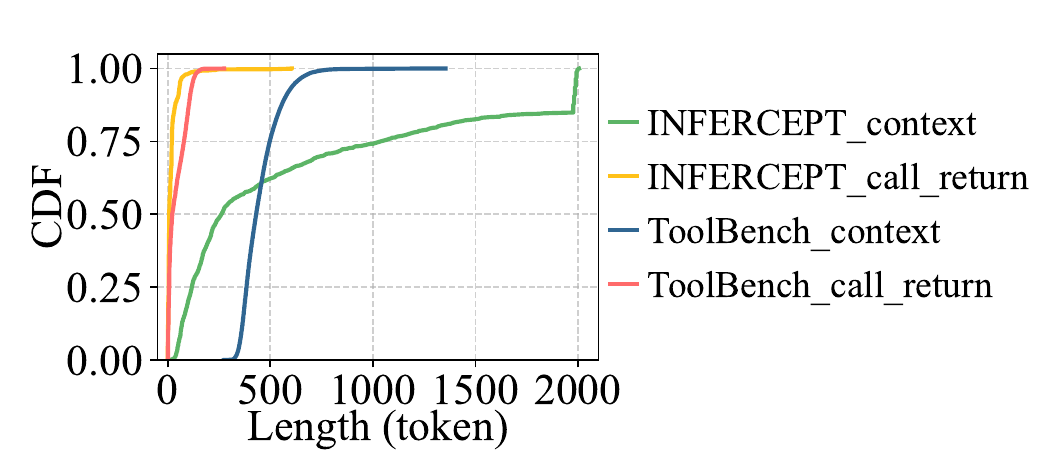}
        % \vspace{-5pt} % Add some space between image and subcaption
        \caption{Context/External\_call\_return length \\ distribution in INFERCEPT and ToolBench datasets.} % Subcaption for the left group
         \label{fig:api_length}
    \end{minipage}
    % \hfill % Add space between minipages
    % % Draw a vertical dashed line using tikz
% \begin{tikzpicture}[baseline={(0,-0.22cm)}] % Adjust -5cm to match image height
%         \draw [dashed, line width=1pt] (0,0) -- (0,3cm); % Height matches image
%     \end{tikzpicture}
    % \hfill % Add space between the line and the right minipage
    \begin{minipage}[b]{0.52\textwidth}
        \centering
        \includegraphics[height=3.0cm]{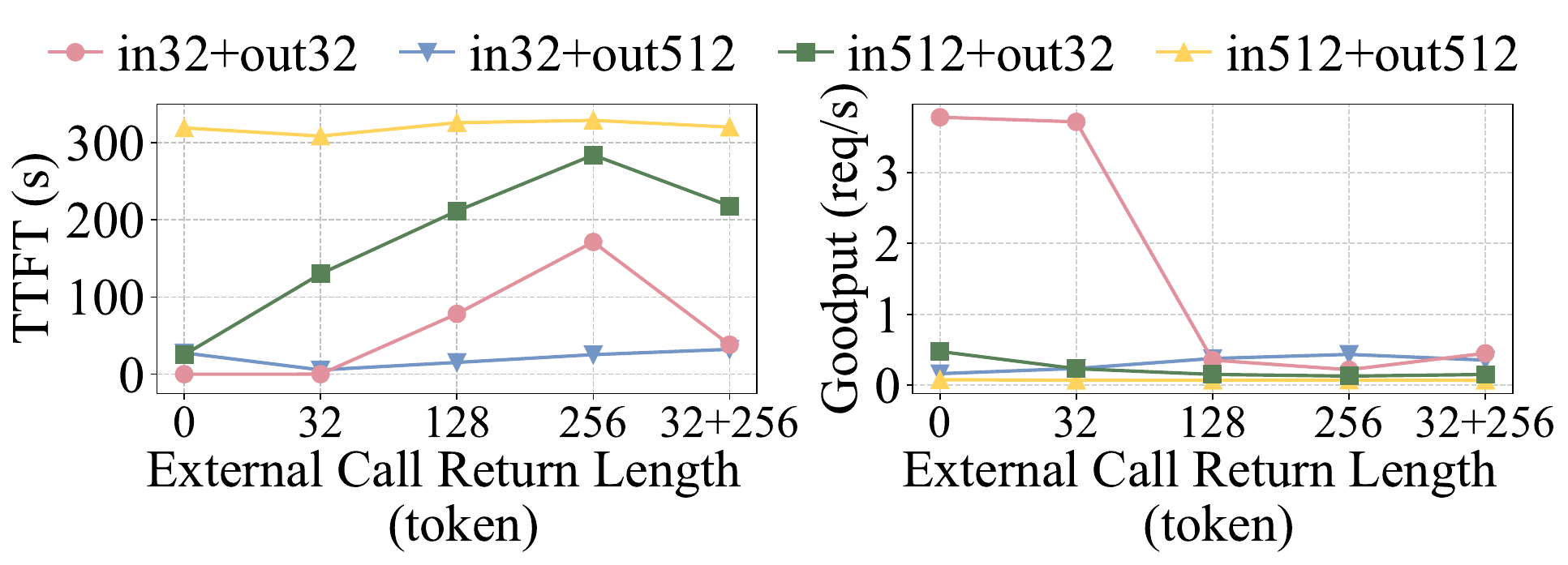} % Adjust path if needed
        % \vspace{-5pt} % Add some space between image and subcaption
        \caption{TTFT and goodput with varying external call return lengths, including fixed lengths and a mix of 32- and 256- tokens.} % 
        \label{fig:api_ttft}
    \end{minipage}
    % Adding a caption for the entire figure
    % \vspace{-18pt}
    % \caption{  Effective throughput (req/s) comparison among vLLM, INFERCEPT, and \name{} on Merge and ToolBench datasets with different models and GPUs. \normalfont{Higher is better.}}
    % Adding a label for referencing
    % \label{fig:slo_tput}
    % \vspace{-13pt}
\end{figure*}

Furthermore, we observe that both the cumulative context lengths at the time of external calls and the external call return lengths are highly variable (\autoref{fig:api_length}).
% This variability leads to dynamic resumption costs after external calls. 
% These costs depend on the context-handling policies, which have distinct memory footprints and recomputation overheads. 
Such variability results in dynamic resumption costs after external calls under different context-handling policies, which have distinct memory footprints and recomputation overheads.
% Under different context-handling policies (e.g., \texttt{Preserve}, \texttt{Discard}, \texttt{Swap})  with distinct memory footprints and recomputation overheads, such variability results in dynamic resumption costs after external calls.
Moreover, the varying external call return lengths significantly affect both TTFT and goodput (\autoref{fig:api_ttft}).
Together, these factors make existing scheduling policies suboptimal for augmented LLM inference.

% \textbf{Opportunity 1:}
% Augmented LLM inference calls for scheduling mechanisms that explicitly account for  heterogeneous requests with distinct memory footprints across execution states and dynamic resumption costs.
\textbf{Opportunity 1:}
Augmented LLM inference calls for scheduling mechanisms that explicitly account for heterogeneous execution states induced by external calls and dynamic resumption overhead.
% \textbf{Opportunity 1:} Augmented LLM inference calls for scheduling mechanisms that incorporate
% heterogeneous requests with distinct memory footprints and recomputation overhead.
% request-specific external call characteristics and adapt to evolving execution state and dynamic resumption cost.

% \textbf{Opportunity 1: Augmented LLM inference calls for scheduling mechanisms that adapt to evolving execution state and context-dependent resumption cost induced by external calls.}
% However, our analysis shows that external call return lengths are highly variable (~\autoref{fig:api_length}) and substantially affect latency and throughput (~\autoref{fig:api_ttft}).
% More importantly, context-handling policies during external calls (e.g., \texttt{Preserve}, \texttt{Discard}, \texttt{Swap}) introduce varying memory and recomputation costs that cannot be captured by token length alone. 
% Consequently, existing length-based scheduling policies remain suboptimal in augmented LLM inference, as further confirmed in \S\ref{sec:end-to-end}.

% 机会：增强LLM推理系统需要结合整体负载以及请求具体信息设计更细粒度的调度方案，能够降低整体的排队延迟，同时防止重尾问题，进一步提升系统的有效吞吐量。
% \textbf{Opportunity 1: Augmented LLM inference necessitates fine-grained scheduling that better adapts to request features and system state to optimize request ordering, improving queueing latency and boosting effective throughput.}
% \textbf{Opportunity 1: Augmented LLM inference requires fine-grained, external-call-aware scheduling to optimize request ordering, reduce queuing latency, and boost effective throughput.}

% \vspace{-8pt}
\subsection{Fixed Batch-Level Token Budget}
\label{sec:mov-budget}
% \vspace{-3pt}

% \vspace{-2pt}
\textbf{Challenge 2: Fixed batch-level token budget restricts throughput under external calls.}
% 批处理token上限决定了系统中一轮迭代forward处理的最多token个数，DeepSpeed-fastgen指出token上限比batch size（请求数量上限）对系统性能的影响更大。
% DeepSpeed-FastGen demonstrates that this limit has a greater impact on system performance than batch size (the maximum number of requests).
% 如果批处理token上限设置的太小，没有完全利用GPU的内存和计算能力，每一轮能处理的请求数量很少，整体的吞吐会很低；如果设置的太大，严重超出GPU的处理能力，会导致更多的资源争用，更多请求已经生成的KV cache所占用的GPU内存会被抢占释放，进而也就需要更多的KV重计算，拉低推理系统整体的处理效率。
The batch-level token budget determines the maximum number of tokens processed in a single forward iteration. 
In augmented LLM inference, external calls introduce paused requests whose contexts may occupy GPU memory, fundamentally complicating budget selection.
While a small budget leads to low per-iteration concurrency and reduced goodput, an overly large budget can trigger resource contention and unnecessary eviction of paused request contexts, incurring redundant recomputation overhead.
Recent systems~\cite{batchllm} adjust the budget based on free GPU memory, but do not model reclaimable memory from paused requests under \texttt{Swap} or \texttt{Discard} context-handling policies, failing to capture the true memory availability under external-call-augmented workloads and constraining goodput.

We evaluate statically configured batch-level token budgets across different hardware, models, and workloads, and observe that both overly small and excessively large budgets degrade goodput (\autoref{tab:maxbatch}). 
Moreover, the optimal budget varies across hardware and workload conditions.
% Building on the configuration in Section~3.1, we further run GPT-J-6B on an NVIDIA 4090 GPU to evaluate how different statically configured batch-level token budgets affect the effective throughput of the inference system. As shown in the table, both excessively small and overly large limits degrade effective throughput. Moreover, the optimal token limit varies across hardware platforms, model sizes, and workload levels.
% \begin{figure}[t]
%     \centering
%     \begin{minipage}[b]{0.49\linewidth}
%         \centering
%         \includegraphics[width=1.0\linewidth,height=4cm,keepaspectratio]{fig/mov/4090_6B_3.0_maxbatch_ttft.pdf}
%         \subcaption{SLO Attainment}
%         \label{fig:maxbatch_4090}
%     \end{minipage}\hfill
%     \begin{minipage}[b]{0.49\linewidth}
%         \centering
%         \includegraphics[width=1.0\linewidth,height=4cm,keepaspectratio]{fig/mov/h800_13B_4.0fcfs_maxbatch_ttft.pdf}
%         \subcaption{Queueing Latency}
%         \label{fig:maxbatch_h800}
%     \end{minipage}
    
%     \vspace{-5pt}
%     \caption{Average TTFT and P95 TTFT with Different Max Batch Tokens Limit, GPUs, Models and Request Rate}
%     \vspace{-10pt}
%     \label{fig:maxbatch}
% \end{figure}

% 机会：根据硬件信息、模型大小、具体的实时负载情况来动态设置批处理token上限将是最优的
% \textbf{Opportunity 2: Dynamically adapting the batch-level token budget to real-time load and hardware status is crucial for maximizing effective throughput and efficiently handling the context of requests paused by external calls.}
\textbf{Opportunity 2:} Dynamically adapting the batch-level token budget based on both free and reclaimable GPU memory from paused requests is crucial for maximizing goodput in augmented LLM inference.

\begin{table}[t]
% \vspace{-5pt}
\centering
\caption{Goodput (req/s) with different batch-level token budgets, optimal token budget varies across hardware and workloads.}
% \vspace{-6pt}
\renewcommand{\arraystretch}{0.65}
% \begin{tabular}{X|c|c|c|c|c|c|}
\begin{tabular}{ >{\centering\arraybackslash}m{3.53cm}|>{\centering\arraybackslash}m{0.38cm} >{\centering\arraybackslash}m{0.38cm} >{\centering\arraybackslash}m{0.52cm} >{\centering\arraybackslash}m{0.52cm} >{\centering\arraybackslash}m{0.52cm} }
% \hline
\toprule
 
\makecell{\textbf{Max batch tokens}}
& \textbf{100} &  \textbf{500} &  \textbf{1000} &  \textbf{1500} &  \textbf{2000} \\
% \hline
\midrule
 
\small{ 2req/s, GPT-J-6B, RTX4090}  & 0.29 & \textbf{0.41} & 0.25 & 0.11 & 0.10 \\
% \hline
 
\small { 4req/s, OPT-13B, H800 } & 0.01 & 0.16 & 0.18 & \textbf{0.22} & 0.19 \\
% \hline
\midrule
\end{tabular}
% \vspace{-18pt}
\label{tab:maxbatch}
\end{table}

\section{Problem Formulation}
% \vspace{-7pt}
% \subsection{Problem Formulation}
\label{sec:problem}
% \vspace{-3pt}

% To address the challenges in augmented LLM inference, this section we first formulate the multi-stage request lifecycle and then define the scheduling objective.
% To reason about scheduling in augmented LLM inference, this section formulates the request lifecycle and defines the corresponding scheduling objective.
To systematize request scheduling in augmented LLM inference, this section formalizes the request lifecycle and establishes the associated scheduling objectives.

\begin{figure}[t]
    \centering 
    \includegraphics[width=1.0\linewidth]{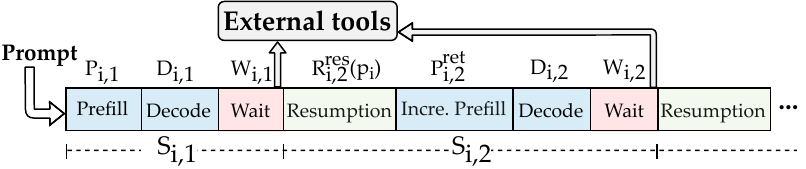}
    % % \vspace{-15pt}
    \caption{ Augmented LLM inference request lifecycle modeling.} 
    % % \vspace{-13pt}
    \label{fig:request}
\end{figure}

\textbf{Request Lifecycle Modeling.}
As shown in \autoref{fig:request}, in augmented LLM inference, a request spans multiple execution stages (e.g., prefill and decode) interleaved with external call waiting stages.
The response length returned by external calls determines the input size and KV cache state for the subsequent execution phase.
% The sequence length returned by external calls dictates the input size and KV cache state for the subsequent execution phase.
% In augmented LLM inference, an request span multiple execution stages and external call waiting stages, where context-handling policies fundamentally alter subsequent resource costs. 
% External calls acts as state transitions that introduce a significant idle period, and their responses determine the input size and KV cache state for the next execution stage.
% This segmentation supports fine-grained, state-aware scheduling by treating each segment as a composite unit with distinct resource requirements.
% follows the incremental execution paradigm adopted in INFERCEPT and 
% Each segment $S_{i,k}$ is a composite unit consists of execution and optional waiting stages for subsequent external call invocation with different resource requirements.
Accordingly, we model each request $R_i$ as a sequence of $n$ service segments, i.e., $R_i = \{S_{i,1}, \dots, S_{i,n}\}$, each ending with an external call.
Formally, the $k$-th segment $S_{i,k}$ is defined as:

% \vspace{-20pt}
% \vspace{-12pt}
\begin{equation}
  S_{i,k} =
\begin{cases}
\langle P_{i,1}, D_{i,1}, W_{i,1} \rangle, & k = 1, \\
\langle R_{i,k}^{{res}}(p_i), P_{i,k}^{{ret}}, D_{i,k}, W_{i,k} \rangle, & 2 \le k \le n.
\end{cases}
\end{equation}
% \vspace{-10pt}
% \vspace{-15pt}

% Although no computation is performed in this state, the request may still occupy GPU memory, incurring non-negligible residency overhead.
% The initial segment comprises the standard prompt prefill phase $P_{i,1}$, the decoding phase $D_{i,1}$, and a tool-wait state $W_{i,1}$, during which the request is suspended while awaiting external API responses, potentially causing memory residency overhead. 
% Subsequent segments additionally include a context resumption stage $R_{i,k}^{{res}}$, 
% whose cost is governed by the context policy $\pi \in \{\texttt{Preserve}, \texttt{Swap}, \texttt{Discard}\}$.
% This is followed by an incremental prefill phase $P_{i,k}^{\mathrm{ret}}$, which integrates new tool call return tokens into the KV cache before the next decoding and potential waiting states.

\begin{figure*}[t]
    % Centering the entire figure
    \centering
   % \vspace{-5pt}
        \includegraphics[width=0.90\linewidth]{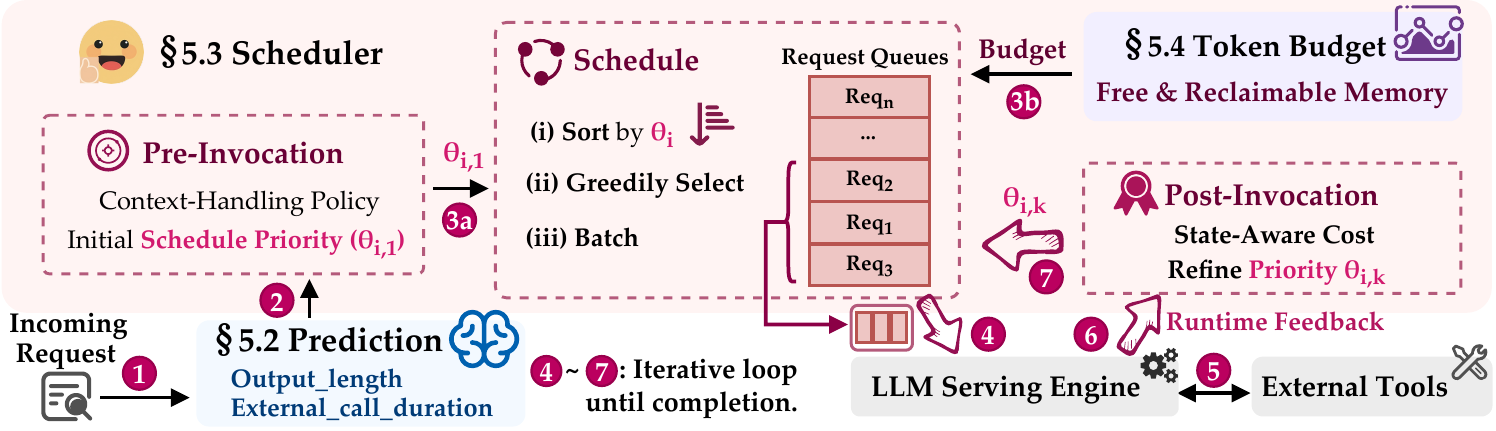}
        % % \vspace{-13pt} % Add some space between image and subcaption
        % \caption{The system architecture of \name{}, featuring adaptively state-aware scheduling and dynamic batch-level token budgeting. Solid arrows indicate control flow, numbers denote the execution order within scheduling iterations.} % Subcaption for the left group
        \caption{System architecture of \name{} with state-aware scheduling and dynamic batch-level token budget.}
% Solid arrows indicate control flow, and numbers denote the execution order within each scheduling iteration.}
         \label{fig:architecture}
         % % \vspace{-13pt}

\end{figure*}

% As shown in Figure~\ref{fig:request},
% during which the request awaits external tool responses and incur memory residency overhead. 

The initial segment $S_{i,1}$ comprises standard prompt prefill stage $P_{i,1}$, decoding stage $D_{i,1}$, and a tool-wait stage $W_{i,1}$.
Subsequent segments ($k \ge 2$) introduce a context resumption stage $R_{i,k}^{{res}}$, where ${res}$ denotes resuming a paused request under a context-handling policy $p_i \in \Pi = \{\texttt{Preserve, Swap, Discard}\}$.
This is followed by an incremental prefill stage $P_{i,k}^{{ret}}$, where ${ret}$ denotes incorporating tool return tokens into the KV cache.
% Subsequent segments ($k \! \ge \! 2$) introduce a context resumption stage $R_{i,k}^{{res}}$, whose cost is governed by context-handling policy $p_i \! \in  \!  \Pi \! = \!\{\texttt{Preserve, Swap, Discard}\}$, followed by an incremental prefill $P_{i,k}^{{ret}}$ to integrate tool return tokens into the KV cache.
This segmentation treats each segment as a composite unit with distinct, state-dependent resource requirements,  providing the necessary granularity for state-aware scheduling.
\textbf{Scheduling Objective.}
% Our goal is to improve the effective throughput of augmented LLM inference, which is influenced by both the service rate of requests and the queueing delays they experience.
Our goal is to maximize goodput in augmented LLM inference, accounting for both request service rates and queuing delays.
Accordingly, we jointly consider how many requests can be served per iteration and how much waiting time can be reduced.
To quantify the per-iteration benefit of scheduling a request, we define a \textit{\textbf{scheduling value $V_i^e$}} capturing both its throughput contribution and waiting-time reduction:
% by scheduling them.
% To capture the per-iteration benefit of scheduling a request, we define a scheduling value $V_i^e$ that reflects both its throughput contribution and waiting-time reduction:
% At each scheduling iteration, executing a request contributes one completed request to throughput and simultaneously eliminates the waiting time it has accumulated so far.
% Therefore, we define a scheduling value $V_i^e$ that quantifies the benefit of servicing request $i$ at iteration $e$, accounting for both throughput contribution and waiting-time reduction.
% Let $w_i^e$ denote the waiting time that would be eliminated if request $i$ is scheduled at iteration $e$. The scheduling value is defined as:
% Let $w_i^e$ denote the cumulative waiting time that request $i$ has incurred up to iteration $e$, which would be eliminated upon being scheduled.
% Let $w_i^e$ denote the cumulative waiting time of request $i$ up to iteration $e$.

% \vspace{-15pt}
% \vspace{-8pt}
\begin{equation}
\label{eq:value}
    V_{i}^e = 1 + \beta \cdot w_{i}^e, 
\end{equation}
% \vspace{-12pt}
% \vspace{-20pt}

where the constant term $1$ represents the fundamental contribution of serving a request to system throughput progress, $w_i^e$ denotes the waiting time that would be eliminated if request $i$ is scheduled at iteration $e$.
% while the term $\beta w_i^e$ quantifies the reduction in waiting time achieved by scheduling request $i$ at iteration $e$.
% The hyperparameter $\beta$ controls the trade-off between throughput-oriented scheduling and delay awareness.
The hyperparameter $\beta$ balances throughput-oriented scheduling and delay awareness.
A smaller $\beta$ prioritizes throughput by allowing the throughput term to dominate $V_i^e$, whereas a larger $\beta$ shifts the scheduling focus toward reduced waiting time.
Given the candidate request set $U^e$, let $\mathbf{x}^e = \{x_i^e \mid i \in U^e\}$ denote the batch selection decisions at iteration $e$, where $x_i^e \in \{0,1\}$ indicates whether request $i$ is selected for execution.
The scheduling objective is to maximize the aggregate scheduling value of the selected batch, subject to the available GPU memory budget $M^e$:
% The scheduling objective is to maximize the aggregate scheduling value of the selected batch, which jointly captures throughput advancement and waiting-time reduction,

% Given the candidate request set $U^e$ and the available GPU memory budget $M^e$, let $x_i^e \in \{0,1\}$ indicates whether request $i$ is selected for execution, the scheduling objective is to make the batch selection decisions to maximize the aggregate scheduling value:

% \vspace{-13pt}
% \vspace{-8pt}
\begin{equation}
\label{eq:goal}
   \begin{aligned}
\text{Maximize } \quad & \sum_{i \in U^e} x_i^e \cdot {V_i^e} \\
\text{s.t. } \quad & \sum_{i \in U^e} x_i^e \cdot m_i(t) \le M^e.
\end{aligned}
\end{equation}
% \vspace{-12pt}
% \vspace{-15pt}

% This formulation aligns with the Generalized Weighted Shortest Processing Time (WSPT) principle, which is optimal for minimizing weighted flow time under resource constraints. By prioritizing requests that yield the highest scheduling value per unit of space--time cost, \name{} effectively mitigates HoL blocking and improves overall system throughput.
% This formulation aligns with the Generalized Weighted Shortest Processing Time (WSPT) rule, which is theoretically optimal for minimizing weighted flow time in resource-constrained environments. By prioritizing requests that release the maximum potential value with minimum space-time cost, \name{} mitigates HoL  blocking and improves overall effective throughput.

% Moreover, the cost $C_i$ in augmented LLM is highly stochastic due to external tool calls. To solve this, we first need a mechanism to estimate these costs accurately (§4.2), before we can apply our scheduling strategy (§4.3).

% \vspace{-5pt}

\section{ Design}
\label{sec:design}

This section details the design of \name{}, a unified inference framework for augmented LLM serving.

\subsection{Design Overview}
As illustrated in ~\autoref{fig:architecture}, \name{} consists of three modules:
(1) \textbf{Prediction module} (\S\ref{sec:predict}) estimates $\textit{output\_length}$ and $\textit{external\_call\_duration}$ to provide priors for scheduling.
(2) \textbf{Scheduler module} (\S\ref{sec:scheduling}) employs a state-aware scheduling policy that incorporates external call characteristics and runtime feedback to optimize request ordering.
(3) \textbf{Token budget module} (\S\ref{sec:budget}) adjusts the batch-level token budget based on free GPU memory and preemptable paused-request contexts.
Together, these modules form a unified framework that reduces queuing latency and maximizes goodput for augmented LLM inference.

% (\S\ref{sec:budget}) adjusts batch-level token budget based on real-time memory availability. 

% \todo{system workflow}
% 系统工作的整体流程是：请求进入服务系统(①)，预测模块根据请求input信息预测输出长度和调用时长，为调度提供预测信息(②)。调度器根据预测信息预估请求发生调度时可能采用的上下文处理策略，估算请求的资源消耗，计算请求的调度优先级，根据优先级对请求进行排序(③)。在每一轮迭代中，动态批处理token模块根据当前负载和内存情况计算token上限，作为调度的约束(④)。调度器在当前的批处理token上限约束下，对排序后的请求队列进行选择请求组成一个batch来执行(⑤)。当推理引擎识别到请求中的外部调用，触发对应的外部调用，请求推理暂停(⑥)。等到调用返回后，调用期间实际应用的上下文处理策略和调用返回长度反馈给调度器，根据运行时反馈调度器计算请求的恢复开销，更新请求的调度价值，并根据对应的处理策略将请求放入对应的队列中，继续参与每一轮迭代的调度(⑧)，直至请求完成服务。
The overall workflow proceeds as follows:
An incoming request (\circled{1}) is first processed by the prediction module to estimate $\textit{output\_length}$ and $\textit{external\_call\_duration}$ as scheduling priors (\circled{2}). 
The scheduler utilizes these to determine the context-handling policy and assign an initial priority (\circled{3a}). 
Concurrently, the token budget module sets a batch-level token budget based on available GPU memory and reclaimable memory from preemptable paused requests (\circled{3b}).
% Concurrently, the token budgeting module sets a batch-level token budget based on real-time memory status (\circled{4}). 
Then the scheduler ranks requests by priority $\theta_{i}$ and selects requests to form an execution batch under the token constraint (\circled{4}). 
If an external tool is invoked, inference is paused until the call returns (\circled{5}), 
% If an external tool is invoked, the engine triggers the call and pauses inference (\circled{6}). 
after which runtime feedback is used to update the request’s scheduling priority (\circled{6}\circled{7}). 
% Upon call return, real-time data regarding the actual context policy and response length are fed back to the scheduler (\circled{7}). 
% The scheduler uses this feedback to calculate the actual resumption cost and refine the request’s $k$-th service segment priority $\theta_{i,k}$. 
Requests are iteratively scheduled until completion.
\subsection{Lightweight Request Prediction}
\label{sec:predict}
% To provide $\textit{output\_length}$ and $\textit{external\_call\_duration}$ for scheduling, 
% we fine-tune a lightweight BERT-base model (110M parameters) to jointly estimate these two quantities.
To support scheduling, 
we fine-tune a lightweight BERT-base model (110M parameters) to jointly estimate 
$\textit{output\_length}$ and $\textit{external\_call\_duration}$,
which provide coarse-grained priors about each request’s future execution behavior.
$\textit{Output\_length}$ prediction is formulated as a classification task by discretizing sequences
into fixed 50-token buckets for robustness \cite{shuffleinfer,s3}.
Using request inputs as features and realized output lengths as labels, the model achieves 85\% bucket accuracy on ToolBench and 65\% on the INFERCEPT dataset.
$\textit{External\_call\_duration}$  is predicted via regression using request input metadata as features and actual execution times as labels, with mean squared errors of approximately 5s on ToolBench and 0.4s on INFERCEPT.
In addition, the predictor is invoked upon request arrival and tool call return, running off the critical execution path, and adds less than 0.1\% overhead to total execution time (\S\ref{sec:ablation}).
Importantly, inaccuracies in early predictions are mitigated by continuously refined scheduling decisions (§\ref{sec:scheduling}).
We empirically validate this robustness to prediction errors in \S\ref{sec:prediction_robustness}.
\subsection{Adaptive Scheduling with State Awareness}
\label{sec:scheduling}
% Since the problem is NP-Hard (Knapsack variant), we propose a Density-based Greedy Algorithm augmented with a Two-Stage Adaptive Mechanism to handle the uncertainty of $C_i$.
% Since finding the optimal subset $\mathbf{x}$ to maximize Eq.~\ref{eq:goal} is NP-Hard, which is equivalent to the Knapsack Problem, we adopt a density-based greedy strategy augmented with Continuous State Re-evaluation to handle the dynamic $C_i$.  This approach treats scheduling as a rolling decision process: at the start of each segment, the cost is re-assessed based on the latest ground truth. For each candidate request $i$ in the global pool, we define the ranking priority $\theta_i$ as the value density, representing the potential value gained per unit of space-time resource consumed.
% 这是一个np-hard的背包问题，然后也是为了解决动机挑战一，我们提出基于价值密度的贪心算法，优先那些具有更高系统效率边际收益的请求
% Given the scheduling objective defined in Eq.~\eqref{eq:goal}, the scheduler must determine which request segments to execute under strict GPU memory constraints. The optimization problem formulated in Eq.~\eqref{eq:goal} is a variant of the classic NP-hard 0-1 knapsack problem. 
% The FCFS-based systems scheduling performs poorly with only account for arrival order \S\ref{sec:mov-fcfs}. Our goal is making scheduling decisions to maximize the efficiency with limited GPU memory. 
% Accordingly, we propose a value density-based greedy scheduling strategy that prioritizes request segments with higher marginal system benefit per unit of resource consumption.
% \textbf{Density-based Greedy Algorithm.}
\textbf{Value-Density-Based Scheduling Policy.}
The scheduling task defined in~\autoref{eq:goal} is a variant of the NP-hard 0-1 knapsack problem, aiming to pack the maximum aggregate value into limited GPU memory. To address this, we propose a greedy value-density strategy that ranks requests by their scheduling efficiency, prioritizing those that yield \textbf{higher scheduling value (\autoref{eq:value}) per unit of resource consumption}. We define $C_{i,k}$ as the resource consumption of the $k$-th service segment of request $i$.
% , which depends on the request’s execution state in augmented LLM inference.
% which is inherently state-dependent under augmented LLM inference. 
The scheduling priority (value density) is then computed as: 
% Existing inference approaches with FCFS scheduling is inadequate, as it ignores different request footprints and causes HoL blocking (\S\ref{sec:mov-fcfs}). 
% To address this, we propose a greedy value-density strategy that ranks requests by their efficiency ratio, prioritizing those that generate higher scheduling value (~\autoref{eq:value}) with unit resource consumption.
% This high-efficiency selection policy enables the scheduler to maximize aggregate throughput while minimizing queuing delays under stringent resource constraints.
% of each segment is then computed as: 
% While the scheduling value $V_i$ captures a request’s contribution to throughput progress and waiting-time reduction, we define the $C_{i,k}$ to denote the resource consumption of $k$-th service segment of request $i$, which is inherently state-dependent under augmented LLM inference. 
% Combining these factors, we define the scheduling priority (value density) of each service segment as: 
% which reflects execution efficiency under the current system state.
% the space--time cost $C_{i,k}$ reflects execution efficiency under the current system state.
% Combining these two factors, we define the scheduling priority (value density) of segment $S_{i,k}$ as:
% We combine these factors to define the scheduling priority (value density) of segment $S_{i,k}$ as:
% $S_{i,k}$ as:

% \vspace{-10pt}
% \vspace{-8pt}
\vskip -7pt
\begin{equation}
\label{eq:theta}
   \theta_{i,k} = \frac{V_i}{C_{i,k}} = \frac{1 + \beta \cdot w_i}{C_{i,k}}.
\end{equation} 
\vskip -3pt
% \vspace{-8pt}
% \vspace{-10pt}

% where $V_i$ reflects the throughput contribution and aging-based fairness, and $C_{i,k}$ is the segment-level space-time cost defined in Eq.~\eqref{eq:cost}.
% With the scheduling value $V_i$ captures a request’s contribution to throughput progress and waiting-time reduction,
Here, scheduling value $V_i$ captures a request’s contribution to throughput progress and waiting-time reduction, computed by tracking the elapsed waiting time $w_i$ since its last scheduling.
This density-based ranking prioritizes requests with higher scheduling efficiency, enabling the scheduler to maximize aggregate throughput while minimizing queuing delays. 
The remaining key challenge lies in accurately modeling the state-dependent resource consumption $C_{i,k}$ across heterogeneous execution phases, which we address next.
% under stringent resource constraints.
% that deliver higher scheduling value per unit of resource, thereby maximizing overall system performance under limited GPU memory and variable external call delays.
% This high-efficiency selection policy enables the scheduler to maximize aggregate throughput while minimizing queuing delays under stringent resource constraints.
% With the scheduling framework established, the core challenge lies in accurately quantifying the space-time cost $C_{i,k}$ across different execution stages, which we detail next.
% We calculate scheduling value $V_i$ by recording the request's latest scheduling time and the current time to obtain the request's waiting time $w_i$. 
% We compute the scheduling value $V_i$ by tracking the elapsed waiting time $w_i$ since the request’s latest scheduling time.
% in augmented LLM inference, which we address next.
% The remaining key challenge lies in accurately calculating and measuring the resource consumption of augmented LLM inference requests across different execution phases.

% 调度器将请求视为一个状态机。在每个服务段（Segment）开始时，根据当前系统状态（Cache Location）和未来预测（Future Prediction），动态生成该段的时空成本（Space-Time Cost）。

% \textit{\textbf{Space-Time Cost Modeling.}}
% However, 
% In augmented LLM serving, defining resource consumption is non-trivial.
% In augmented LLM serving, accurately defining resource consumption is non-trivial due to the highly variable external calls behaviors and context management.
% External calls introduce highly variable paused-waiting durations and response sizes, 
\textbf{State-Aware Resource Consumption Modeling.}
In augmented LLM serving, resource consumption depends not only on request length but also on external call behavior and context-handling policies.
External calls induce variable pause durations and response sizes,
while different context-handling policies incur distinct memory and recomputation costs.
In addition, execution costs vary across service stages, including prefill, decoding, waiting, and resumption.
These factors make length-based or instantaneous memory-based cost models inadequate.
% In addition, execution costs vary across service stages, including prefill, decoding, and external waiting phases.
% cost models inadequate, motivating a state-aware, segment-level space-time cost formulation that captures both memory and temporal resource usage.
% During external calls, different context policies induce different memory and computation costs. Moreover, execution costs vary significantly across service stages, such as prefill, decoding, and external waiting phases.
% Consequently, request length alone is insufficient to capture the true resource consumption of augmented inference, motivating a more expressive cost model.
% Due to the execution heterogeneity and resource cost uncertainty introduced by external calls (\S\ref{sec:mov-call}), conventional length-based metrics fail to capture the multi-dimensional resource footprints across different service segments.
% To more accurately capture resource usage,
We therefore adopt a space-time cost model that quantifies resource consumption as the integral of memory occupancy over the segment’s service duration.
Formally, the space-time cost of service segment $S_{i,k}$ under context-handling policy $p_i$ is defined as:
% we adopt a space-time cost model that measures the resource consumption of a service segment as the integral of memory occupancy over its execution time.
% To this end, we introduce space-time cost modeling that quantifies the total resource consumption as the integral of its memory occupancy over time
% To address this, we introduce space-time cost modeling,
% which quantifies resource consumption as the cumulative product of memory footprint and residency duration
% which integrates dynamic memory occupancy over the residency duration of a service segment. 
% Under this model, the cost of a service segment $S_{i,k}$ represents the total memory capacity held by a request over time rather than just its computational load. We formally define this cost as the integral of the instantaneous memory occupancy $m_i(t,p_i)$ of request $i$ under policy $p_i$ over the total residency duration $\tau_{i,k}$:
% Unlike static metrics, this approach quantifies the opportunity cost of GPU memory being occupied during external waiting stages or recomputation phases. We formally define the space-time cost of segment $S_{i,k}$ under policy $p_i$ as:
% To formally quantify such consumption, we define the space--time cost of segment $S_{i,k}$ as:

% \vspace{-15pt}
% \vspace{-10pt}
\vskip -3pt
\begin{equation}
\label{eq:cost}
    C_{i,k}(p_i) =   \int_{0}^{\tau_{i,k}} m_i(t, p_i) \, dt ,
\end{equation}
\vskip -3pt
% \vspace{-10pt}
% \vspace{-10pt}
 
% This formulation captures both the amount of GPU memory used and the duration it is held, 
where $m_i(t,p_i)$ denotes the instantaneous memory occupancy of request $i$ under policy $p_i$, and $\tau_{i,k}$ is the segment’s residency duration. 
This formulation measures how long and how much GPU memory a request occupies over its service duration,
providing a unified metric to evaluate resource overhead across heterogeneous execution stages.
To enable effective scheduling, 
% To enable effective density-based scheduling, 
we further construct $C_{i,k}$
% we further construct the space-time cost $C_{i,k}$ 
in a state-aware manner, conditioned on the execution state, context-handling policy, and realized external call outcomes.
This construction proceeds in two regimes, including Pre-Invocation estimation and Post-Invocation refinement.
\textit{\textbf{(1) Pre-Invocation Estimation:}}
Upon request arrival, \name{} estimates the space-time cost $C_{i,1}$ to compute the initial scheduling priority $\theta_{i,1}$ before the external call returns. 
Specifically, a lightweight prediction module (\S\ref{sec:predict}) provides the predicted \textit{external\_ call\_duration} \(\hat{\tau}_{i}^{call}\) and \textit{output\_length} \(\hat{l}_{i}^{out}\) \footnote{We use $\hat{\cdot}$ to denote predicted values throughout the paper.}. 
% Specifically, the system invokes a lightweight prediction module (\S\ref{sec:predict}) to estimate the external call duration $\hat{\tau}_i^{call}$ and the output length $\hat{l}_i^{out}$.
% and following the memory-waste minimization principle in INFERCEPT, 
With these predictions, 
\name{} selects an initial context-handling policy $\hat{p}_{i,1}$ 
that minimizes the expected memory waste $\text{Waste}_{i}^{p_i}$ caused by the external call:
% that minimizes the expected waiting-phase memory waste $\text{Waste}_i^p_i$ under the external call:
% Specifically, the system first invokes a prediction module (\S\ref{sec:predict}) to estimate call duration ($\hat{\tau}_i^{call}$) and output length ($\hat{l}_i^{out}$). 
% With prediction information and follow the cost-minimization principle in INFERCEPT, \name{} selects a tentative context-handling policy $p_i$ to minimize memory waste $\text{Waste}_i^p_i$ during the waiting phase:

% \vspace{-15pt}
% \begin{equation}
%     \hat p_i_{i,1} = \arg\min_{p_i} \text{Waste}_i^p_i(\hat{l}_i^{out}, \hat{\tau}_i^{call}) , \quad p_i \in p_i
% \end{equation}
% \vspace{-20pt}

% \vspace{-17pt}
% \vskip -8pt
\vskip -2pt
\begin{equation}
\label{eq:context_policy}
    \hat {p}_{i,1} = \arg\min_{{p} \in \Pi} \text{Waste}_{i}^{p}(\hat{l}_{i}^{out}, \hat{\tau}_{i}^{call}) .
\end{equation}
\vskip -2pt
% \vskip -10pt
% \vspace{-18pt}

Detailed formulations and symbol definitions are provided in Appendix \ref{app:policy}. 
Next, \name{} constructs the cost ${C}_{i,1}$ using \autoref{eq:cost}, combining: 
(i) the execution cost during the prefill $C_{i,1}^{prefill}$ and decoding stages $C_{i,1}^{decode}$ ($P_{i,1}$ and $D_{i,1}$ in \autoref{fig:request}), and (ii) the memory residency cost $C_{i,1}^{call}$ during the external call under policy $\hat{p}_{i,1}$ ($W_{i,1}$ in \autoref{fig:request}):
\begin{equation}
 C_{i,1}(\hat p_{i,1})  \!\!
=  C_{i,1}^{prefill}  + C_{i,1}^{decode} + C_{i,1}^{call}(\hat p_{i,1})
\end{equation}
% \vspace{-13pt}
\begin{equation}
C_{i,1}^{call}(\hat p_{i,1}) \!=\!   
\begin{cases} 
C_{i,1}^{preserve} & \!\!\! \hat{p}_{i,1}\! =\! \texttt{Preserve} \\
0             & \!\!\! \hat{p}_{i,1} \!= \!\texttt{Discard} \\
C_{i,1}^{swap-out}  & \!\!\! \hat{p}_{i,1} \! = \!\texttt{Swap}
\end{cases}.
\end{equation}
With this estimated space-time efficiency, the scheduler initializes request priority $\theta_{i,1}$ with \autoref{eq:theta}.
\textit{\textbf{(2) Post-Invocation Refinement:}} 
% When the external call returns, request transitions from a tool-wait to a resumption state, \name{} adjusts the resource consumption of the request based on actual runtime feedback.
When the external call returns, the request transitions from the tool-wait state to the resumption state. \name{} updates the resource consumption based on actual runtime feedback. 
Specifically, the realized return length and the context-handling policy $p_{i,k-1}$ applied during the previous waiting phase are fed back to refine the cost of the next segment.
% Specifically, the system feeds back the realized return length and the context-handling policy $p_i_{i,k-1}$ applied during the waiting stage to update the cost of the next segment.
To capture these evolving resource demands, \name{} first computes a feedback realization term $C_{i,k}^{fb}$ that quantifies the actual overhead incurred during the transition,
including the context resumption stage ($R^{res}_{i,2}(p_i)$ in ~\autoref{fig:request}) and the incremental prefill stage for return tokens ($P^{ret}_{i,2}$ in~\autoref{fig:request}):
\vskip -8pt
\begin{equation}
 \label{eq:c_fb}
C_{i,k}^{fb}(p_{i,k-1}) \!=\! C_{i,k}^{return } + \underbrace{
\begin{cases} 
0 & \!\!\! {p}_{i,k-1}\! =\! \texttt{Preserve} \\
C_{i,1}^{recompute}             & \!\!\! {p}_{i,k-1} \!= \!\texttt{Discard} \\
C_{i,1}^{swap-in}  & \!\!\! {p}_{i,k-1} \! = \!\texttt{Swap}
\end{cases}}_{\text{Resumption Cost}} \!\! .
\end{equation}
% \vskip -2pt
% \begin{equation}
% \label{eq:c_fb}
% C_{i,k}^{fb}(p_{i,k-1}) = \underbrace{\int_{t \in \tau_{i,k}^{res}}  m_i(t, p_{i,k-1}) dt}_{\text{Resumption Cost}} + \underbrace{\int_{t \in \tau_{i,k}^{ret}} m_i(t) dt}_{\text{Return Tokens Cost}}
% \end{equation}

% \vspace{-8pt}
% \vspace{-15pt}

% In this formulation, Resumption Cost captures the realized overhead of restoring execution state under the previously applied context policy $p_i_{i,k-1}$.
% For example, if a request was discarded during the waiting phase, this term reflects the substantial KV-cache recomputation cost.
% Return Tokens Cost quantifies the actual resource occupancy required to process the returned tokens.
% Here, Resumption Cost reflects the actual overhead of restoring execution state under the context-handling policy $p_i_{i,k-1}$ of previous external call.
% Here, Resumption Cost reflects the actual overhead of restoring execution state under $p_i_{i,k-1}$.
% For example, if a request adopt \texttt{Discard} policy, this term reflects the KV-cache recomputation cost.
Here, Resumption Cost reflects the overhead of restoring the execution state under $p_{i,k-1}$. 
For instance, if $p_{i,k-1}$ adopts the \texttt{Discard} policy, this term captures the KV cache recomputation cost.
Return tokens cost
$C_{i,k}^{return}$ reflects the memory required to process the call returned tokens.
By replacing predictive estimates with realized execution feedback, $C_{i,k}^{fb}$ resolves uncertainty from prior predictions. 
Subsequently, \name{} predicts the remaining execution ($D_{i,2}$ in \autoref{fig:request}) and any subsequent external calls ($W_{i,2}$ in \autoref{fig:request}) to construct the final $C_{i,k}$:
% By replacing predictive estimates with realized execution feedback, the feedback-driven term $C_{i,k}^{fb}$ settles the resumption cost under the actual context-handling policy, resolving uncertainty carried over from earlier predictions.
% Subsequently, \name{} incorporates a new round of prediction for the remaining execution and any potential subsequent tool call to construct the final $C_{i,k}$:
% Finally, \name{} synthesizes the total cost $C_{i,k}$ by augmenting the realized feedback with updated predictions for the remaining decoding phase and potential subsequent tool calls:

% \vspace{-16pt}

% \begin{equation} 
% \label{c_ik_new}
% C_{i,k} \!=\! C_{i,k}^{fb}\!+\! \underbrace{\int_{t \in \tau_{i,k}^{exec}} \!\!m_i(t) dt}_{\text{Decoding Cost}} + \!\underbrace{\int_{t \in \hat{\tau}_{i,k}^{call}}\!\!\! m_i(t, \hat p_{i,k}) dt}_{\text{Future Call Cost}}
% \end{equation}
\vskip -8pt
\begin{equation} 
\label{c_ik_new}
C_{i,k}(\hat p_{i,k}) \!=\! C_{i,k}^{fb}(p_{i,k-1})\!+ C_{i,k}^{decode} + C_{i,k}^{call}(\hat p_{i,k}).
\end{equation}
\vskip -4pt

% \vspace{-8pt}
% \vspace{-15pt}

% This segment-level refinement replaces cumulative uncertainty with runtime feedback, balancing historical resource demands against anticipated future overheads, aligning scheduling decisions with actual system states and effectively mitigating prediction-induced inefficiencies.
% This segment-level refinement uses realized execution feedback to correct prior predictive estimates, integrating observed and anticipated resource demands to align scheduling decisions with actual system states and mitigate inefficiencies from earlier predictions.
This segment-level refinement uses realized execution feedback to correct prior predictive estimates, aligning scheduling decisions with actual system states and effectively mitigating inefficiencies from earlier predictions. For requests comprising multiple service segments, this refinement process repeats iteratively until the request completes.
\textbf{Scheduling Procedure.}
At each scheduling decision, the scheduler proceeds as follows:
\begin{itemize}[leftmargin=2em, itemsep=0pt, topsep=-2pt, parsep=0pt, partopsep=0pt]
\item \textbf{Cost Construction:}
For each request, construct its space-time cost $C_{i,k}$ and compute the priority $\theta_{i,k}$.

\item \textbf{Ranking:} All requests in the global pool $U^e$ are ranked in descending order of their current value density $\theta_i$.

\item \textbf{Greedy Packing:}
Requests are selected greedily from the ranked queue until the aggregate memory footprint $\sum x_i^e \cdot m_i(t)$ reaches the system memory limit $M^e$.
% limit $M^e$.
\end{itemize}
% Select requests from the top of the ranked list ($x_i^e = 1$) until the total memory occupancy $\sum x_i^e \cdot m_i(t)$ reaches the memory limit $M^e$.
    % \item \textbf{Packing:} The scheduler greedily selects requests from the top of the ranked list to form a batch $B^e$ ($x_i^e = 1$).
    % \item \textbf{Constraint Check:} Requests are added until the cumulative instantaneous memory occupancy $\sum_{i \in B^e} m_i(t)$ reaches the dynamic memory 

% This procedure dynamically adapts to the evolving request pool, requests execution states, and external call outcomes.\
This procedure dynamically schedules requests based on state-aware space-time costs and value-density priorities, adapting to evolving request states and external calls. 
The complete pseudocode is provided in Appendix~\ref{appendix:symbol},
and a theoretical analysis is provided in Appendix~\ref{app:proof}.

% The complete pseudocode of the scheduling procedure is provided in Appendix~\S\ref{appendix:symbol}.

% With the scheduling framework established, the core challenge lies in accurately quantifying the space-time cost $C_{i,k}$ across different execution stages, which we detail next.

\subsection{Dynamic Batch-Level Token Budget}
\label{sec:budget}
% The memory constraint in Eq.~\eqref{eq:goal} is enforced as a batch-level token budget, limiting the number of tokens processed in each scheduling iteration.
% At each iteration, the scheduler ranks requests and greedily packs them under the current budget.
% Traditional systems use a fixed token budget, which can underutilize GPU memory or force discarding of paused request contexts in augmented LLM inference with external calls.
% To maximize effective throughput, \name{} adapts the token budget based on the available GPU memory and the contexts of paused requests eligible for preemption.
In the scheduling procedure (\S\ref{sec:scheduling}), 
the memory constraint $M^e$ governs greedy packing through a batch-level token budget that limits the number of tokens processed per iteration.
% the memory constraint $M^e$ governs greedy packing, presented as a batch-level token budget that limits the number of tokens processed.  
% This is enforced as a batch-level token budget, limiting the number of tokens processed in each scheduling iteration.
However, using a fixed token budget can underutilize GPU memory or trigger unnecessary discarding of paused request contexts in augmented LLM inference.
% with external calls.
% force discarding of paused request contexts in augmented LLM inference with external calls.
We dynamically adjust the token budget based on 
(i) currently available GPU memory $G_{{free}}(t)$, and 
(ii) reclaimable memory $G_{\text{kv}}^{{preempt}}(t)$ from preemptable paused requests, i.e., those selecting \texttt{Swap} or \texttt{Discard} policies by \autoref{eq:context_policy}.
Formally, the token budget $\mathcal{B}_{{token}}(t)$ is computed as:
% Specifically, we set only the paused requests adopting \texttt{Swap} or \texttt{Discard} policies (\autoref{eq:context_policy}) are considered preemptable.
% Formally, the token budget $\mathcal{B}_{\text{token}}(t)$ is computed as:
% To address this, we dynamically adjust the batch-level token budget based on available GPU memory and reclaimable memory from paused requests.
% Specifically, only the paused requests which selects the \texttt{Swap} or \texttt{Discard} policy by \autoref{eq:context_policy} are considered preemptable.
% Formally, let $G_{\text{free}}(t)$ denote free GPU memory, and $G_{\text{kv}}^{\text{preempt}}(t)$ the memory of preemptable paused requests. 
% The token budget $\mathcal{B}_{\text{token}}(t)$ is calculated as:

% \vspace{-20pt}
% \begin{equation}
% \label{eq:token}
% % \small
% \mathcal{B}_{token}(t)\!= \!\left\lfloor \!\frac{G_{{free}}(t)\! + \!\gamma \cdot G_{{kv}}^{{paused}}(t)}{{\mu}} \!\right\rfloor,  \gamma \!\in \! [0,1],
% \end{equation}
% \vspace{-15pt}

% \vspace{-13pt}
% \vspace{-8pt}
\vskip -6pt
\begin{equation}
\label{eq:token}
\mathcal{B}_{{token}}(t)
= \left\lfloor
\frac{G_{{free}}(t) + G_{{kv}}^{{preempt}}(t)}{\mu}
\right\rfloor,
\end{equation}
\vskip -3pt
% \vspace{-8pt}
% \vspace{-13pt}

% Where ${{\mu}}$ denotes the per-token footprint and $\gamma$ controls the reclaimable fraction of paused-context memory. When memory demand exceeds available physical capacity, paused requests become subject to swapping or discarding.
% By default, $\gamma \! = \! 1$ allows paused contexts to serve as elastic buffers, triggering swapping only when memory demand exceeds available physical capacity.
% allows pause blocks to act as elastic buffers and trigger swapping only when demand exceeds physical free memory. 
% To avoid instability from transient memory fluctuations, 
where $\mu$ denotes the per-token memory footprint.
To prevent instability from transient memory fluctuations, $\mathcal{B}_{{token}}(t)$ is clipped to
$[\beta_{{low}} \cdot  T_{\max}, \beta_{{high}} \cdot T_{\max}]$, where $T_{\max}$ is a reference offline budget and $\beta_{{low}}, \beta_{{high}}$ are scaling factors.
This design balances throughput with controlled use of preemptable paused requests’ memory while maintaining scheduling stability under fluctuating GPU memory conditions.

\begin{figure*}[t]
    \centering
    
    % --- 第一行 ---
    \begin{minipage}[b]{0.48\textwidth}
        \centering
        \includegraphics[width=\textwidth]{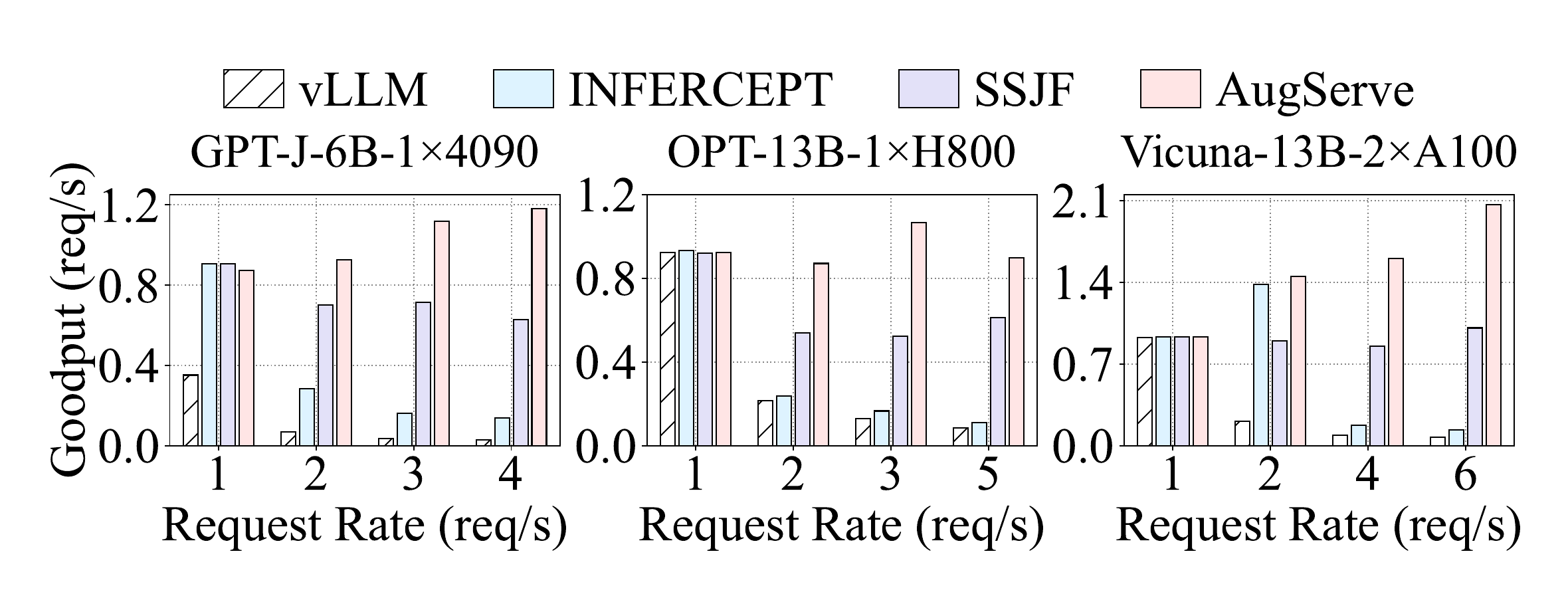}
    \end{minipage}
    \hfill
    \begin{tikzpicture}[baseline={(0,0)}] 
        \draw [dashed, line width=1pt] (0,0) -- (0,3cm); % 根据图片高度微调 3cm
    \end{tikzpicture}
    \hfill
    \begin{minipage}[b]{0.48\textwidth}
        \centering
        \includegraphics[width=\textwidth]{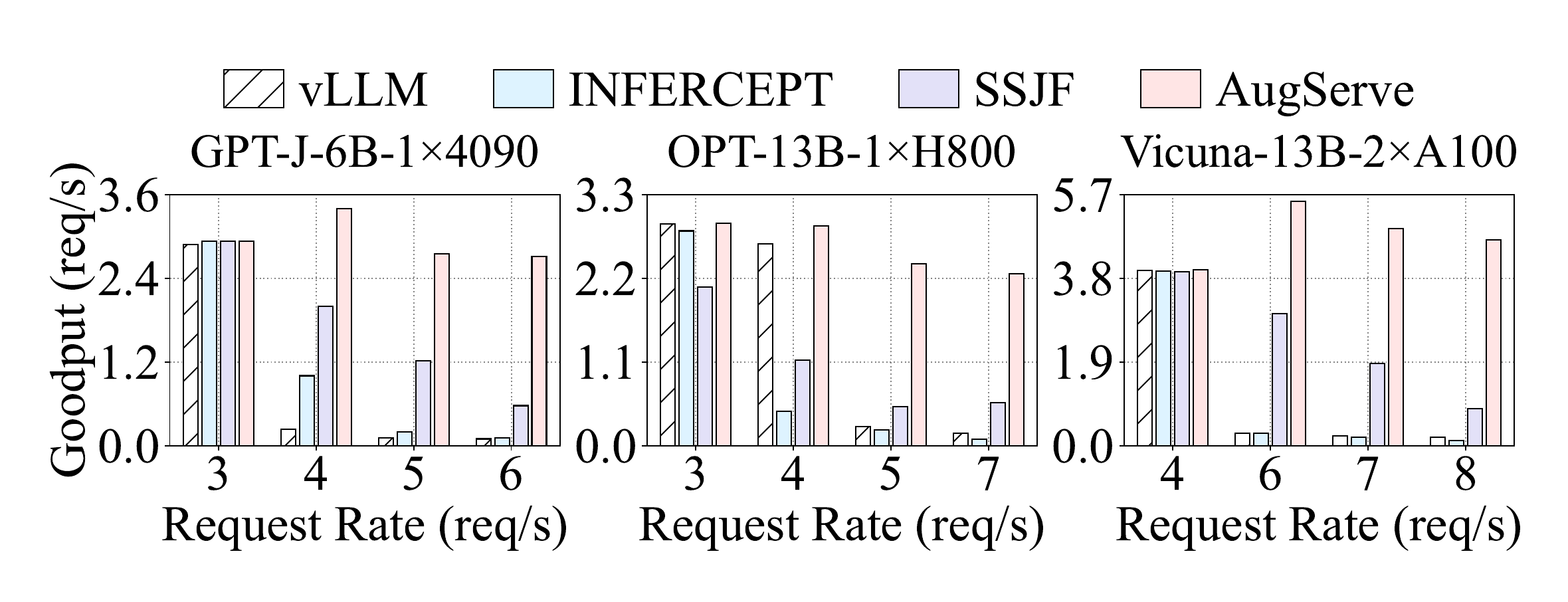}
    \end{minipage}

    % \vspace{1ex} % 行间距
\vskip -10pt
    % --- 第二行 ---
    \begin{minipage}[b]{0.48\textwidth}
        \centering
        \includegraphics[width=\textwidth]{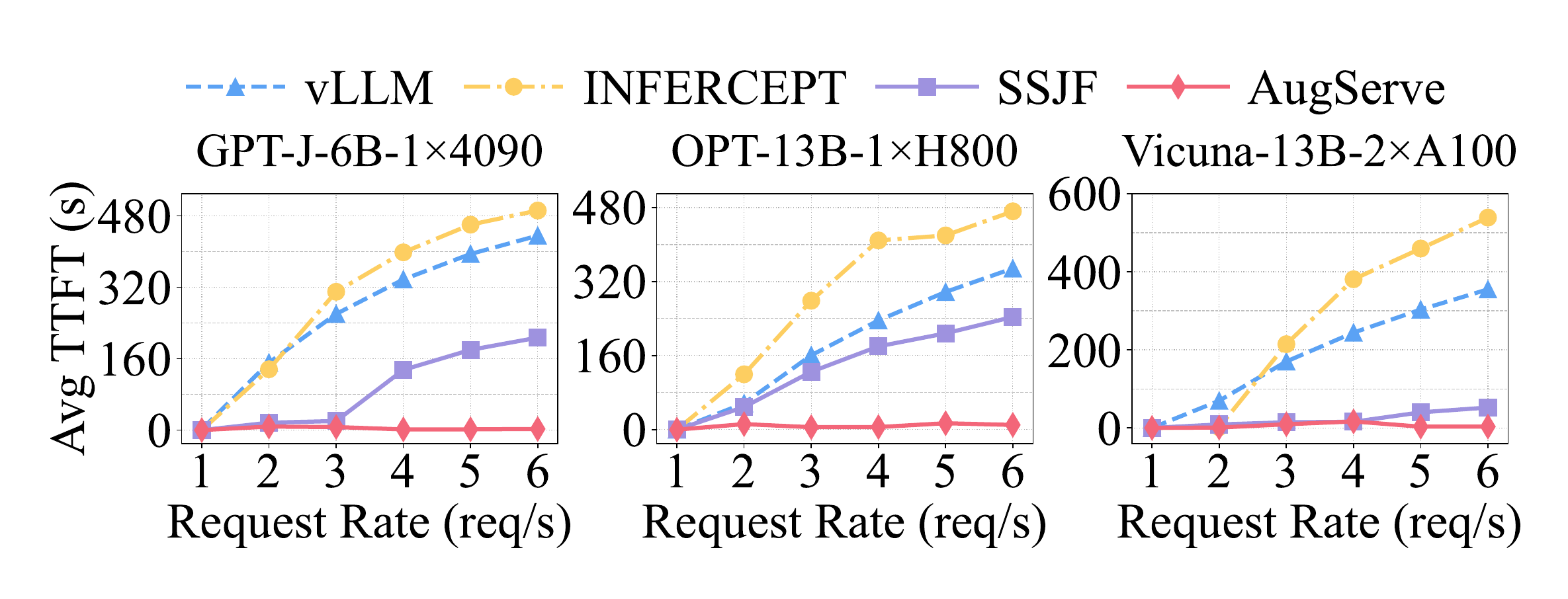}
    \end{minipage}
    \hfill
    \begin{tikzpicture}[baseline={(0,0)}] 
        \draw [dashed, line width=1pt] (0,0) -- (0,3cm);
    \end{tikzpicture}
    \hfill
    \begin{minipage}[b]{0.48\textwidth}
        \centering
        \includegraphics[width=\textwidth]{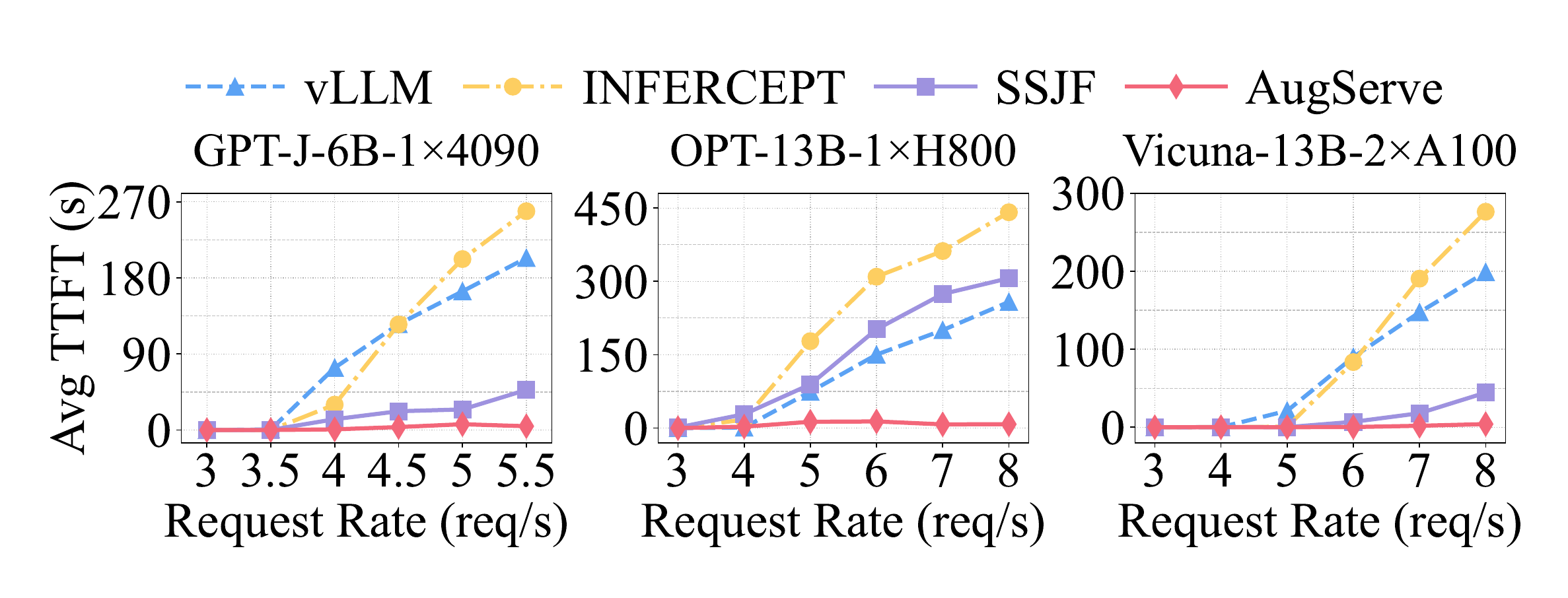}
    \end{minipage}

    % \vspace{1ex} % 行间距
% \vspace{-10pt}
\vskip -10pt
    % --- 第三行 (带 Subcaption) ---
    \begin{minipage}[b]{0.48\textwidth}
        \centering
        \includegraphics[width=\textwidth]{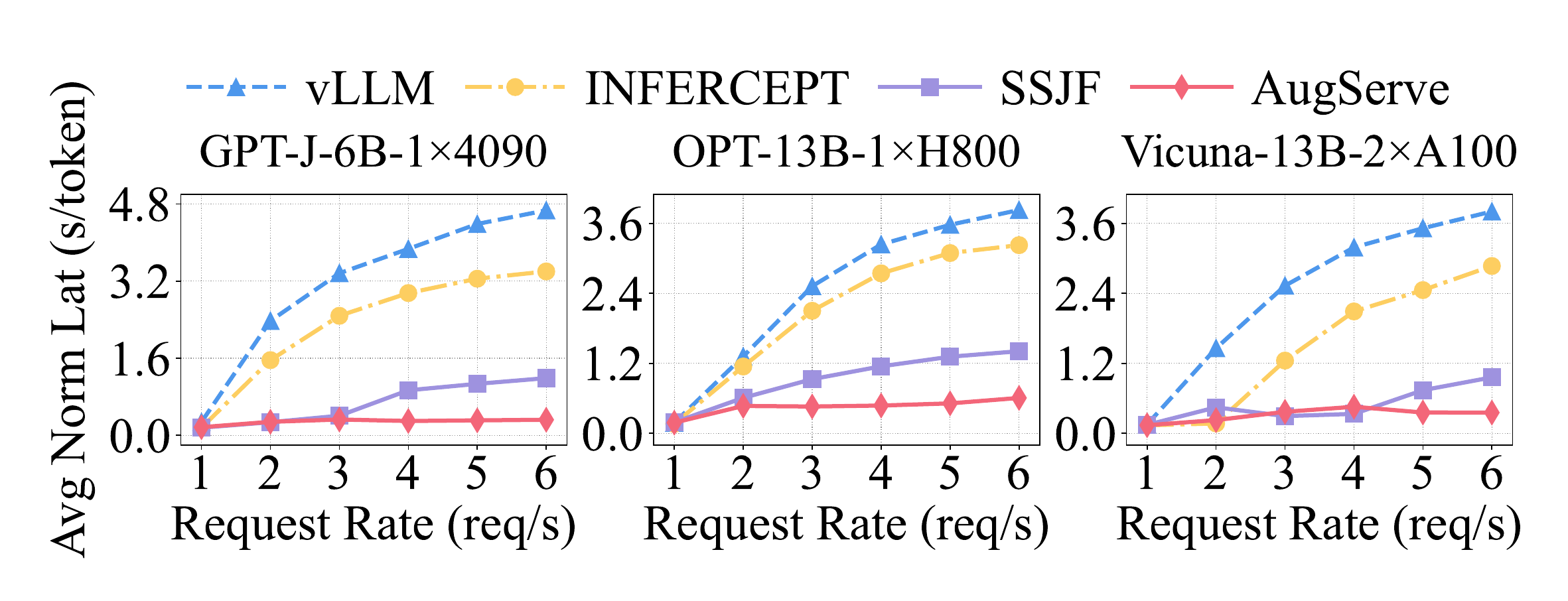}
        % \vspace{2pt}
        \subcaption{\small \textbf{INFERCEPT dataset.}}
    \end{minipage}
    \hfill
    \begin{tikzpicture}[baseline={(0,0)}] % 调整 baseline 确保线在图中间而非 caption 中间
        \draw [dashed, line width=1pt] (0,0) -- (0,3.7cm);
    \end{tikzpicture}
    \hfill
    \begin{minipage}[b]{0.48\textwidth}
        \centering
        \includegraphics[width=\textwidth]{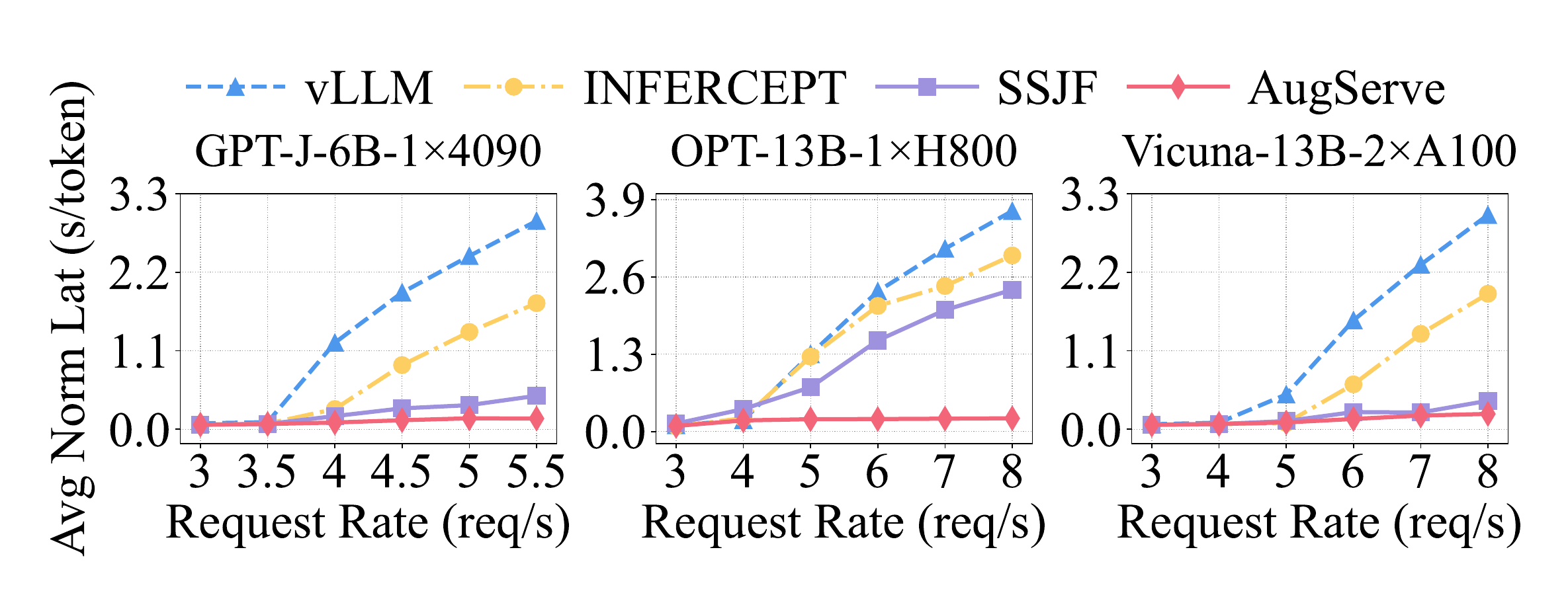}
        % \vspace{2pt}
        \subcaption{\small \textbf{ToolBench dataset.}}
    \end{minipage}

    % --- 总 Caption ---
    % \vspace{5pt}
    \caption{\small Comparison of goodput, TTFT, and normalized latency among vLLM, INFERCEPT, Speculative-SJF scheduling, and AugServe with different models and GPUs on INFERCEPT and ToolBench datasets. First row: Goodput (higher is better). Second row: TTFT (lower is better). Third row: Normalized latency (lower is better).}
    % among different systems on INFERCEPT and ToolBench datasets. 
    \label{fig:e2e}
\end{figure*}

\section{Evaluation}
\label{sec:eval}
We evaluate our approach and baselines across different hardware, models, and datasets. 
\begin{itemize}[leftmargin=*, itemsep=0pt, topsep=-2pt, parsep=0pt, partopsep=0pt]
% \item \textbf{Baselines:} 
\item \textbf{Baselines:}
We compare \name{} against vLLM and INFERCEPT with FCFS scheduling,
and a Speculative Shortest-Job-First (SSJF) baseline representing length-based scheduling using predicted output length.
% We compare \name{} against vLLM, INFERCEPT, and Speculative-SJF (SSJF), which prioritizes requests based on predicted output length.
% \item \textbf{Baselines:} We compare \name{} against vLLM, INFERCEPT, and include Random scheduling and Speculative-SJF (SSJF) as additional baselines. SSJF naively prioritizes requests by predicted output length.
% , illustrating the limitations of length-based scheduling in augmented LLM inference where external API costs are variable.
% SSJF prioritizes requests using predicted output lengths, serving to demonstrate that traditional length-based scheduling is suboptimal in augmented LLM services as it ignores variable external API return lengths.
% Our testbed includes: (i) RTX 4090 (24GB) serving GPT-J-6B, representing memory-constrained consumer hardware; (ii) H800 (80GB) serving OPT-13B for high-throughput single-card scenarios; (iii) 2$\times$A100 (40GB each) with Vicuna-13B to align with prior benchmarks; and (iv) 4$\times$A100 with Llama-3-70B-Instruct to evaluate multi-GPU scalability.
\item \textbf{Setup and models:} 
We run GPT-J-6B on an RTX 4090 GPU (24GB), OPT-13B on an H800 GPU (80GB), 
Vicuna-13B on two A100 GPUs (40GB each), and Llama-3-70B-Instruct on four A100 GPUs.
% We run GPT-J-6B on an RTX 4090 GPU (24GB), OPT-13B on an H800 GPU (80GB).
% Besides, we follow INFERCEPT to run Vicuna-13B on two A100 GPUs (40GB each) and Llama-3-70B-Instruct on four A100 GPUs.
% and, following INFERCEPT, Vicuna-13B on two A100 (40GB each) GPUs and Llama-3-70B-Instruct on four A100 GPUs.
% Our experiments are conducted on diverse hardware platforms to demonstrate the generality of our systems, we run GPT-J-6B on an NVIDIA RTX 4090 GPU and OPT-13B on H800 GPU, to align with INFERCEPT, we also run 13B Vicuna model on two A100 GPUS with 40GB memory each. Otherwise, the Llama-3-70B-Instruct are runned on four A100 GPUS. 
% \item \textbf{Datasets:} To evaluate augmented LLM inference,
\item \textbf{Datasets:} We utilize the INFERCEPT dataset and the ToolBench dataset ~\cite{toolllm}, which features thousands of API calls across various categories.
% To assess the effectiveness of \name{} in augmented LLM inference, we adopt INFERCEPT dataset and ToolBench dataset provided by TollLLM \cite{toolllm}, which contains thousands of API calls spanning dozens of categories. 
\item \textbf{Workloads:} \textbf{(W1)} Poisson arrivals over 30 minutes; 
% \textbf{(W2)} fixed 2k requests for completion efficiency; 
\textbf{(W2)} Gamma arrivals with varying coefficients of variation.
% \item \textbf{Workloads:} \textbf{(W1)} Poisson arrival process over 30 minutes for stochastic request streams; \textbf{(W2)} fixed 2k requests for system completion efficiency; \textbf{(W3)} Gamma arrival for robustness against bursty traffic with varying coefficients of variation.
% \item \textbf{Workloads:} \textbf{(W1)} Poisson arrival process over 30 minutes to emulate stochastic request streams; \textbf{(W2)} fixed 2k requests to measure system completion efficiency; and \textbf{(W3)} a Gamma arrival to test robustness against bursty traffic via varying coefficients of variation.
% We consider three types of workload generation: \textbf{(W1)} a Poisson arrival process, running for 30 minutes to simulate real-world random request streams; \textbf{(W2)} a fixed request count of 2k requests to evaluate system completion efficiency; and \textbf{(W3)} a Gamma arrival process, also running for 30 minutes, where the coefficient of variation (CV) is adjusted to control burstiness under the same average arrival rate, serving as a robustness test.
% We report TTFT (including queuing delay) and normalized latency (end-to-end latency normalized by output length). System efficiency is measured by Goodput, defined as SLO-satisfying requests per unit time.
% the number of SLO-satisfying requests per unit time.
\item \textbf{Metrics and SLOs:} We report Time-to-First-Token (TTFT), normalized latency (end-to-end latency divided by output length), and goodput (number of SLO-satisfying requests per unit time).
% Following prior work~\cite{aptserve, fastserve}, we adopt commonly used SLOs in LLM serving, setting $\text{TTFT} < 1\text{s}$ and normalized latency $< 10\times$ single-iteration time.
Following prior work~\cite{aptserve, fastserve}, SLOs are set as $\text{TTFT} \!\!< \!\!1\text{s}$ and normalized latency $\!\!<\!\! 10\times$ single-iteration time. 
Performance improvements of \name{} over baselines are reported as the geometric mean across all experimental settings.
\end{itemize}

\subsection{End-to-End Performance}
\label{sec:end-to-end}
\textbf{(1) Goodput.}
\autoref{fig:e2e} reports the goodput of \name{} and the baselines across three model configurations and request rates under the W1 workload.
\name{} consistently outperforms all baselines, achieving a geometric mean goodput of $7.5\times$ that of vLLM, $5.7\times$ that of INFERCEPT, and $1.9\times$ that of SSJF.
The performance advantage is particularly pronounced under high contention.
For example, on the H800 GPU with the ToolBench dataset, at a request rate of 5.0~req/s, the goodput of vLLM and INFERCEPT drops to near zero, whereas \name{} sustains a goodput of 2.4~req/s.
% For example, on the H800 GPU with the ToolBench dataset, while the goodput of vLLM and INFERCEPT drops to near zero beyond 5.0~req/s, \name{} sustains a high goodput of 2.4~req/s.
This gap arises because FCFS scheduling in vLLM and INFERCEPT exacerbates HoL blocking under increasing load.
As a result, queuing delays frequently exceed SLOs, sharply reducing SLO attainment (i.e., the fraction of requests meeting SLOs, Appendix \ref{app:slo_rate}) and thereby degrading goodput.
% This gap stems from the fact that FCFS scheduling in vLLM and INFERCEPT exacerbates HoL blocking as load increases.
% The resulting queueing delays frequently exceed SLOs, sharply reducing SLO attainment (i.e., the fraction of requests meeting SLOs, \autoref{app:slo_rate}) and thereby degrading goodput.
% and, in turn, degrading goodput.
% This gap stems from that FCFS scheduling in vLLM and INFERCEPT exacerbates HoL blocking as load increases, causing queueing delays to exceed SLOs and sharply reducing SLO attainment (i.e., the fraction of requests meeting SLOs, \autoref{app:slo_rate}), which directly degrades goodput.
% SSJF partially alleviates HoL blocking by prioritizing requests based on predicted output lengths, but it ignores the execution variability and latency introduced by external calls, leading to suboptimal performance under heavy workloads.
SSJF partially alleviates HoL blocking by prioritizing requests based on predicted output lengths. However, it ignores the execution-stage heterogeneity and latency variability introduced by external calls, leading to suboptimal performance under heavy load.
In contrast, \name{} continuously refines request priorities by incorporating external-call behaviors and cross-round execution states.
This state-aware design effectively mitigates HoL blocking in augmented LLM inference, sustaining high SLO attainment under peak request rates (Appendix \ref{app:slo_rate}) and thereby achieving substantially higher goodput.

\textbf{(2) Latency Performance.}
% \autoref{fig:e2e} compares the TTFT and normalized latency of \name{} with the baselines. These two metrics jointly determine effective throughput, representing request-level responsiveness and token-level execution efficiency, respectively. TTFT captures the queueing delay before the first token is generated.
\autoref{fig:e2e} also compares the TTFT and normalized latency of \name{} with the baselines.
These two metrics jointly influence goodput, capturing request-level responsiveness and token-level execution efficiency, respectively.

TTFT reflects the queuing delay before the first token is generated.
Across all evaluated scenarios, \name{} consistently achieves substantially lower TTFT, with geometric mean reductions of 95.6\%, 96.0\%, and 92.5\% compared to vLLM, INFERCEPT, and SSJF, respectively.
Under heavy load, existing systems suffer from severe HoL blocking caused by requests stalled at external calls,
which prevents the scheduler from prioritizing runnable and resource-efficient executions.
In contrast, \name{} maintains high responsiveness by scheduling requests based on their execution stage and state-dependent resource efficiency.
% Across all evaluated scenarios, \name{} consistently achieves substantially lower TTFT, with geometric mean reductions of 95.6\%, 96.0\%, and 92.5\% compared to vLLM, INFERCEPT, and SSJF, respectively.
% Under heavy loads, existing systems suffer from severe HoL blocking, where requests stalled by external calls delay subsequent executions, whereas \name{} maintains high responsiveness by scheduling requests according to their execution stage and resource efficiency.
% across inference rounds.
% Under heavy loads, baseline systems suffer from severe HoL blocking, where a small number of stalled or externally blocked requests delay the entire queue, whereas \name{} preserves high responsiveness through adaptive priority refinement.
% Across all evaluated scenarios, \name{} consistently achieves substantially lower TTFT, with geometric mean reductions of 94.9\% and 96.0\% compared to vLLM and INFERCEPT, and 93\% of SSJF. 
% Under heavy loads, baseline systems suffer from severe HoL blocking, where a small number of stalled requests delay the entire queue, whereas \name{} preserves high responsiveness through adaptive priority refinement.
Meanwhile, normalized latency captures average token-level execution efficiency while mitigating bias from varying sequence lengths, enabling fair comparison across requests.
Overall, \name{} reduces normalized latency by 77.8\%, 72.8\%, and 36.5\% relative to vLLM, INFERCEPT, and SSJF, respectively.
These gains in both responsiveness and token-level efficiency contribute to \name{}’s superior goodput.
Additional analyses of tail latency (P95) are provided in Appendix~\ref{app:tail}.

\textbf{(3) Additional Results.}
Appendix \ref{appendix:performance} presents additional experimental results that further validate the effectiveness and generality of \name{}.
We first evaluate \name{} under diverse models and workload settings, including mixed external-call and non-external-call workloads (\ref{app:mix}) and Llama-3-70B-Instruct (\ref{app:70b}).
% Then we provide complementary performance analyses, covering robustness to prediction errors (\ref{app:predict}) and bursty arrivals (\ref{app:sens}), as well as comparison with memory-based SJF scheduling (\ref{app:mars}).
Then, we provide complementary analyses of sensitivity to $\beta$ (\ref{app:beta}), robustness to bursty arrivals (\ref{app:sens}), and comparison with memory-based SJF scheduling (\ref{app:mars}).
Finally, we analyze the GPU memory occupancy (\ref{app:memory}).

\subsection{Ablation Study}
\label{sec:ablation}

\begin{figure*}[t]
    % Centering the entire figure
    \centering
    
        \includegraphics[width=\textwidth]{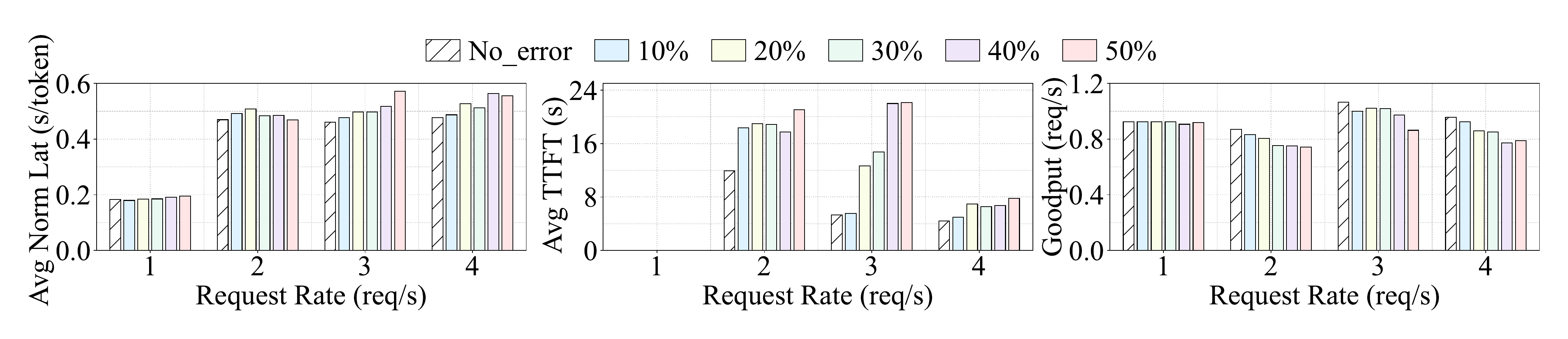}
        % \vspace{-22pt} % Add some space between image and subcaption
        % \caption{Average normalized latency (s/token), TTFT (s), goodput (req/s) comparison with prediction error injection (10\%, 20\%, 30\%, 40\%, 50\%), INFERCEPT dataset with OPT-13B on H800 GPU.} % Subcaption for the left group
        \caption{\small Comparison of normalized latency, TTFT, and goodput under prediction error injection with OPT-13B on an H800 GPU using the INFERCEPT dataset. Prediction errors are injected at 10\%, 20\%, 30\%, 40\%, and 50\%. Left: Normalized latency (lower is better). Middle: TTFT (lower is better). Right: Goodput (higher is better).}
        \label{fig:predict}
\end{figure*}

% 为了更进一步理解\name{}设计的有效性，我们breakdown \name{}的组成，具体来说，我们首先加上两阶段调度策略，比较这部分设计与使用FCFS策略的vLLM和 INFERCEPT的性能差异；接着我们在此基础上再加上动态的token batching，进一步衡量性能改进。我们维持先前的实验设置，使用INFERCEPT和ToolBench两个数据集，OPT-13B以及H800 GPU，在不同负载下进行观察。
% 为进一步理解 \name{} 设计的有效性，我们进行了消融实验，对其核心组件逐步展开分析。具体而言，我们首先加入两阶段调度策略，与采用 FCFS 调度的 vLLM 和INFERCEPT 进行比较；随后在此基础上再引入动态 token batching，以评估进一步的性能改进。实验设置与前文一致，使用 INFERCEPT 和 ToolBench 两个数据集、OPT-13B 模型以及 H800 GPU，并在不同负载下进行测试。

% To further understand the effectiveness of \name{}’s design, we conduct an ablation study by breaking down its core components. Specifically, we first add the two-stage scheduling strategy (\name{} w/ Scheduling in Figure \ref{fig:split}) and compare it against vLLM andINFERCEPT, both of which use FCFS scheduling. We then incorporate dynamic token batching on top of scheduling to assess additional performance improvements. The experimental setup remains the same as before, using the INFERCEPT and ToolBench datasets with OPT-13B on an H800 GPU under different request loads.

% 为进一步理解 \name{} 设计的有效性，我们进行了消融实验，对其核心组件逐步展开分析。具体而言，我们首先加入动态 token batching来测试吞吐量的提升，接着引入两阶段调度策略，与采用 FCFS 调度的 vLLM 和INFERCEPT 进行比较，以评估进一步的性能改进。实验设置使用 INFERCEPT数据集、OPT-13B 模型以及 H800 GPU。

\begin{table}[t]
    \centering
    \caption{Ablation study of \name{}.}
    % \caption{Ablation study of \name{} on the INFERCEPT dataset with OPT-13B on an H800 GPU.}
    \label{tab:ablation}
    \small
    \renewcommand{\arraystretch}{0.8}
    \begin{tabular}{
        >{\centering\arraybackslash}m{0.6cm} |
        >{\centering\arraybackslash}m{1.6cm} |
        >{\centering\arraybackslash}m{2.3cm}
    }
        \toprule
        \textbf{} & \textbf{TTFT (s)} & \textbf{Goodput (req/s)} \\
        % \textbf{} & \textbf{(s)} & \textbf{(req/s)} \\
        \midrule
        Base & 306.26 & 0.15 \\
        +B   & 293.51 & 0.21 \\
        +S   & 14.53  & 0.93 \\
        Aug  & 3.88   & 1.07 \\
        \bottomrule
    \end{tabular}
\end{table}

\begin{figure}[t]
    \centering
    \includegraphics[width=0.83\linewidth,keepaspectratio]{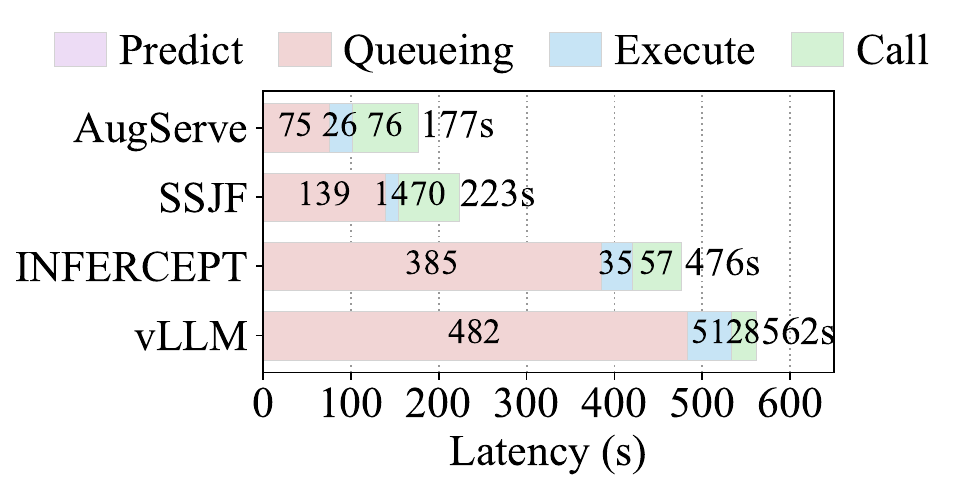}
    \caption{Latency breakdown.}
    % \caption{Latency breakdown of \name{}, vLLM, INFERCEPT, and SSJF on the INFERCEPT dataset with OPT-13B on an H800 GPU.}
    \label{fig:time-breakdown}
\end{figure}

\textbf{(1) Ablation.}
% To isolate the contributions of \name{}'s core designs, we perform an ablation study using the OPT-13B model on an H800 GPU (3.0 req/s, INFERCEPT dataset).
% Figure \ref{fig:ablation} presents the results, highlight the impact of each component on responsiveness and effective throughput.
% We first evaluate the baseline augmented LLM serving system equipped only with \textit{batch-level token budget}. This component optimizes resource allocation by preventing memory fragmentation, yielding a 20\% and 30\% increase in effective throughput over vLLM and INFERCEPT, respectively. 
% We conduct an ablation study to isolate the contributions of \name{}’s two core components, dynamic batch-level token budget and state-aware scheduling, using OPT-13B on an H800 GPU (3.0 req/s, INFERCEPT dataset). \autoref{tab:ablation} summarizes their impact on responsiveness and goodput, Base in \autoref{tab:ablation} denotes INFERCEPT.
We conduct an ablation study to evaluate the contributions of \name{}’s two core components: dynamic batch-level token budget and state-aware scheduling.
Experiments use OPT-13B on an H800 GPU with a 3.0~req/s workload on the INFERCEPT dataset. \autoref{tab:ablation} summarizes their impact on TTFT and goodput, with Base denoting the INFERCEPT baseline.
% We conduct an ablation study to isolate the contributions of \name{}’s core components Token Budget and \name{} Scheduler using OPT-13B on an H800 GPU (3.0 req/s, INFERCEPT dataset). \autoref{fig:ablation} summarizes the impact of each design on responsiveness and effective throughput.

We first enable the dynamic batch-level token budget alone (+B in \autoref{tab:ablation}) on INFERCEPT. By adapting batch capacity to available GPU memory, including reclaimable memory from paused requests, this component improves resource utilization and increases goodput by 41.5\%.
% We first enable dynamic batch-level token budget alone (+B in \autoref{tab:ablation}) on INFERCEPT. By adapting batch capacity to dynamic GPU memory availability, including reclaimable memory from paused requests, this component improves resource utilization and increases goodput by 41.5\%.
% To evaluate the efficiency of dynamic batch-level token budgeting, we enabled this mechanism (denoted as +Batch Budget in \autoref{fig:ablation}) on vLLM and INFERCEPT. By adapting batch capacity to the dynamic memory availability from paused requests, it improves resource utilization and achieves a 38.7\% and 35.7\% increase in goodput over vLLM and INFERCEPT, respectively.
% Next, we isolate state-aware scheduling by replacing FCFS with \name{}’s scheduler on INFERCEPT
Next, we replaced the original FCFS algorithm with AugServe's state-aware scheduling (+S in \autoref{tab:ablation}) while keeping a fixed batch budget. This reduces average TTFT by 95.2\% and boosts goodput by $4.2\times$.
Finally, combining both components in the full \name{} (Aug in \autoref{tab:ablation}) achieves the best overall performance, yielding lower TTFT and higher goodput than either component alone. This confirms that dynamic batch-level capacity adaptation and state-aware scheduling are complementary and jointly essential for efficient augmented LLM inference.
\textbf{(2) Latency Breakdown.}
Figure~\ref{fig:time-breakdown} shows the breakdown of end-to-end request latency for \name{} and the baselines. The results indicate that \name{}’s performance gains are primarily attributed to a substantial reduction in queuing time.
Moreover, the prediction module in \name{} incurs negligible overhead, accounting for 0.1\% of end-to-end latency on average (0.18\,s per request).

\subsection{Robustness to Prediction Errors}
\label{sec:prediction_robustness}

% To assess the robustness of \name{} to prediction inaccuracies, we inject zero-mean Gaussian noise into both \textit{external\_call\_duration} and \textit{output\_length} predictions. 
% Specifically, the injected noise follows $\mathcal{N}(0, (p \times m)^2)$, where $p$ controls the error magnitude and $m$ denotes the measured value. 

To assess the robustness of \name{} to prediction inaccuracies, we inject noise into both \textit{external\_call\_duration} and \textit{output\_length} predictions. 
Specifically, for each prediction $\hat{m}$, we use $\hat{m}(1+\epsilon)$ as the noisy prediction, where $\epsilon$ is randomly chosen from $\{-p, +p\}$ with equal probability.
We vary $p$ from 10\% to 50\% and evaluate OPT-13B on an H800 GPU using the INFERCEPT dataset.

\autoref{fig:predict} reports the impact of prediction errors on normalized latency, TTFT, and goodput. 
Across all error levels, goodput and normalized latency remain largely stable, showing that \name{} is insensitive to moderate-to-large prediction noise. 
TTFT increases mildly as the prediction error grows, with an absolute increase of roughly 10--20\,s even under high error rates of 40--50\%. 
Overall, the performance impact of mispredictions remains limited, demonstrating that \name{} is robust to prediction inaccuracies.
This robustness arises from \name{}’s feedback-driven design, which continuously refines scheduling decisions using realized runtime information upon external call returns, rather than relying solely on static predictions.

\subsection{Scheduling Overhead}
\label{sec:scheduler_overhead}
\begin{table}[t]
\centering
\caption{Average per-iteration scheduling overhead (s) under different request rates on the INFERCEPT dataset with OPT-13B and an H800 GPU.}
\renewcommand{\arraystretch}{0.6}
\begin{tabular}
{>{\centering\arraybackslash}m{2.5cm} |>{\centering\arraybackslash}m{2.0cm} >{\centering\arraybackslash}m{2.0cm}}
\toprule
\textbf{Req/s} & \textbf{INFERCEPT} & \textbf{\name{}} \\
\midrule
3.0 & 0.013 & 0.018 \\
4.0 & 0.015 & 0.022 \\
\bottomrule
\end{tabular}
\label{tab:scheduler_overhead_rate}
\end{table}

Since \name{} performs iteration-level scheduling, we further measure its per-iteration scheduling overhead.
Table~\ref{tab:scheduler_overhead_rate} reports the average scheduling overhead on the INFERCEPT dataset with OPT-13B on an H800 GPU.
\name{} incurs slightly higher overhead than INFERCEPT due to its richer state-aware scheduling logic, but the overhead remains small.
Compared with the substantial queuing-time reduction achieved by \name{}, this additional overhead is negligible.
Additional scalability results under different numbers of paused requests are provided in Appendix~\ref{app:scheduler}.
% \input{content/relatedwork}
% \vspace{-4pt}
\section{Conclusion}
% \name{}的核心是设计一个自适应的请求调度策略以及动态token batching，通过两个阶段的调度以及根据运行时信息的决策修正，高效地适应请求需求以及负载变化，降低整体的排队延迟，有效提升推理服务的有效吞吐量。实验很好地证明了\name{}的高效性和稳定性。
% 本文提出了 \name{} 来优化增强型 LLM 推理服务中的 SLO。\name{} 的核心是一种自适应请求调度策略和动态令牌批处理，通过分析和结合增强型 LLM 推理的独特特性进行两阶段调度决策，以及通过运行时调度修正，实现有效地自适应不同的请求需求和工作负载条件。结合这两点,\name{}降低了整体排队延迟，并显著提高了有效吞吐量。实验证明了 \name{} 的高效性和稳定性。
% This paper proposes \name{} to optimize SLOs in augmented LLM inference serving. The core of \name{} is an adaptive request scheduling strategy and dynamic token batching, designed to address the unique characteristics of augmented LLM inference. By using two-stage scheduling and runtime decision corrections, it efficiently adapts to varying request demands and workload conditions. This approach reduces overall queueing delays and significantly improves effective throughput. Experiments demonstrate the high efficiency and stability of \name{}.

% This paper proposes \name{} to optimize SLOs in augmented LLM inference services. The core of \name{} is an adaptive request scheduling strategy with dynamic token batching, which efficiently adapts to varying request demands and workload conditions through two-stage scheduling decisions and runtime scheduling corrections, tailored to the unique characteristics of augmented LLM inference. This approach reduces overall queueing delays and significantly improves effective throughput. Experiments demonstrate the high efficiency and stability of \name{}.

This paper presents \name{}, a system for improving inference efficiency in augmented LLM services. 
\name{} combines adaptively state-aware request scheduling with a dynamic batch-level token budget to reduce queuing delays and substantially improve effective throughput.

\section*{Acknowledgements}
We would like to thank the anonymous reviewers for their tremendous feedback and comments, which have substantially improved the content and presentation of this paper.

\section*{Impact Statement}
This paper presents work whose goal is to advance the field of machine learning systems. There are many potential societal consequences of our work, none of which we feel must be specifically highlighted here.

\nocite{langley00}
\bibliography{content/paper}

@article{mcpzero,
  title={Mcp-zero: Proactive toolchain construction for llm agents from scratch},
  author={Fei, Xiang and Zheng, Xiawu and Feng, Hao},
  journal={arXiv e-prints},
  pages={arXiv--2506},
  year={2025}
}

@article{nips24advancing,
  title={Advancing tool-augmented large language models: Integrating insights from errors in inference trees},
  author={Chen, Sijia and Wang, Yibo and Wu, Yi-Feng and Chen, Qingguo and Xu, Zhao and Luo, Weihua and Zhang, Kaifu and Zhang, Lijun},
  journal={Advances in Neural Information Processing Systems},
  volume={37},
  pages={106555--106581},
  year={2024}
}

@inproceedings{acl24gear,
    title = "{GEAR}: Augmenting Language Models with Generalizable and Efficient Tool Resolution",
    author = "Lu, Yining  and
      Yu, Haoping  and
      Khashabi, Daniel",
    editor = "Graham, Yvette  and
      Purver, Matthew",
    booktitle = "Proceedings of the 18th Conference of the European Chapter of the Association for Computational Linguistics (Volume 1: Long Papers)",
    month = mar,
    year = "2024",
    address = "St. Julian{'}s, Malta",
    publisher = "Association for Computational Linguistics",
    url = "https://aclanthology.org/2024.eacl-long.7/",
    doi = "10.18653/v1/2024.eacl-long.7",
    pages = "112--138",
}

@article{toolformer,
  title={Toolformer: Language models can teach themselves to use tools},
  author={Schick, Timo and Dwivedi-Yu, Jane and Dess{\`\i}, Roberto and Raileanu, Roberta and Lomeli, Maria and Hambro, Eric and Zettlemoyer, Luke and Cancedda, Nicola and Scialom, Thomas},
  journal={Advances in Neural Information Processing Systems},
  volume={36},
  pages={68539--68551},
  year={2023}
}

@article{toolkengpt,
  title={Toolkengpt: Augmenting frozen language models with massive tools via tool embeddings},
  author={Hao, Shibo and Liu, Tianyang and Wang, Zhen and Hu, Zhiting},
  journal={Advances in neural information processing systems},
  volume={36},
  pages={45870--45894},
  year={2023}
}

@article{nature2025study,
  title={A study on classification based concurrent API calls and optimal model combination for tool augmented LLMs for AI agent},
  author={Go, HeounMo and Park, SangHyun},
  journal={Scientific Reports},
  volume={15},
  number={1},
  pages={20579},
  year={2025},
  publisher={Nature Publishing Group UK London}
}

@article{alm2023survey,
  title={Augmented language models: a survey},
  author={Mialon, Gr{\'e}goire and Dess{\`\i}, Roberto and Lomeli, Maria and Nalmpantis, Christoforos and Pasunuru, Ram and Raileanu, Roberta and Rozi{\`e}re, Baptiste and Schick, Timo and Dwivedi-Yu, Jane and Celikyilmaz, Asli and others},
  journal={arXiv preprint arXiv:2302.07842},
  year={2023}
}

@article{toollearning,
  title={Tool learning with foundation models},
  author={Qin, Yujia and Hu, Shengding and Lin, Yankai and Chen, Weize and Ding, Ning and Cui, Ganqu and Zeng, Zheni and Zhou, Xuanhe and Huang, Yufei and Xiao, Chaojun and others},
  journal={ACM Computing Surveys},
  volume={57},
  number={4},
  pages={1--40},
  year={2024},
  publisher={ACM New York, NY}
}

@inproceedings{
webrl,
title={Web{RL}: Training {LLM} Web Agents via Self-Evolving Online Curriculum Reinforcement Learning},
author={Zehan Qi and Xiao Liu and Iat Long Iong and Hanyu Lai and Xueqiao Sun and Jiadai Sun and Xinyue Yang and Yu Yang and Shuntian Yao and Wei Xu and Jie Tang and Yuxiao Dong},
booktitle={The Thirteenth International Conference on Learning Representations},
year={2025},
url={https://openreview.net/forum?id=oVKEAFjEqv}
}

@inproceedings{webpiot,
author = {Zhang, Yao and Ma, Zijian and Ma, Yunpu and Han, Zhen and Wu, Yu and Tresp, Volker},
title = {WebPilot: a versatile and autonomous multi-agent system for web task execution with strategic exploration},
year = {2025},
isbn = {978-1-57735-897-8},
publisher = {AAAI Press},
url = {https://doi.org/10.1609/aaai.v39i22.34505},
doi = {10.1609/aaai.v39i22.34505},
booktitle = {Proceedings of the Thirty-Ninth AAAI Conference on Artificial Intelligence and Thirty-Seventh Conference on Innovative Applications of Artificial Intelligence and Fifteenth Symposium on Educational Advances in Artificial Intelligence},
articleno = {2607},
numpages = {9},
series = {AAAI'25/IAAI'25/EAAI'25}
}

@article{toolbox,
  title={A Toolbox, Not a Hammer--Multi-TAG: Scaling Math Reasoning with Multi-Tool Aggregation},
  author={Yao, Bohan and Yadav, Vikas},
  journal={arXiv preprint arXiv:2507.18973},
  year={2025}

}

@inproceedings{acl24good,
    title = "From Good to Great: Improving Math Reasoning with Tool-Augmented Interleaf Prompting",
    author = "Chen, Nuo  and
      Li, Hongguang  and
      Wang, Baoyuan  and
      Li, Jia",
    editor = "Dalvi Mishra, Bhavana  and
      Durrett, Greg  and
      Jansen, Peter  and
      Lipkin, Ben  and
      Neves Ribeiro, Danilo  and
      Wong, Lionel  and
      Ye, Xi  and
      Zhao, Wenting",
    booktitle = "Proceedings of the 2nd Workshop on Natural Language Reasoning and Structured Explanations (@ACL 2024)",
    month = aug,
    year = "2024",
    address = "Bangkok, Thailand",
    publisher = "Association for Computational Linguistics",
    url = "https://aclanthology.org/2024.nlrse-1.7/",
    pages = "64--79",
    
}

@article{dragin,
  title={DRAGIN: dynamic retrieval augmented generation based on the information needs of large language models},
  author={Su, Weihang and Tang, Yichen and Ai, Qingyao and Wu, Zhijing and Liu, Yiqun},
  journal={arXiv preprint arXiv:2403.10081},
  year={2024}
}

@INPROCEEDINGS{abouttime,
  author={Gade, Anoushka and Jetcheva, Jorjeta G. and Trivedi, Hardi},
  booktitle={2025 IEEE Conference on Artificial Intelligence (CAI)}, 
  title={It's About Time: Incorporating Temporality in Retrieval Augmented Language Models}, 
  year={2025},
  volume={},
  number={},
  pages={75-82},
  doi={10.1109/CAI64502.2025.00019}}

@article{fastserve,
  title={Fast distributed inference serving for large language models},
  author={Wu, Bingyang and Zhong, Yinmin and Zhang, Zili and Liu, Shengyu and Liu, Fangyue and Sun, Yuanhang and Huang, Gang and Liu, Xuanzhe and Jin, Xin},
  journal={arXiv preprint arXiv:2305.05920},
  year={2023}
}

@article{aptserve,
  title={Apt-Serve: Adaptive Request Scheduling on Hybrid Cache for Scalable LLM Inference Serving},
  author={Gao, Shihong and Zhang, Xin and Shen, Yanyan and Chen, Lei},
  journal={Proceedings of the ACM on Management of Data},
  volume={3},
  number={3},
  pages={1--28},
  year={2025},
  publisher={ACM New York, NY, USA}
}

@inproceedings{vllm,
  title={Efficient memory management for large language model serving with pagedattention},
  author={Kwon, Woosuk and Li, Zhuohan and Zhuang, Siyuan and Sheng, Ying and Zheng, Lianmin and Yu, Cody Hao and Gonzalez, Joseph and Zhang, Hao and Stoica, Ion},
  booktitle={Proceedings of the 29th symposium on operating systems principles},
  pages={611--626},
  year={2023}
}

@inproceedings{infercept,
author = {Abhyankar, Reyna and He, Zijian and Srivatsa, Vikranth and Zhang, Hao and Zhang, Yiying},
title = {INFERCEPT: efficient intercept support for augmented large language model inference},
year = {2024},
publisher = {JMLR.org},
booktitle = {Proceedings of the 41st International Conference on Machine Learning},
articleno = {3},
numpages = {15},
location = {Vienna, Austria},
series = {ICML'24}
}

@article{asynclm,
  title={Asynchronous LLM Function Calling},
  author={Gim, In and Lee, Seung-seob and Zhong, Lin},
  journal={arXiv preprint arXiv:2412.07017},
  year={2024}
}

@misc{mcp,
  author = {Anthropic},
  title = {Model Context Protocol (MCP)},
  year = {2024},
  howpublished = {\url{https://modelcontextprotocol.io}}
}

@inproceedings{orca,
  title={Orca: A distributed serving system for $\{$Transformer-Based$\}$ generative models},
  author={Yu, Gyeong-In and Jeong, Joo Seong and Kim, Geon-Woo and Kim, Soojeong and Chun, Byung-Gon},
  booktitle={16th USENIX Symposium on Operating Systems Design and Implementation (OSDI 22)},
  pages={521--538},
  year={2022}
}

@inproceedings{chunk_prefill,
  title={Taming $\{$Throughput-Latency$\}$ tradeoff in $\{$LLM$\}$ inference with $\{$Sarathi-Serve$\}$},
  author={Agrawal, Amey and Kedia, Nitin and Panwar, Ashish and Mohan, Jayashree and Kwatra, Nipun and Gulavani, Bhargav and Tumanov, Alexey and Ramjee, Ramachandran},
  booktitle={18th USENIX Symposium on Operating Systems Design and Implementation (OSDI 24)},
  pages={117--134},
  year={2024}
}

@inproceedings{distserve,
  title={$\{$DistServe$\}$: Disaggregating prefill and decoding for goodput-optimized large language model serving},
  author={Zhong, Yinmin and Liu, Shengyu and Chen, Junda and Hu, Jianbo and Zhu, Yibo and Liu, Xuanzhe and Jin, Xin and Zhang, Hao},
  booktitle={18th USENIX Symposium on Operating Systems Design and Implementation (OSDI 24)},
  pages={193--210},
  year={2024}
}

@inproceedings{flashgen,
  title={Accelerating LLM Serving for Multi-turn Dialogues with Efficient Resource Management},
  author={Jeong, Jinwoo and Ahn, Jeongseob},
  booktitle={Proceedings of the 30th ACM International Conference on Architectural Support for Programming Languages and Operating Systems, Volume 2},
  pages={1--15},
  year={2025}
}

@article{shuffleinfer,
  title={ShuffleInfer: Disaggregate LLM Inference for Mixed Downstream Workloads},
  author={Hu, CunChen and Huang, HeYang and Xu, LiangLiang and Chen, XuSheng and Wang, Chenxi and Xu, Jiang and Chen, Shuang and Feng, Hao and Wang, Sa and Bao, Yungang and others},
  journal={ACM Transactions on Architecture and Code Optimization},
  year={2025},
  publisher={ACM New York, NY}
}

@article{tightllm,
  title={TightLLM: Maximizing Throughput for LLM Inference via Adaptive Offloading Policy},
  author={Hu, Yitao and Liu, Xiulong and Yang, Guotao and Li, Linxuan and Zeng, Kai and Zhao, Zhixin and Chen, Sheng and Zhao, Laiping and Li, Wenxin and Li, Keqiu},
  journal={IEEE Transactions on Computers},
  year={2025},
  publisher={IEEE}
}

@inproceedings{flexgen,
  title={Flexgen: High-throughput generative inference of large language models with a single gpu},
  author={Sheng, Ying and Zheng, Lianmin and Yuan, Binhang and Li, Zhuohan and Ryabinin, Max and Chen, Beidi and Liang, Percy and R{\'e}, Christopher and Stoica, Ion and Zhang, Ce},
  booktitle={International Conference on Machine Learning},
  pages={31094--31116},
  year={2023},
  organization={PMLR}
}

@inproceedings{toolllm,
title={Tool{LLM}: Facilitating Large Language Models to Master 16000+ Real-world {API}s},
author={Yujia Qin and Shihao Liang and Yining Ye and Kunlun Zhu and Lan Yan and Yaxi Lu and Yankai Lin and Xin Cong and Xiangru Tang and Bill Qian and Sihan Zhao and Lauren Hong and Runchu Tian and Ruobing Xie and Jie Zhou and Mark Gerstein and dahai li and Zhiyuan Liu and Maosong Sun},
booktitle={The Twelfth International Conference on Learning Representations},
year={2024},
url={https://openreview.net/forum?id=dHng2O0Jjr}
}

@article{opt,
  title={Opt: Open pre-trained transformer language models},
  author={Zhang, Susan and Roller, Stephen and Goyal, Naman and Artetxe, Mikel and Chen, Moya and Chen, Shuohui and Dewan, Christopher and Diab, Mona and Li, Xian and Lin, Xi Victoria and others},
  journal={arXiv preprint arXiv:2205.01068},
  year={2022}
}

@inproceedings{s3,
author = {Jin, Yunho and Wu, Chun-Feng and Brooks, David and Wei, Gu-Yeon},
title = {S3: increasing GPU utilization during generative inference for higher throughput},
year = {2023},
publisher = {Curran Associates Inc.},
address = {Red Hook, NY, USA},
articleno = {791},
numpages = {13},
location = {New Orleans, LA, USA},
series = {NIPS '23}
}

@misc{revisiting_slo,
      title={Revisiting Service Level Objectives and System Level Metrics in Large Language Model Serving}, 
      author={Zhibin Wang and Shipeng Li and Yuhang Zhou and Xue Li and Zhonghui Zhang and Nguyen Cam-Tu and Rong Gu and Chen Tian and Guihai Chen and Sheng Zhong},
      year={2025},
      eprint={2410.14257},
      archivePrefix={arXiv},
      primaryClass={cs.LG},
      url={https://arxiv.org/abs/2410.14257}, 
}

@inproceedings{optimizing_goodput,
  title={Optimizing Goodput through Sharing for Batch Analytics with Deadlines.},
  author={Karthik, Srinivas and Sioulas, Panagiotis and Pradhan, Ahana and Subramanya, Raghunandan and Mytilinis, Ioannis and Ailamaki, Anastasia},
  booktitle={EDBT},
  pages={332--344},
  year={2024}
}

@inproceedings{nsdi23_shepherd_goodput,
  title = {{SHEPHERD}: Serving {DNNs} in the Wild},
  author = {Zhang, Hong and Tang, Yupeng and Khandelwal, Anurag and Stoica, Ion},
  booktitle = {20th USENIX Symposium on Networked Systems Design and Implementation (NSDI 23)},
  pages = {787--808},
  year = {2023}
}

@inproceedings{splitwise,
author = {Patel, Pratyush and Choukse, Esha and Zhang, Chaojie and Shah, Aashaka and Goiri, \'{I}\~{n}igo and Maleki, Saeed and Bianchini, Ricardo},
title = {Splitwise: Efficient Generative LLM Inference Using Phase Splitting},
year = {2025},
isbn = {9798350326581},
publisher = {IEEE Press},
url = {https://doi.org/10.1109/ISCA59077.2024.00019},
doi = {10.1109/ISCA59077.2024.00019},
booktitle = {Proceedings of the 51st Annual International Symposium on Computer Architecture},
pages = {118–132},
numpages = {15},
location = {Buenos Aires, Argentina},
series = {ISCA '24}
}

@article{nips24_rank,
  title={Efficient llm scheduling by learning to rank},
  author={Fu, Yichao and Zhu, Siqi and Su, Runlong and Qiao, Aurick and Stoica, Ion and Zhang, Hao},
  journal={Advances in Neural Information Processing Systems},
  volume={37},
  pages={59006--59029},
  year={2024}
}

@inproceedings{
iclr25-dont,
title={{DON}{\textquoteright}T {STOP} {ME} {NOW}: {EMBEDDING} {BASED} {SCHEDULING} {FOR} {LLMS}},
author={Rana Shahout and eran malach and Chunwei Liu and Weifan Jiang and Minlan Yu and Michael Mitzenmacher},
booktitle={The Thirteenth International Conference on Learning Representations},
year={2025},
url={https://openreview.net/forum?id=7JhGdZvW4T}
}

@inproceedings{nips23-response,
author = {Zheng, Zangwei and Ren, Xiaozhe and Xue, Fuzhao and Luo, Yang and Jiang, Xin and You, Yang},
title = {Response length perception and sequence scheduling: an LLM-empowered LLM inference pipeline},
year = {2023},
publisher = {Curran Associates Inc.},
address = {Red Hook, NY, USA},
booktitle = {Proceedings of the 37th International Conference on Neural Information Processing Systems},
articleno = {2859},
numpages = {14},
location = {New Orleans, LA, USA},
series = {NIPS '23}
}

@inproceedings{
lamps,
title={Fast Inference for Augmented Large Language Models},
author={Rana Shahout and Cong Liang and Shiji Xin and Qianru Lao and Yong Cui and Minlan Yu and Michael Mitzenmacher},
booktitle={The Thirty-ninth Annual Conference on Neural Information Processing Systems},
year={2025},
url={https://openreview.net/forum?id=uNqTxj5brQ}
}

@inproceedings{
icml25BFCL,
title={The Berkeley Function Calling Leaderboard ({BFCL}): From Tool Use to Agentic Evaluation of Large Language Models},
author={Shishir G Patil and Huanzhi Mao and Fanjia Yan and Charlie Cheng-Jie Ji and Vishnu Suresh and Ion Stoica and Joseph E. Gonzalez},
booktitle={Forty-second International Conference on Machine Learning},
year={2025},
url={https://openreview.net/forum?id=2GmDdhBdDk}
}

@inproceedings{toolmaker,
    title = "{LLM} Agents Making Agent Tools",
    author = {W{\"o}lflein, Georg  and
      Ferber, Dyke  and
      Truhn, Daniel  and
      Arandjelovic, Ognjen  and
      Kather, Jakob Nikolas},
    editor = "Che, Wanxiang  and
      Nabende, Joyce  and
      Shutova, Ekaterina  and
      Pilehvar, Mohammad Taher",
    booktitle = "Proceedings of the 63rd Annual Meeting of the Association for Computational Linguistics (Volume 1: Long Papers)",
    month = jul,
    year = "2025",
    address = "Vienna, Austria",
    publisher = "Association for Computational Linguistics",
    url = "https://aclanthology.org/2025.acl-long.1266/",
    doi = "10.18653/v1/2025.acl-long.1266",
    pages = "26092--26130",
    ISBN = "979-8-89176-251-0",
}

@misc{batchllm,
      title={BatchLLM: Optimizing Large Batched LLM Inference with Global Prefix Sharing and Throughput-oriented Token Batching}, 
      author={Zhen Zheng and Xin Ji and Taosong Fang and Fanghao Zhou and Chuanjie Liu and Gang Peng},
      year={2025},
      eprint={2412.03594},
      archivePrefix={arXiv},
      primaryClass={cs.CL},
      url={https://arxiv.org/abs/2412.03594}, 
}
\bibliographystyle{icml2026}

%%%%%%%%%%%%%%%%%%%%%%%%%%%%%%%%%%%%%%%%%%%%%%%%%%%%%%%%%%%%%%%%%%%%%%%%%%%%%%%
%%%%%%%%%%%%%%%%%%%%%%%%%%%%%%%%%%%%%%%%%%%%%%%%%%%%%%%%%%%%%%%%%%%%%%%%%%%%%%%
% APPENDIX
%%%%%%%%%%%%%%%%%%%%%%%%%%%%%%%%%%%%%%%%%%%%%%%%%%%%%%%%%%%%%%%%%%%%%%%%%%%%%%%
%%%%%%%%%%%%%%%%%%%%%%%%%%%%%%%%%%%%%%%%%%%%%%%%%%%%%%%%%%%%%%%%%%%%%%%%%%%%%%%
\newpage
\appendix
\onecolumn
% \input{content/appendix/symbol}
% \section{Appendix}
% \section{Greedy Scheduling}
\section{State-Aware Adaptive Scheduling Algorithm}
\label{appendix:symbol}

\begin{table}[t]
\centering
\caption{Notation and definitions of key symbols.}
\label{table:symbol}
\begin{tabular}{l@{\hspace{4pt}}l}
\toprule
Symbol & Definition \\
\midrule
$R_i$ & The $i$-th request in the system \\
$n_i$ & Total number of service segments for request $R_i$ \\
$S_{i,k}$ & The $k$-th service segment of request $R_i$. \\
$P_{i,1}$ & Standard prompt prefill stage in the initial segment. \\
$D_{i,k}$ & Decoding stage within segment $k$. \\
$W_{i,k}$ & Tool-wait stage where the request awaits external responses. \\
$R_{i,k}^{res}(p_i)$ & Context resumption stage, governed by policy $p_i$. \\
$P_{i,k}^{ret}$ & Incremental prefill for processing returned tool tokens. \\
$\Pi$ & Set of context-handling policies: $\{\texttt{Preserve, Swap, Discard}\}$. \\
$V_i^e$ & Scheduling value for request $i$ at iteration $e$. \\
$w_i^e$ & Cumulative waiting time for request $i$ up to iteration $e$. \\
$\beta$ & Starvation prevention aging factor. \\
$x_i^e$ & Binary decision variable: 1 if request $i$ is scheduled, 0 otherwise. \\
$M^e$ & Instantaneous GPU memory capacity constraint at iteration $e$. \\
$m_i(t)$ & Instantaneous memory occupancy of request $i$ at time $t$. \\
$C_{i,k}(p_{i,k})$ & Space-time cost (action) of segment $S_{i,k}$ under policy $p_{i,k}$. \\
$\tau_{i,k}$ & Total residency duration of the request segment. \\
$\theta_{i,k}$ & Scheduling priority defined as the value density $V_i / C_{i,k}$. \\
$\hat{l}_i^{out}$ & Predicted output token length. \\
$\hat{\tau}_i^{call}$ & Predicted duration of the external tool call. \\
$\hat p_{i,k}$ & Predicted context policy during the wait stage. \\
$C_{i,k}^{fb}$ & Feedback-driven realized cost term for $k > 1$. \\
${p}_{i,k-1}$ & Context policy actually executed during the previous wait phase. \\
$\mathcal{B}_{token}(t)$ & Dynamic batch-level token budget at time $t$. \\
$G_{free}(t)$ & Number of currently available free GPU memory blocks. \\
$G_{kv}^{preempt}(t)$ & Reclaimable memory blocks held by paused request contexts. \\
$\mu$ & Memory footprint per token. \\
% $\gamma$ & Reclamation factor for treating paused contexts as elastic capacity. \\
$T_{max}$ & Static offline reference token budget. \\
$\beta_{low}, \beta_{high}$ & Scaling factors for the smoothing range of the token budget. \\

\bottomrule
\end{tabular}
\end{table}

\begin{algorithm}[tp]
  \caption{\name{}: State-Aware Adaptive Scheduling with Dynamic Token Budget}
  \label{alg:augserve}
  \begin{algorithmic}[1]
    \STATE {\bfseries Input:} Request queues: $\mathrm{running}, \mathrm{swapped}, \mathrm{waiting}, \mathrm{paused}$; Predictor $\mathrm{Pred}$; Aging parameter $\beta$; Scaling bounds $\beta_{low}, \beta_{high}$. 
    \vskip 4pt 
    \WHILE{True}

    % \vspace{0.7em}  % 在这里插入一个 0.5 字符高度的空行
    \vskip 4pt
    \STATE \texttt{// (1) Request arrival \& Predictive Initialization}
    \FOR{each $r \in$ arrivals}
    \STATE $(\hat{\tau}_i^{call}, \hat{l}_i^{out}) \leftarrow \mathrm{Pred}(r)$ 
    \STATE $\hat{p}_{i,1} \leftarrow \arg\min_{p \in \Pi} \text{Waste}_i^{p}(\hat{l}_i^{out}, \hat{\tau}_i^{call})$  \quad // Select policy
    \STATE Estimate $\hat{C}_{i,1}$ using $\hat{p}_{i,1}$ and $(\hat{\tau}_i^{call}, \hat{l}_i^{out})$ by \autoref{eq:cost} 
    \STATE $r.\theta_{i,1} \leftarrow (1 + \beta \cdot w_i) / \hat{C}_{i,1}$ by \autoref{eq:value} and \autoref{eq:theta} 
    \STATE $\mathrm{waiting}.\text{push}(r)$
    \ENDFOR

    % \vspace{0.5em}  % 在这里插入一个 0.5 字符高度的空行
    \vskip 4pt
    \STATE \texttt{// (2) State-Aware refinement upon tool return}
    \FOR{each $(r, p_{i,k-1}) \in$ $\mathrm{paused}$}
    \IF{$r.\mathrm{apiFinished}()$}
    \STATE Update $C_{i,k}^{fb}$ using realized policy ${p}_{i,k-1}$ by \autoref{eq:c_fb} 
    \STATE $C_{i,k} \leftarrow C_{i,k}^{fb} + C_{i,k}^{decode} + \hat{C}_{i,k}^{call}$ 
    \IF{${p}_{i,k-1} = \mathrm{Preserve}$} 
    \STATE $\mathrm{running}.\text{push}(r)$
    \ELSIF{${p}_{i,k-1} = \mathrm{Swap}$} 
    \STATE $\mathrm{swapped}.\text{push}(r)$ 
    \ELSE 
    \STATE $\mathrm{waiting}.\text{push}(r)$ \quad // Discard policy
    \ENDIF
    \STATE Update priority $r.\theta_{i,k} \leftarrow (1 + \beta \cdot w_i) / C_{i,k}$ 
    \ENDIF
    \ENDFOR

    % \vspace{0.5em}  % 在这里插入一个 0.5 字符高度的空行
    \vskip 4pt
    \STATE \texttt{// (3) Dynamic token budget adjustment}
    \STATE $\mathcal{B}_{token}(t) \leftarrow \lfloor (G_{free}(t) + G_{\text{kv}}^{\text{preempt}}(t)) / \mu \rfloor$ by \autoref{eq:token} 
    \STATE $\mathcal{B}_{token}(t) \leftarrow \text{clip}(\mathcal{B}_{token}(t), \beta_{low} \cdot T_{max}, \beta_{high} \cdot T_{max})$

    % \vspace{0.5em}  % 在这里插入一个 0.5 字符高度的空行
    \vskip 4pt
    \STATE \texttt{// (4) Global queue ranking \& Greedy packing}
    \STATE Sort all $r$ in candidate queues by $\theta_{i,k}$ 
    \STATE $\mathrm{scheduled} \leftarrow \emptyset$
    \FOR{each $r \in$ sorted candidates}
    \IF{$\mathrm{active\_tokens} + \mathrm{tokens}(r) \le \mathcal{B}_{token}(t)$} 
    \STATE $\mathrm{scheduled} \leftarrow \mathrm{scheduled} \cup \{r\}$ 
    \STATE $\mathrm{active\_tokens} \leftarrow \mathrm{active\_tokens} + \mathrm{tokens}(r)$   
    \STATE Update $r.waiting\_time$
    \ELSE
    \STATE \textbf{break} \quad // Budget reached
    \ENDIF
    \ENDFOR
    
    % \vspace{0.5em}  % 在这里插入一个 0.5 字符高度的空行
    \vskip 4pt
    \STATE \texttt{// (5) Execute one iteration}
    \STATE \textbf{forward}($\mathrm{scheduled}$) 
    \ENDWHILE
  \end{algorithmic}
\end{algorithm}

% \autoref{table:symbol} summarizes the symbols used throughout the paper. \autoref{alg:augserve} presents the overall workflow of \name{}
\autoref{table:symbol} summarizes the symbols used throughout the paper. 
The overall workflow of \name{} is presented in \autoref{alg:augserve}
, which integrates state-aware adaptive scheduling with dynamic batch-level token budget.

Upon request arrival, \name{} invokes a lightweight prediction module to estimate the \textit{external\_call\_duration} and \textit{output\_length}.
These estimates are used to select an initial context-handling policy and to compute the request’s initial scheduling priority based on its expected execution characteristics (Lines~4--10).

When an external tool call completes, \name{} performs state-aware refinement by replacing predictive estimates with realized runtime feedback. 
The scheduling cost and priority are updated based on the actual context-handling policy and return length, and the request is reinserted into the appropriate queue according to its execution state (Lines~12--25).

To respect runtime memory constraints, \name{} dynamically adjusts the batch-level token budget based on available GPU memory and reclaimable memory from paused requests under different context-handling policies.
Bounded smoothing is applied to prevent abrupt budget fluctuations and ensure system stability (Lines~27--28).

Finally, \name{} globally ranks all runnable requests according to their refined scheduling priorities (Line~30).
A batch is then greedily constructed under the current token budget constraint (Lines~32--40) and executed in the next forward iteration.

\section{Context-Handling Policy}
\label{app:policy}
\begin{figure}[t]
    \centering 
    \includegraphics[width=0.4\linewidth]{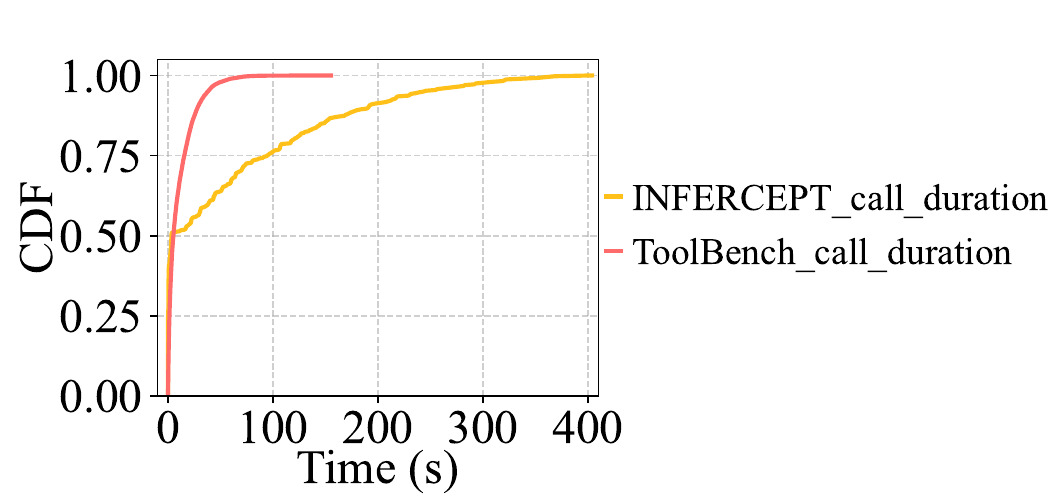}
    % \vspace{-9pt}
    \caption{
    % As the load increases, TTFT rises (driven by queueing latency), causing goodput (effective throughput) to drop.}
    % Effective throughput (req/s) and TTFT (s) with FCFS and SJF scheduling under different request rates.
  % SJF scheduling outperforms FCFS in both goodput (effective throughput) and TTFT, but remains suboptimal under high load.
  External call duration distribution in INFERCEPT and ToolBench datasets.}

    % \vspace{-18pt}
    \label{fig:time}
\end{figure}

The construction of the predictive space-time cost $\hat{C}_{i,k}$ (\S\ref{sec:scheduling}) is inherently tied to the choice of context-handling policy $\hat{\pi}_{i,k}$.
During an external tool call, the resource consumption of a request depends critically on how its KV cache is managed, reflecting a fundamental trade-off between GPU memory occupancy and computational overhead.
Specifically, the \texttt{Preserve} policy avoids recomputation by keeping the context in GPU memory throughout the call, but incurs high memory occupancy,
whereas the \texttt{Discard} policy releases GPU memory at the cost of recomputation when the request resumes.
% As shown in \autoref{fig:time}, external call durations exhibit highly heterogeneous distributions across both datasets.
As shown in \autoref{fig:api_length} and \autoref{fig:time}, external call durations and cumulative context lengths exhibit highly heterogeneous distributions across both datasets.
This variability implies that a fixed context-handling policy is suboptimal:
short external calls with small context favor preserving context to avoid recomputation overhead,
while long calls with large context benefit from releasing GPU memory via discarding or swapping.
% This variability implies that a fixed context-handling policy is suboptimal:
% short external calls favor preserving context to avoid recomputation overhead,
% while long calls benefit from releasing GPU memory via discarding or swapping.
To minimize the total resource overhead induced by external calls,
\name{} selects the most efficient context-handling policy based on the request’s current context length $L_i$,  its predicted \textit{output\_length} $\hat{l}_{i}^{out}$ 
and predicted \textit{external\_call\_duration} $\hat{\tau}_i^{{call}}$ (\S\ref{sec:predict}).
To formalize this optimization, we adopt the memory-waste formulation introduced by INFERCEPT~\cite{infercept},
where the waste metric captures the opportunity cost of GPU memory being occupied or reclaimed during the external call.
% The construction of the predictive space-time cost $\hat{C}_{i,k}$ (\S\ref{sec:scheduling}) is inherently tied to the choice of context handling policy $\hat{\pi}_{i,k}$. 
% During an external tool call, the resource consumption of a request depends heavily on how its KV cache is managed, involving a fundamental trade-off between memory occupancy and computational overhead. \texttt{Preserve} policy avoids recomputation but incurs significant memory occupancy costs as the context remains idle in GPU memory throughout the call duration, while \texttt{Discard} releases memory resources but introduces substantial recomputation overhead when the request resumes execution.
% % As shown in \autoref{fig:time}, external call durations exhibit highly heterogeneous distributions across both datasets. Such variability implies that a fixed context-handling policy is suboptimal, where short calls favor preserving context to avoid recomputation, while long calls benefit from releasing GPU memory via discarding or swapping.
% To minimize the total resource overhead induced by external calls, \name{} must select the most efficient policy based on the request's current context length $L_i$ and its predicted call duration $\hat{\tau}_i^{\mathrm{INT}}$ (\S\ref{sec:predict}).
% To formalize this optimization, we adopt the equations developed by INFERCEPT \cite{infercept} to minimize the memory-waste, where the waste metric captures the opportunity cost of GPU memory occupied or reclaimed during the external call.

For a request $i$, let $M$ denote the per-token memory footprint, 
let
$\tilde{L}_i = L_i + \hat{l}_i^{out}$ denote the estimated context length at the moment the external call is issued,
where $L_i$ is the current context length and $\hat{l}_i^{out}$ is the predicted number of tokens generated before the call.
We denote by $\hat{\tau}_i^{{call}}$ the predicted duration of the external call,
and by $\tau^{{fwd}}(L)$ the execution time of a forward iteration with context length $L$.
We further denote by $L_{{other}}$ the aggregate context length of other runnable requests,
$\tau^{{swap}}(L)$ the time to swap a context of length $L$, and
$N^{{fwd}}_{\max}$ the maximum number of tokens that can be swapped per forward iteration.
% , which determines the effective swap throughput.
% $N^{{fwd}}_{\max}$ the maximum number of forward iterations
% that may be blocked by swapping.

% $L_i$ the context length (in tokens), $\hat{\tau}_i^{{call}}$ denote the duration of the external call, and $\tau^{{fwd}}(L)$ the execution time of a forward iteration with context length $L$.
% We further denote by $L_{{other}}$ the aggregate context length of other runnable requests,
% $\tau^{{swap}}(L)$ the time to swap a context of length $L$, and $N^{{fwd}}_{\max}$ the maximum number of forward iterations
% that may be blocked by swapping.

\paragraph{Preserve.}
Under the \texttt{Preserve} policy, the request retains its context in GPU memory throughout the external call.
The resulting memory waste equals the memory footprint multiplied by the call duration:

% \vspace{-3pt}
\begin{equation}
\small
\text{Waste}^{\text{Preserve}}_i
= \hat{\tau}^{{call}}_i \cdot \tilde{L}_i \cdot M .
\label{eq:waste_preserve}
\end{equation}

\paragraph{Discard.}
With the \texttt{Discard} policy, the context is freed during waiting and recomputed upon resumption.
The waste arises from recomputation overhead, including both the request itself and interference with other active requests:

% \vspace{-8pt}
\begin{equation}
\small
\text{Waste}^{\text{Discard}}_i
=
\tau^{{fwd}}(\tilde{L}_i)\cdot \tilde{L}_i \cdot M
+
\tau^{{fwd}}(\tilde{L}_i)\cdot L_{{other}}\cdot M .
\label{eq:waste_discard}
\end{equation}

\paragraph{Swap.}
Under the \texttt{Swap} policy, the context is temporarily swapped to secondary storage and restored upon resumption.
The waste is dominated by bidirectional swapping overhead that may stall concurrent forward execution:

% \vspace{-8pt}
\begin{equation}
\small
\text{Waste}^{\text{Swap}}_i
=
2 \cdot \tau^{{swap}}(\tilde{L}_i)\cdot N^{{fwd}}_{\max}\cdot M .
\label{eq:waste_swap}
\end{equation}

\paragraph{Policy Selection.}
We select the context-handling policy $\hat{p}_i$ that minimizes the expected memory waste during the external call:
% waiting phase:

% \vspace{-8pt}
\begin{equation}
\small
\hat{p}_i
=
\arg\min_{p \in \{\texttt{Preserve}, \texttt{Discard}, \texttt{Swap}\}}
\text{Waste}^{p}_i .
\label{eq:waste_si}
\end{equation}

While INFERCEPT utilizes these formulas within a First-Come-First-Served (FCFS) scheduling framework, \name{} integrates this waste-aware selection into our state-aware cost model. By quantifying waiting-phase overheads in advance, the scheduler can internalize the downstream memory impact of external calls into its value-density computation. This enables state-aware prioritization that aligns immediate execution decisions with long-term global resource efficiency.

% This waste-aware policy selection directly feeds into the construction of the space--time cost $C_{i,k}$ in the main scheduler.
% By quantifying waiting-phase overheads in advance, the scheduler can internalize the downstream memory impact of external calls
% into its value-density computation, enabling state-aware prioritization that aligns short-term execution decisions with
% long-term resource efficiency.

% \input{content/appendix/cost}
% \input{content/appendix/implementation}

\begin{figure*}[t]
    % Centering the entire figure
    \centering
    
        \includegraphics[width=\textwidth]{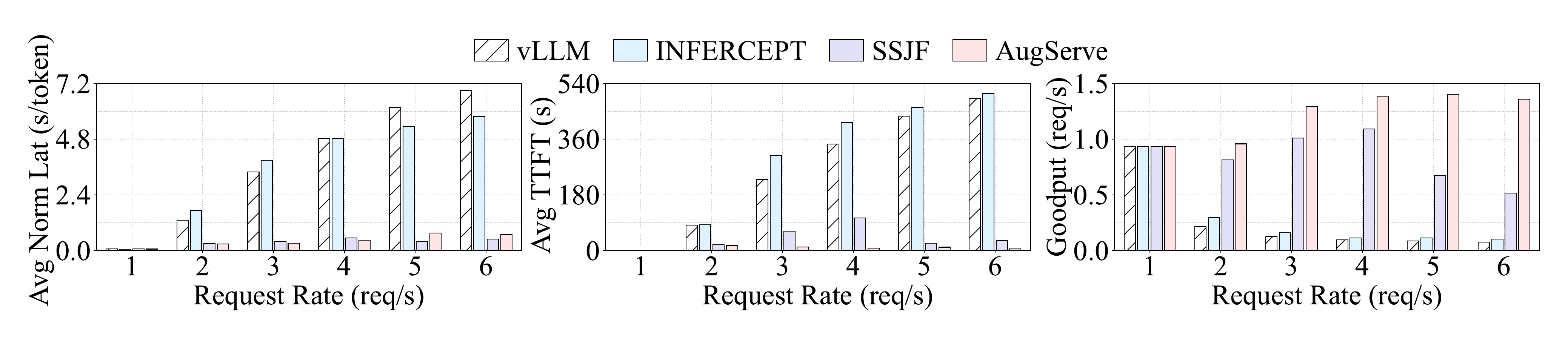}
        % \vspace{-22pt} % Add some space between image and subcaption
        \caption{Average normalized latency (s/token), TTFT (s), goodput (req/s) comparison among vLLM, INFERCEPT, Speculative-SJF scheduling, and \name{} with OPT-13B using mixed workloads of external-call and non-external-call requests on an H800 GPU.} % Subcaption for the left group
        \label{fig:mix}
\end{figure*}

\begin{figure*}[t]
    % Centering the entire figure
    \centering
    
        \includegraphics[width=\textwidth]{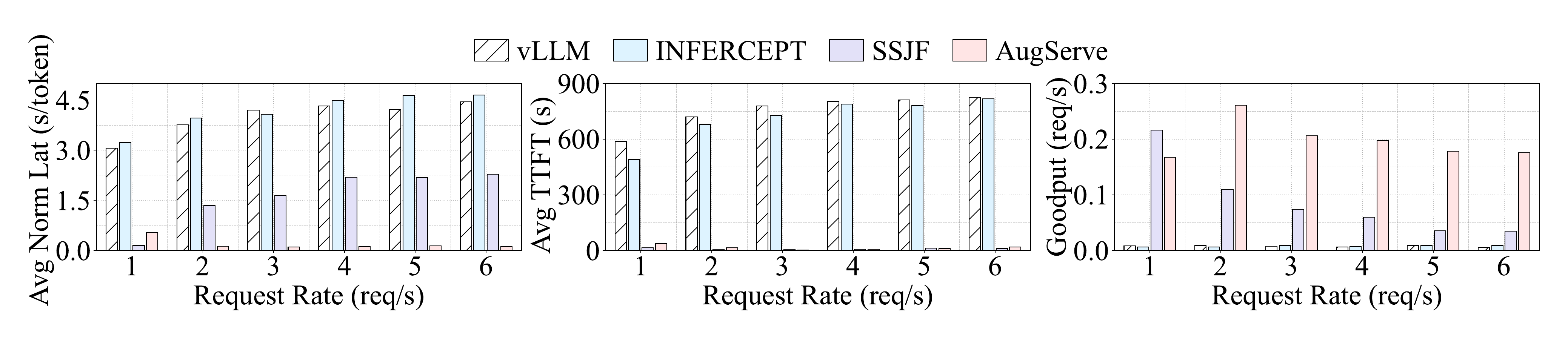}
        % \vspace{-22pt} % Add some space between image and subcaption
        \caption{Average normalized latency (s/token), TTFT (s), goodput (req/s) comparison among vLLM, INFERCEPT, Speculative-SJF scheduling, and \name{} with Llama-3-70B-Instruct using INFERCEPT dataset on 4 A100 GPUs.} % Subcaption for the left group
        \label{fig:70b}
\end{figure*}

% \vspace{-4pt}
\section{Related Work}
% 现在已经又很多努力在优化LLM 推理，这里对几个方面做一些总结
% Existing works optimize LLM inference on different directions.
% There has been substantial effort to optimize LLM inference on different directions.
% , and we summarize related work along several key directions.

% vLLM引入PageAttention的内存管理方式来提高GPU内存利用率，将内存分配为固定大小的blocks并gradually分配，减少内存碎片来降低浪费。InferCept提出增强LLM推理暂停期间的动态上下文处理策略，选择性保留或交换KV cache到CPU内存上，最小化推理暂停期间的内存浪费。FlexGen和FlashGen提出多级KV cache管理，卸载KV到CPU内存或SSD上，释放更多内存处理更多请求，提高吞吐量。Sarathi使用分块预填充将输入分成块在迭代中与解码一起处理，最大化计算资源的使用。这些内存优化工作和\name{}是正交的，可以集成到\name{}中。
% vLLM 引入了 PageAttention 内存管理方式，通过将内存分配为固定大小的块并逐步分配，减少内存碎片和浪费，从而提高 GPU 内存利用率。InferCept 提出了增强 LLM 推理暂停期间的动态上下文处理策略，选择性地保留或将 KV cache 交换到 CPU 内存中，以最小化推理暂停期间的内存浪费。FlexGen 和 FlashGen 提出了多级 KV cache 管理，将 KV 数据卸载到 CPU 内存或 SSD 上，释放更多内存以处理更多请求，从而提高吞吐量。Sarathi 使用分块预填充，将输入拆分为块，与解码一起处理，最大化计算资源的利用。这些内存优化方法与 \name{} 是正交的，可以与 \name{} 集成，进一步提升性能。
\textbf{Memory Optimizations.}
% vLLM \cite{vllm} improves GPU memory utilization with PageAttention, allocating memory in fixed-size blocks and gradually assigning them, reducing fragmentation and memory waste. 
% InferCept \cite{infercept} proposes a dynamic context-handling strategy during augmented LLM inference pauses, selectively swapping KV cache to CPU memory to minimize memory waste. 
% Sarathi \cite{chunk_prefill} employs chunked prefill, splitting the input into blocks and processing them together with decoding to maximize resource utilization.
% FlexGen \cite{flexgen} and FlashGen \cite{flashgen} optimize KV cache management by offloading data to CPU memory or SSDs to boost throughput. 
% These memory optimization techniques, including PageAttention, dynamic context-handling, and chunked prefill, are already integrated into \name{} to enhance performance.
vLLM~\cite{vllm} improves GPU memory utilization with PagedAttention, allocating KV cache in fixed-size blocks to reduce fragmentation.
INFERCEPT~\cite{infercept} proposes dynamic context-handling policies for augmented LLM inference, selectively preserving, swapping, or discarding KV cache during external-call pauses.
Sarathi~\cite{chunk_prefill} employs chunked prefill to interleave prefill and decoding for improved resource utilization.
FlexGen~\cite{flexgen} and FlashGen~\cite{flashgen} further optimize KV cache management by offloading data to CPU memory or SSDs.
These techniques primarily focus on memory efficiency and serve as complementary building blocks for high-performance LLM serving.

% Orca提出迭代级批处理替代原先的以请求为粒度的调度方式，允许提前完成的请求先提出，并允许新到达的请求能立即加入ongoing batch，fully利用GPU并行提高系统吞吐量，迭代级批处理也广泛应用到现有推理框架中。FastServe通过多级反馈队列实现抢占式时间切片机制，降低请求完成时间。还有一些工作专注于通过预测请求输出长度来实现接近最短作业优先的顺序调度，降低整体的排队延迟。虽然在调度上面这些工作做出了努力，然而先前的研究在增强LLM推理场景下缺乏对调用特征的结合，
% 近年来，许多研究在优化LLM推理调度方面取得了显著进展。例如，Orca提出了迭代级批处理方法，允许已完成的请求提前退出，并允许新请求加入正在进行的批处理，充分利用GPU并行计算，从而提高系统吞吐量。FastServe采用多级反馈队列和抢占式时间切片机制，降低请求完成时间。此外，还有一些方法通过预测输出长度来实现接近最短作业优先（SJF）的调度，从而降低整体排队延迟。然而，尽管这些工作在传统LLM推理中取得了一定的进展，它们未能考虑增强LLM推理中特有的调用特征，如API调用时间和推理暂停期间的上下文管理，因此不完全适用于增强LLM场景。
% Recent research has made efforts in optimizing LLM inference scheduling. For instance, Orca \cite{orca} introduces iteration-level batching to improve throughput by fully utilizing GPU parallelism, while FastServe \cite{fastserve} uses preemptive time-slicing to reduce completion times. Some methods \cite{s3, nips24_rank, shuffleinfer,nips23-response,iclr25-dont} also predict output lengths to achieve near-shortest-job-first (SJF) scheduling and reduce queuing delays. However, these approaches fail to address the unique characteristics of augmented LLM inference, such as API call times and return length during pauses, making them less suitable for augmented LLM scenarios.
\textbf{Scheduling.}
Recent research has explored improving request scheduling for LLM inference.
Orca~\cite{orca} leverages iteration-level batching to increase GPU utilization, 
while FastServe~\cite{fastserve} employs input-length–aware, token-level preemptive scheduling to mitigate HoL blocking.
% while FastServe~\cite{fastserve} employs preemptive scheduling based on input length to reduce completion time.
% reduces head-of-line blocking by enabling token-level preemptive scheduling based on input length.
% while FastServe~\cite{fastserve} employs preemptive time-slicing to reduce completion time.
Other approaches~\cite{s3,nips24_rank,shuffleinfer,nips23-response,iclr25-dont} approximate Shortest-Job-First (SJF) scheduling by predicting output lengths to reduce queuing delay.
While effective for text-only inference, these methods do not explicitly account for the heterogeneous and multi-stage execution behavior introduced by external calls in augmented LLM inference.
% Recent work such as MARS~\cite{lamps} extends SJF-style scheduling by predicting memory usage to approximate job size.
% However, it makes one-shot scheduling decisions at each execution round and treats requests after external calls as newly arriving jobs.
% MARS~\cite{lamps} approximates SJF scheduling by predicting memory usage and making one-shot scheduling decisions at each execution round.
% Requests are treated as newly arriving jobs after external calls, without explicitly incorporating realized runtime feedback.
% As a result, it does not explicitly model cross-round execution dynamics or the resumption cost induced by cumulative context growth and external call return variability.
MARS~\cite{lamps} approximates SJF scheduling by predicting memory usage and making scheduling decisions independently at each execution round.
Requests are treated as newly arriving jobs after external calls, without explicitly incorporating realized runtime feedback from prior execution stages.
As a result, it lacks an explicit coupling across rounds to model cross-round execution dynamics or the resumption cost induced by cumulative context growth and variability in external call returns.
In contrast, \name{} introduces a state-aware adaptive scheduling strategy that continuously refines scheduling priorities using runtime feedback upon external call returns.
By explicitly capturing execution state transitions and realized resumption costs, \name{} corrects predictive inaccuracies and adapts to dynamic external environments, enabling more robust scheduling and higher effective throughput in augmented LLM inference.

\section{Additional Experimental Results}
\label{appendix:performance}

\subsection{Mixed Workloads of External-Call and non-External-Call Requests}
\label{app:mix}
\autoref{fig:mix} reports the average normalized latency, TTFT, and goodput for vLLM, INFERCEPT, SSJF, and \name{} under mixed workloads of external-call and non-external-call requests.
We construct the mixed workload from the INFERCEPT dataset by disabling external calls for 50\% of the requests, while preserving the original prompts and output lengths.
All experiments are conducted using OPT-13B on an H800 GPU.
The results show that \name{} consistently achieves the best performance across all metrics.
Under mixed workloads, FCFS-based systems suffer from severe HoL blocking when external-call requests stall execution, which propagates queuing delays to non-external-call requests and degrades overall responsiveness.
SSJF partially mitigates this issue by prioritizing shorter requests, but remains unaware of external-call-induced execution states and thus yields limited gains.
In contrast, \name{} explicitly accounts for heterogeneous execution states across requests, effectively
mitigating the blocking impact of stalled external-call requests on fast non-external-call requests and maintaining high goodput under mixed workloads.

\subsection{Llama-3-70B-Instruct Results}
\label{app:70b}
% We further evaluate \name{} on a large-scale model to assess its scalability. Experiments are conducted using Llama-3-70B-Instruct on four A100 GPUs with the INFERCEPT dataset.

\autoref{fig:70b} compares average normalized latency, TTFT, and goodput across vLLM, INFERCEPT, SSJF scheduling, and \name{} when serving Llama-3-70B-Instruct on four A100 GPUs using the INFERCEPT dataset.
Despite the significantly larger model size and KV cache footprint, \name{} consistently outperforms all baselines across all metrics.
FCFS-based systems suffer from severe queuing delays under external-call-augmented workloads, while SSJF provides only limited improvement due to its lack of awareness of external-call-induced execution states and cross-round resumption dynamics.
In contrast, \name{} maintains low TTFT and high goodput, demonstrating that its state-aware scheduling and dynamic batch-level adaptation remain effective under large-model, multi-GPU inference serving.
% \autoref{fig:70b} compares average normalized latency, TTFT, and goodput across vLLM, INFERCEPT, SSJF scheduling, and \name{} serving with Llama-3-70B-Instruct on four A100 GPUs with the INFERCEPT dataset.
% Despite the significantly larger model size and memory footprint, \name{} consistently outperforms all baselines across all metrics. FCFS-based systems experience severe queueing delays under external-call-augmented workloads, while SSJF provides limited improvement due to its lack of awareness of external-call execution states. In contrast, \name{} maintains low TTFT and high goodput, demonstrating that its state-aware scheduling and dynamic batch adaptation remain effective at scale.

\begin{table}[t]
\centering
\caption{Sensitivity to the balancing parameter $\beta$ on the INFERCEPT dataset with OPT-13B on an H800 GPU.
The default value used in the main experiments is marked with $^\ast$.}
\label{tab:beta_sensitivity}

\begin{tabular}
{>{\centering\arraybackslash}m{2.2cm}|>{\centering\arraybackslash}m{1.1cm} |>{\centering\arraybackslash}m{1.8cm} >{\centering\arraybackslash}m{1.8cm} >{\centering\arraybackslash}m{2.1cm} >{\centering\arraybackslash}m{1.8cm} >
{\centering\arraybackslash}m{1.8cm}}
\toprule
\textbf{Metric} & \textbf{Req/s} & \textbf{$\beta=10^{-3}$} & \textbf{$\beta=10^{-4}$} & \textbf{$\beta=5{\times}10^{-5}$ $^\ast$} & \textbf{$\beta=10^{-5}$} & \textbf{$\beta=10^{-6}$} \\
\midrule
\multirow{2}{*}{Goodput (req/s) }
& 3.0 & 1.01 & 1.12 & 1.07 & 1.04 & 1.02 \\
& 5.0 & 0.82 & 0.94 & 0.90 & 0.94 & 0.86 \\
\midrule
\multirow{2}{*}{P99 TTFT (s)}
& 3.0 & 276.02 & 109.05 & 98.66 & 111.89 & 278.10 \\
& 5.0 & 298.65 & 152.84 & 157.45 & 166.99 & 226.58 \\
\bottomrule
\end{tabular}

\end{table}

\subsection{$\beta$ Sensitivity}
\label{app:beta}

We further evaluate the sensitivity of \name{} to the balancing parameter $\beta$ in \autoref{eq:value}, which balances throughput progress and waiting-time reduction in the scheduling value. 
All experiments are conducted on the INFERCEPT dataset with OPT-13B on an H800 GPU. 
The default value used in the main experiments is $\beta=5\times10^{-5}$.
As shown in \autoref{tab:beta_sensitivity}, \name{} is not overly sensitive to the choice of $\beta$. 
Across different request rates, strong performance is consistently achieved within a stable range around $10^{-5}$--$10^{-4}$, rather than at a single sharply tuned value. 
The default value $\beta=5\times10^{-5}$ falls within this stable range and achieves balanced performance across workloads.
When $\beta$ is too large, e.g., $\beta=10^{-3}$, the scheduler over-emphasizes waiting time and may prioritize long-waiting requests at the cost of execution efficiency, leading to worse tail latency and lower goodput. 
When $\beta$ is too small, e.g., $\beta=10^{-6}$, the scheduler behaves closer to a throughput-dominant greedy policy, which weakens delay awareness and can degrade tail latency. 
Overall, \name{} remains effective over a relatively wide range of $\beta$ values and does not require fine-grained tuning for specific workloads.

\begin{table}[t]
\centering
\caption{Average per-iteration scheduling overhead (s) under different numbers of paused requests.}
\renewcommand{\arraystretch}{0.85}
\begin{tabular}
{>{\centering\arraybackslash}m{2.8cm} |>{\centering\arraybackslash}m{2.0cm} >{\centering\arraybackslash}m{2.0cm}}
\toprule
\textbf{Paused Requests} & \textbf{INFERCEPT} & \textbf{\name{}} \\
\midrule
$\sim$50 & 0.002 & 0.002 \\
50--100 & 0.014 & 0.008 \\
100--150 & 0.013 & 0.016 \\
150--200 & -- & 0.025 \\
$\geq$200 & -- & 0.021 \\
\bottomrule
\end{tabular}
\label{tab:scheduler_overhead_paused}
\end{table}

\subsection{Scheduling Overhead with Increasing Paused Requests}
\label{app:scheduler}
In \S\ref{sec:scheduler_overhead}, we report the average per-iteration scheduling overhead of \name{} under different request rates.
Here, we further examine how this overhead changes as the number of paused requests increases, which is important for augmented LLM workloads where many requests may concurrently wait for external call returns.
As shown in \autoref{tab:scheduler_overhead_paused}, the per-iteration scheduling overhead of \name{} 
increases slightly with the number of paused requests, but remains at the millisecond level even with hundreds of paused requests.
This indicates that iteration-level scheduling does not become a bottleneck under high concurrency.
This behavior is expected, as the scheduling procedure is mainly dominated by sorting and greedy selection over candidate requests, whose cost scales smoothly with the number of active and paused requests.
Overall, \name{} maintains stable and efficient scheduling under high concurrency. 

\begin{table}[t]
\centering
\caption{ Goodput (req/s) comparison of vLLM, INFERCEPT, Speculative-SJF
scheduling, and \name{} under different request rates and arrival burstiness (CV) on INFERCEPT dataset with OPT-13B and an H800 GPU.}
% \vspace{-8pt}
\renewcommand{\arraystretch}{0.85}
\begin{tabular}
% {lccccc} % 修正为 6 列：l + 5c
{>{\centering\arraybackslash}m{3.5cm} |>{\centering\arraybackslash}m{0.5cm} |>{\centering\arraybackslash}m{0.8cm} >{\centering\arraybackslash}m{1.7cm} >{\centering\arraybackslash}m{1.1cm} >{\centering\arraybackslash}m{1.1cm}}
\toprule
 \textbf{Request Rate (req/s)} & \textbf{CV} & \textbf{vLLM} & \textbf{INFERCEPT} & \textbf{SSJF} & \textbf{\name{}} \\
\midrule
 \multirow{3}{*}{2.0} & 1 & 0.21 & 0.28 & 0.49 & 0.85 \\
 & 1.5 & 0.09 & 0.13 & 0.52 & 0.82 \\
 & 2 & 0.06 & 0.08 & 0.39 & 0.78 \\
% \cmidrule{2-6}
 \midrule
\multirow{3}{*}{3.0}  & 1 & 0.14 & 0.20 & 0.50 & 1.03 \\
 & 1.5 & 0.07 & 0.11 & 0.54 & 0.81 \\
 & 2 & 0.04 & 0.06 & 0.23 & 0.42\\

\bottomrule
\end{tabular}
\label{tab:cv}
% \vspace{-10pt}
\end{table}

\subsection{Robustness to Bursty Arrivals}
\label{app:sens}

% 为了测试\name{}与两个baseline 系统在突发流量下的robustness，我们使用伽马分布来模拟请求到达过程，使用变异系数来控制负载波动程度。具体来说，我们在两个数据集下固定不同的request rate，然后设置不同的变异系数，比较几个推理系统的有效吞吐量
% 表6显示了具体的goodput比较，但是在所有的数据集和负载下，\name{}都表现出比vLLM和INFERCEPT更稳定的性能。如表所示，随着突发请求的程度增加，推理系统的效率都随之下降，比如在Merge数据下负载为2.0req/s时，当CV=1.5时，vLLM和INFERCEPT的有效吞吐量都下降到0.1req/s左右，但是\name{}仍然维持1.15req/s，表现出良好的稳定性。
% \subsection{Robustness to Bursty Arrivals.}
To evaluate robustness under bursty traffic, we model request arrivals using a Gamma distribution and control burstiness via the coefficient of variation (CV).
We fix the average request rate and vary CV to induce different levels of load fluctuation, using the INFERCEPT dataset with OPT-13B on an H800 GPU.
\autoref{tab:cv} reports goodput under different burst levels.
Across all evaluated load conditions, \name{} consistently exhibits more stable performance than baselines.
% As burstiness increases, the goodput of FCFS-based systems degrades sharply, while SSJF offers only limited improvement.
As burstiness increases, the goodput of FCFS-based systems degrades sharply, while SSJF offers only limited improvement due to its lack of awareness of external-call-induced execution states.
At 2.0 req/s with CV = 1.5, the goodput of vLLM and INFERCEPT drops to approximately 0.1 req/s, whereas \name{} sustains a throughput of 0.82 req/s.
% As burstiness increases, the goodput of FCFS-based systems degrades sharply.
% For example, a
% At 2.0 req/s with CV = 1.5, the goodput of vLLM and INFERCEPT drops to approximately 0.1 req/s, whereas \name{} sustains a throughput of 0.88 req/s.
This result demonstrates that \name{} is significantly more resilient to bursty arrivals, benefiting from its state-aware scheduling and adaptive capacity control that mitigate burst-induced queue buildup.

\begin{figure*}[t]
    % Centering the entire figure
    \centering
    
        \includegraphics[width=\textwidth]{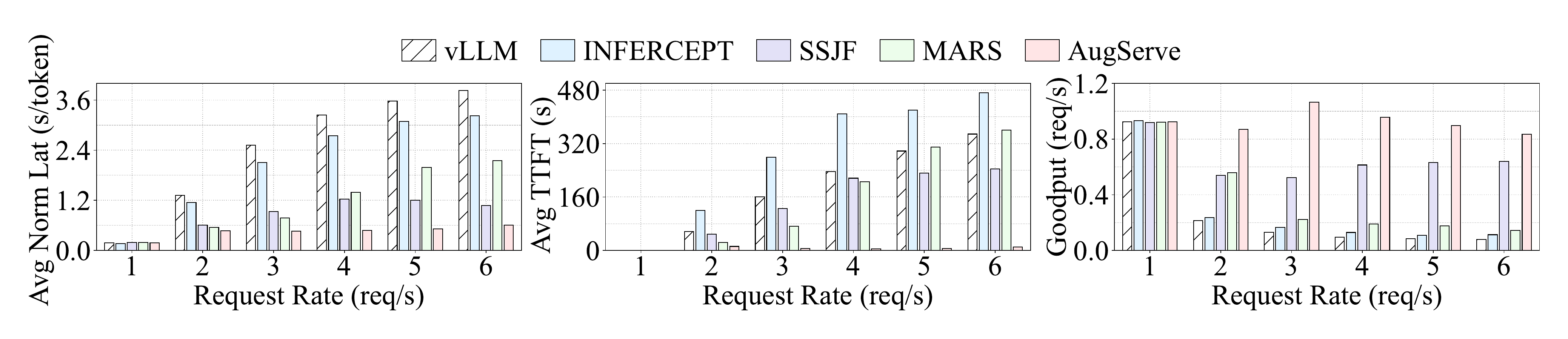}
        % \vspace{-22pt} % Add some space between image and subcaption
        \caption{Average normalized latency (s/token), TTFT (s), goodput (req/s) comparison among vLLM, INFERCEPT, Speculative-SJF scheduling, MARS, and \name{} with OPT-13B using INFERCEPT dataset on an H800 GPU.} % Subcaption for the left group
        \label{fig:mars1}
\end{figure*}

\begin{figure*}[t]
    % Centering the entire figure
    \centering
    
        \includegraphics[width=\textwidth]{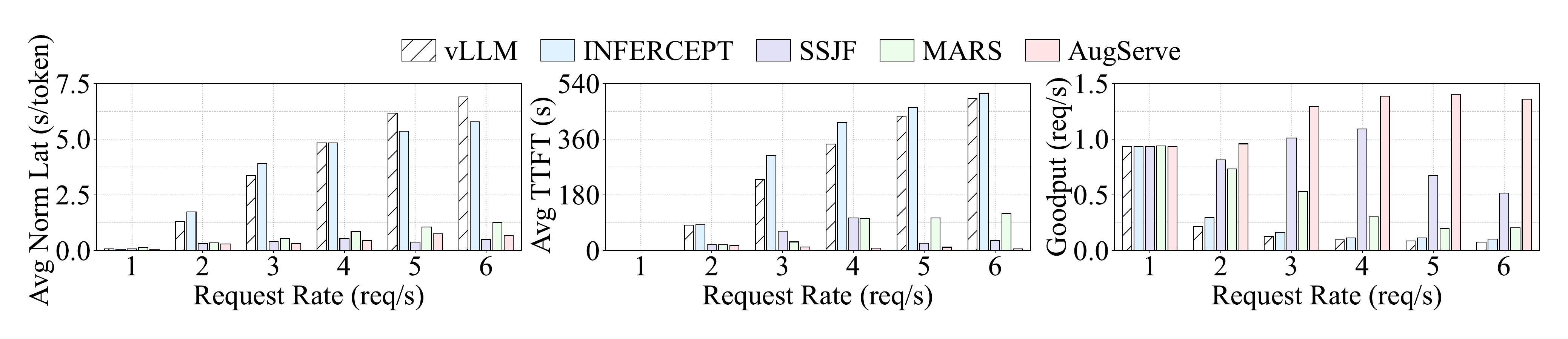}
        % \vspace{-22pt} % Add some space between image and subcaption
        \caption{Average normalized latency (s/token), TTFT (s), goodput (req/s) comparison among vLLM, INFERCEPT, Speculative-SJF scheduling, MARS, and \name{} with OPT-13B using mixed workloads of external-call and non-external-call requests on an H800 GPU.} % Subcaption for the left group
        \label{fig:mars2}
\end{figure*}

\subsection{Comparison with Memory-Based SJF Scheduling}
\label{app:mars}
% We additionally compare \name{} against a memory-based approximate SJF scheduling baseline MARS.
% This class of methods estimates job sizes based on memory-time cost and prioritizes requests accordingly.
We further compare \name{} with MARS, a representative memory-based SJF scheduler for augmented LLM inference, which prioritizes requests based on estimated memory cost.
Following MARS, each service round is treated as an independent scheduling unit with static, per-call cost estimation, without explicitly modeling cross-round execution state.
Since MARS does not provide a prediction model for the INFERCEPT workload, we re-implement its scheduling policy and priority formulation within our system and evaluate it under the same prediction interface as other baselines to ensure a fair comparison.
% We further compare \name{} with MARS, a representative memory--based SJF scheduler designed for augmented LLM inference. This class of methods estimates job sizes based on memory cost and prioritizes requests accordingly.
% As MARS does not provide a prediction model for the INFERCEPT workload and relies on oracle execution statistics, we re-implement its scheduling policy and priority formulation within our system, and evaluate it under the same predicted interface as other baselines for fair comparison.
% Since MARS does not provide a prediction model for the INFERCEPT workload and directly uses oracle execution statistics, we reuse the scheduling policy and priority formulation of MARS, and evaluate it under a unified prediction interface with other baselines for a fair comparison.
% Specifically, we reuse the scheduling policy and priority formulation of MARS, while adopting a unified prediction interface for API latency and return length across all schedulers.
% Following MARS, each service round is treated as an independent scheduling unit with static, per-API cost estimation.
% Following MARS, each service round is treated as an independent scheduling unit with static, per-API cost estimation, without modeling cross-round execution state.
Consistent with prior observations, memory-based SJF improves over FCFS-style baselines under light to moderate load, reducing TTFT and  normalized latency while maintaining higher goodput.
However, as load increases and requests exhibit larger cumulative context growth and higher variability in external call returns, its per-round static cost abstraction becomes increasingly brittle, leading to degraded performance under heavy load.
In some scenarios, its performance can approach or even fall below simpler length-based SJF heuristics.
In contrast, \name{} explicitly models execution-stage-dependent state and dynamically refines scheduling decisions across service rounds, enabling more stable performance across load levels.
As shown in \autoref{fig:mars1} and \autoref{fig:mars2}, \name{} achieves a geometric mean goodput of $3.07\times$ that of MARS across the INFERCEPT and mixed datasets, and up to $4.78\times$ higher under heavy load.
% As shown in \autoref{fig:mars1} and \autoref{fig:mars2}, \name{} achieves a geometric mean goodput of $3.07\times$ that of MARS, and up to $4.78\times$ higher under heavy load on the INFERCEPT dataset.
It also reduces TTFT by 92.1\% and normalized latency by 38.9\%, demonstrating both higher efficiency and robustness.

\begin{figure*}[t]
    % Centering the entire figure
    \centering
    % Using minipage to split the figure into two parts
    \begin{minipage}[b]{0.492\textwidth}
        \centering
        \includegraphics[width=\textwidth]{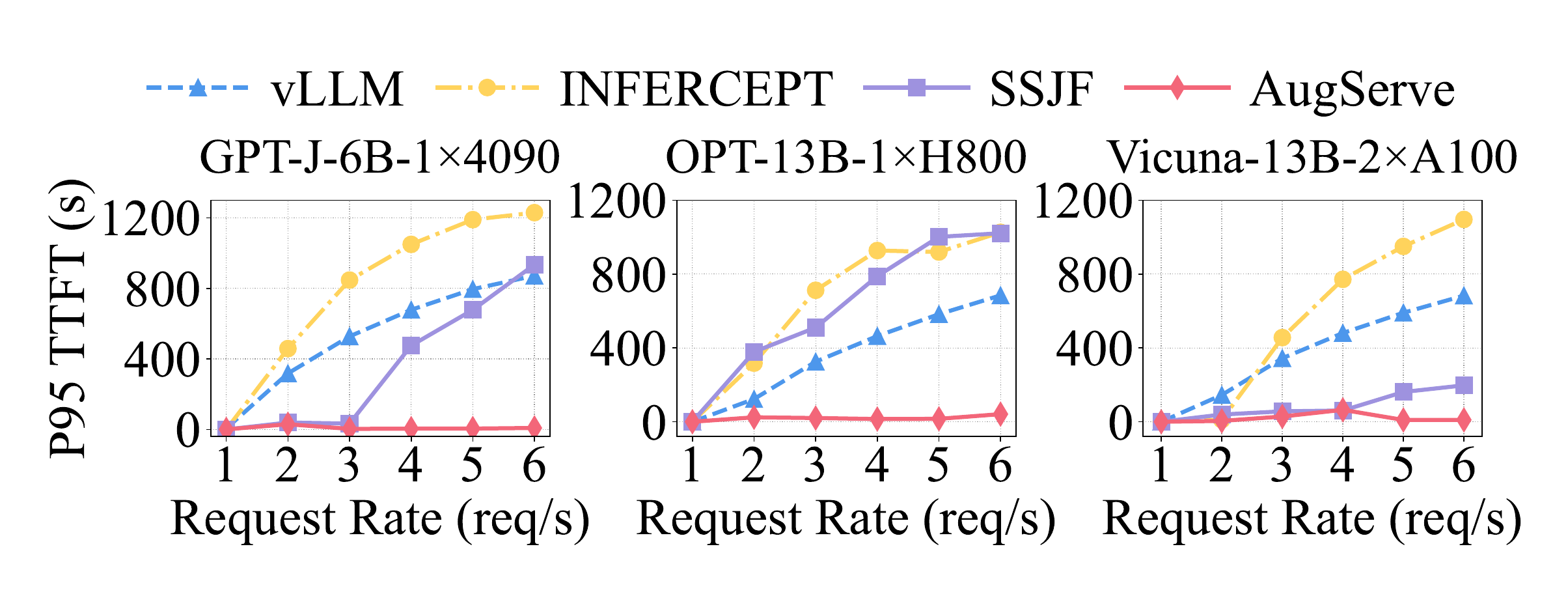}
        % \vspace{-22pt} % Add some space between image and subcaption
        \subcaption{ \textbf{INFERCEPT dataset.}} % Subcaption for the left group
    \end{minipage}
    % \hfill % Add space between minipages
    % % Draw a vertical dashed line using tikz
\begin{tikzpicture}[baseline={(0,-0.22cm)}] % Adjust -5cm to match image height
        \draw [dashed, line width=1pt] (0,0) -- (0,2.7cm); % Height matches image
    \end{tikzpicture}
    % \hfill % Add space between the line and the right minipage
    \begin{minipage}[b]{0.492\textwidth}
        \centering
        \includegraphics[width=\textwidth]{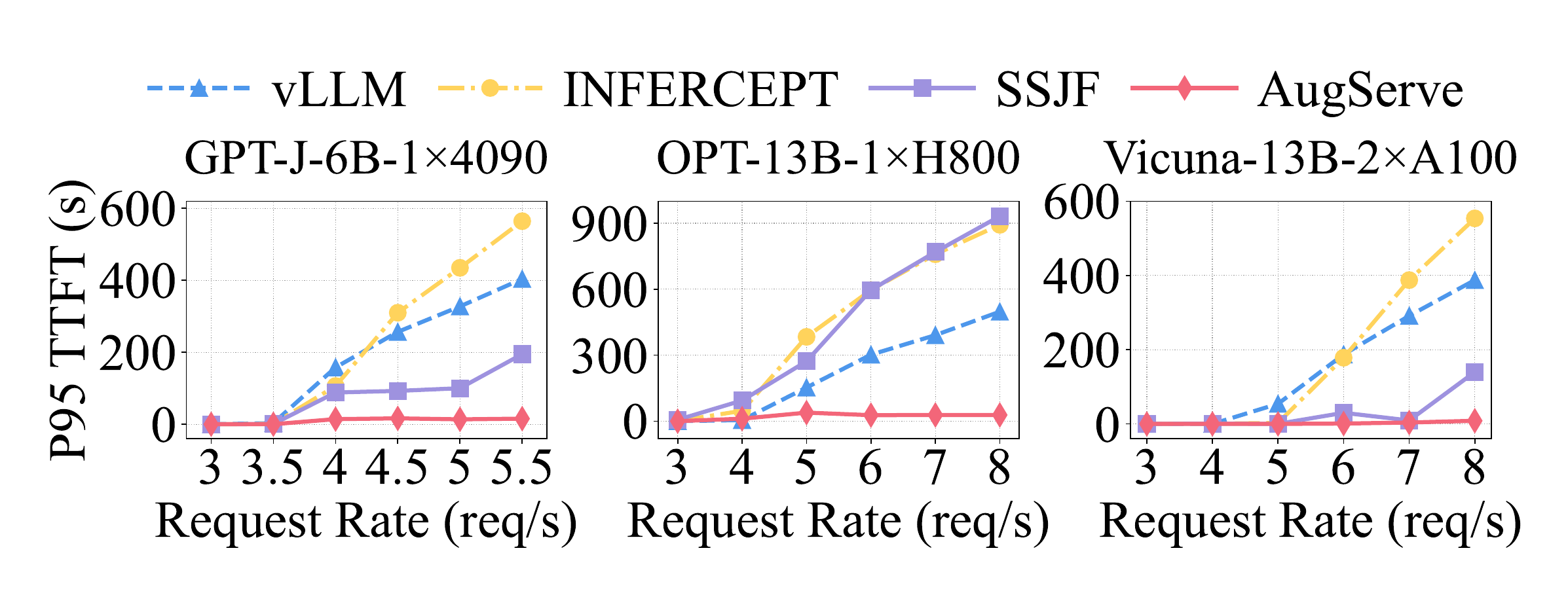} % Adjust path if needed
        % \vspace{-22pt} % Add some space between image and subcaption
        \subcaption{\small \textbf{ToolBench dataset.}} % Subcaption for the right group
    \end{minipage}
    % Adding a caption for the entire figure
    % \vspace{-12pt}
    \caption{P95 Time-to-First-Token (TTFT) (s) comparison among vLLM, INFERCEPT, Speculative-SJF scheduling, and \name{} on INFERCEPT and ToolBench datasets with different models and GPUs. Lower right is better, i.e., shorter response time and queuing time.}
    % Adding a label for referencing
    \label{fig:slo_p95ttft}
    % \vspace{-8pt}
\end{figure*}

\begin{figure*}[t]
    % Centering the entire figure
    \centering
    % Using minipage to split the figure into two parts
    \begin{minipage}[b]{0.492\textwidth}
        \centering
        \includegraphics[width=\textwidth]{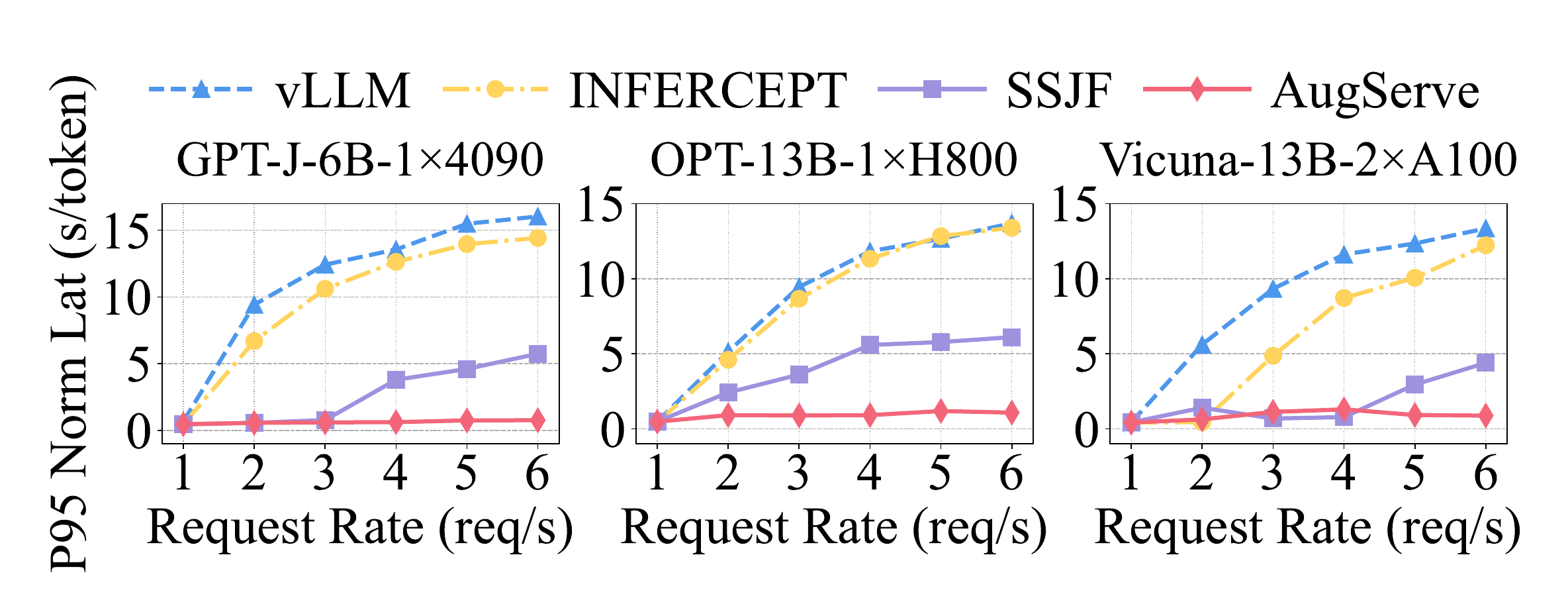}
        % \vspace{-22pt} % Add some space between image and subcaption
        \subcaption{\textbf{INFERCEPT dataset.}} % Subcaption for the left group
    \end{minipage}
    % \hfill % Add space between minipages
    % % Draw a vertical dashed line using tikz
\begin{tikzpicture}[baseline={(0,-0.22cm)}] % Adjust -5cm to match image height
        \draw [dashed, line width=1pt] (0,0) -- (0,2.7cm); % Height matches image
    \end{tikzpicture}
    % \hfill % Add space between the line and the right minipage
    \begin{minipage}[b]{0.492\textwidth}
        \centering
        \includegraphics[width=\textwidth]{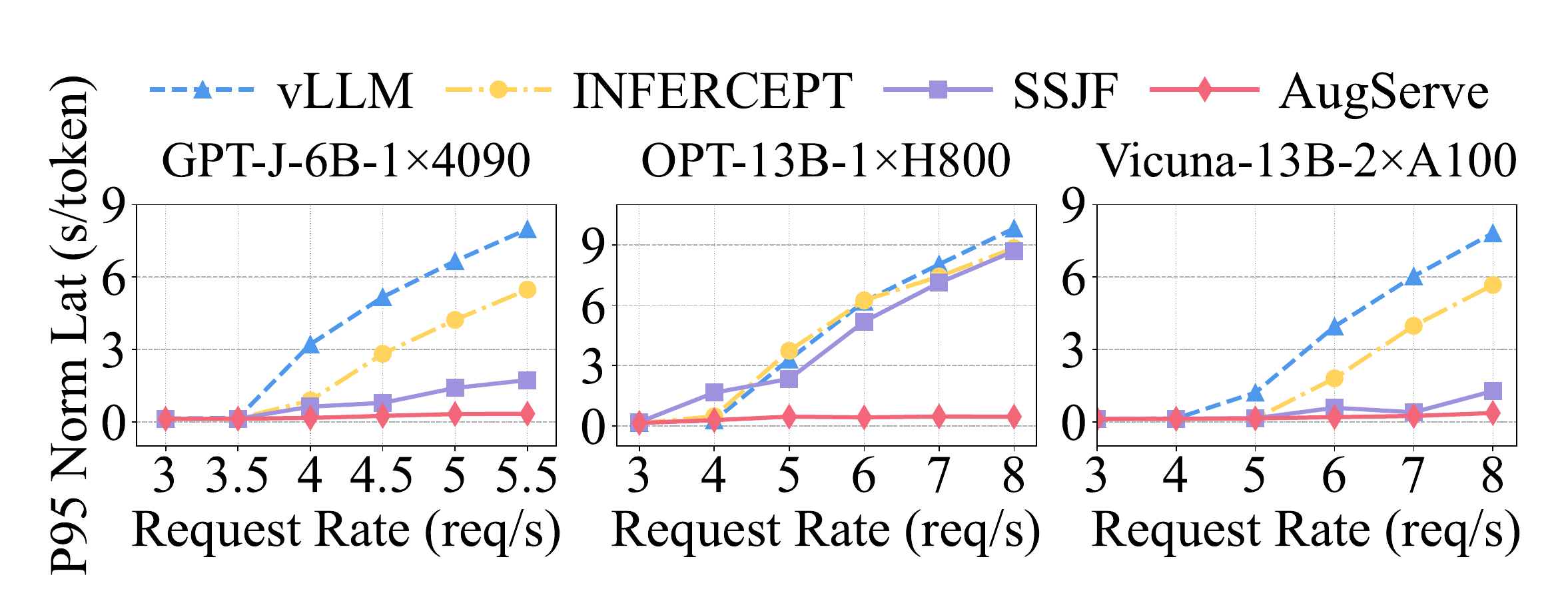} % Adjust path if needed
        % \vspace{-22pt} % Add some space between image and subcaption
        \subcaption{\textbf{ToolBench dataset.}} % Subcaption for the right group
    \end{minipage}
    % Adding a caption for the entire figure
    % \vspace{-12pt}
    \caption{P95 normalized Latency (s/token) comparison among vLLM, INFERCEPT, Speculative-SJF scheduling, and \name{} on INFERCEPT and ToolBench datasets with different models and GPUs. Lower right is better, i.e., sustains higher serving load.}
    % Adding a label for referencing
    \label{fig:p95_norm}
    % \vspace{-8pt}
\end{figure*}
\subsection{Tail Latency Performance}
\label{app:tail}
% 为了证明系统延迟性能的稳定性和用户体验，我们进一步分析了系统在不同负载下的P95 平均 Token 延迟 和 P95 TTFT性能，实验设置与\S\ref{sec:end-to-end}保持一致，结果在\ref{}和\ref{}展示。具体来说，vLLM和INFERCEPT的尾延迟都会随着负载增加而逐渐升高，这也说明了二者在高负载时系统中排队延迟和资源争用情况很严重，从而导致性能下降。但\name{}仍能维持很低的尾部延迟以及良好的稳定性，提供更好的用户体验。

% 为了证明系统延迟性能的稳定性和用户体验，我们进一步分析了系统在不同负载下的 P95 平均 Token 延迟和 P95 TTFT 性能。实验设置与\S\ref{sec:end-to-end}保持一致，结果展示在图 \ref{} 和 \ref{} 中。具体来说，vLLM 和 INFERCEPT 的尾延迟随着负载增加逐渐升高，这也表明它们在高负载时系统中排队延迟和资源争用问题较为严重，从而导致性能下降。相比之下，\name{} 仍能维持较低的尾部延迟和良好的稳定性，提供更优的用户体验。
\autoref{fig:slo_p95ttft} and \autoref{fig:p95_norm} report the P95 TTFT and P95 normalized latency under different load levels, using the same experimental setup as in \S\ref{sec:end-to-end}.
As load increases, the tail latency of vLLM and INFERCEPT grows sharply, reflecting severe queuing delays and resource contention under high load.
SSJF provides limited improvement by prioritizing shorter requests, but still exhibits high tail latency when queuing delay is dominated by external-call-induced stalls.
In contrast, \name{} consistently maintains low P95 TTFT and normalized latency across load levels, demonstrating robust tail latency control under external-call-augmented workloads.

\begin{figure*}[t]
    % Centering the entire figure
    \centering
    % Using minipage to split the figure into two parts
    \begin{minipage}[b]{0.492\textwidth}
        \centering
        \includegraphics[width=\textwidth]{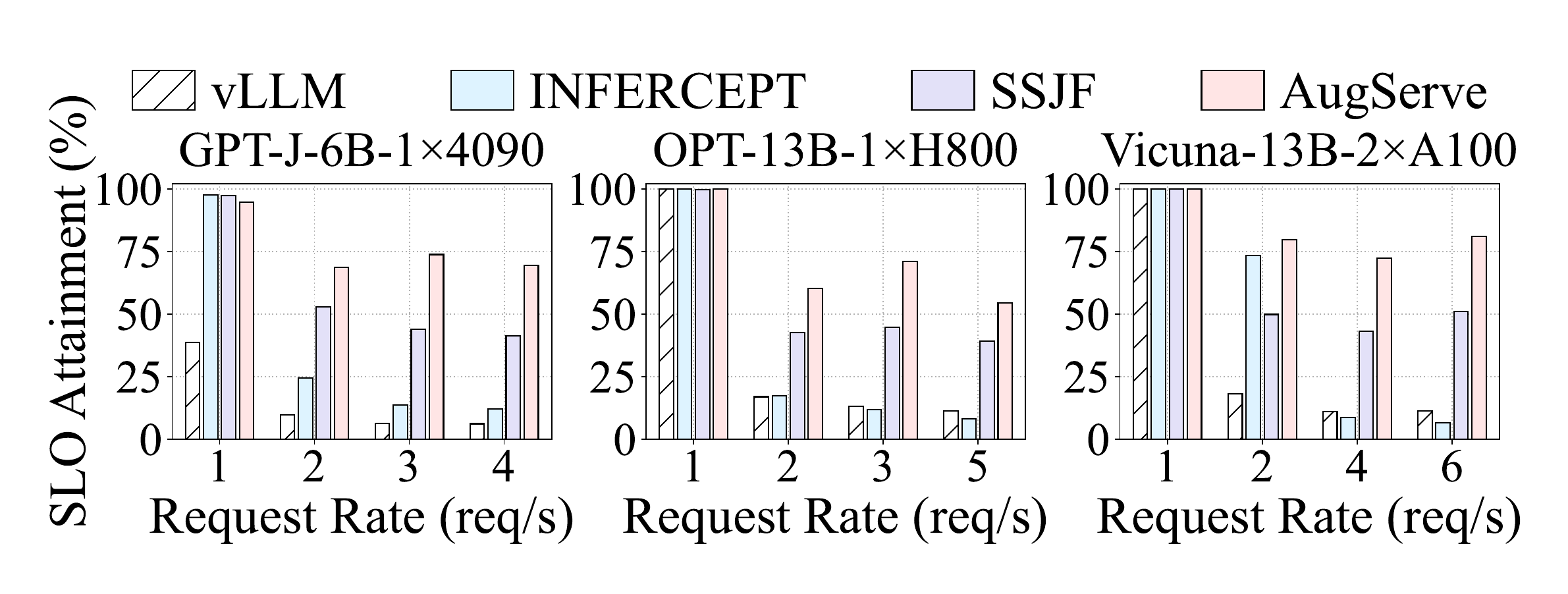}
        % \vspace{-22pt} % Add some space between image and subcaption
        \subcaption{ \textbf{INFERCEPT dataset.}} % Subcaption for the left group
    \end{minipage}
    % \hfill % Add space between minipages
    % % Draw a vertical dashed line using tikz
\begin{tikzpicture}[baseline={(0,-0.22cm)}] % Adjust -5cm to match image height
        \draw [dashed, line width=1pt] (0,0) -- (0,2.7cm); % Height matches image
    \end{tikzpicture}
    % \hfill % Add space between the line and the right minipage
    \begin{minipage}[b]{0.492\textwidth}
        \centering
        \includegraphics[width=\textwidth]{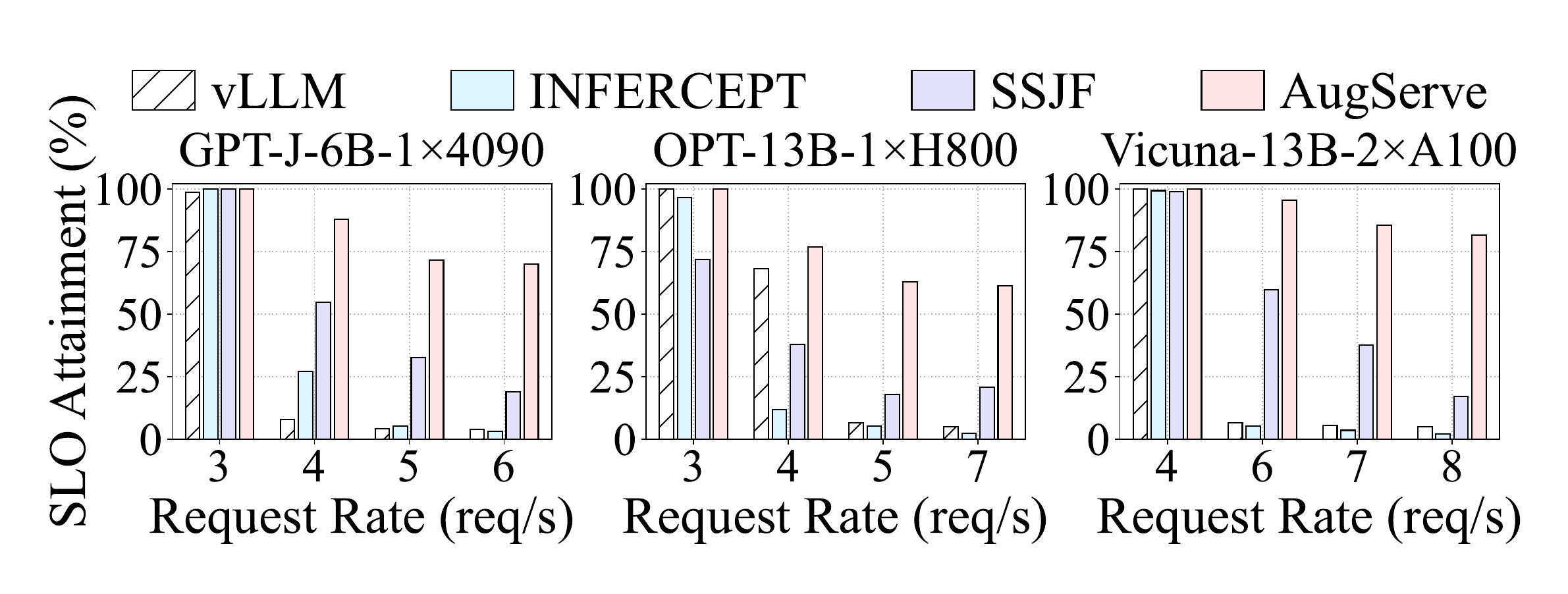} % Adjust path if needed
        % \vspace{-22pt} % Add some space between image and subcaption
        \subcaption{ \textbf{ToolBench dataset.}} % Subcaption for the right group
    \end{minipage}
    % Adding a caption for the entire figure
    % \vspace{-12pt}
    \caption{ SLO attainment (\%) with SLOs comparison among vLLM, INFERCEPT, Speculative-SJF scheduling, and \name{} on INFERCEPT and ToolBench datasets with different models and GPUs. Higher is better.}
    % Adding a label for referencing
    \label{fig:slo_attainment}
    % \vspace{-8pt}
\end{figure*}

\begin{figure*}[t]
    % Centering the entire figure
    \centering
    
        \includegraphics[width=\textwidth]{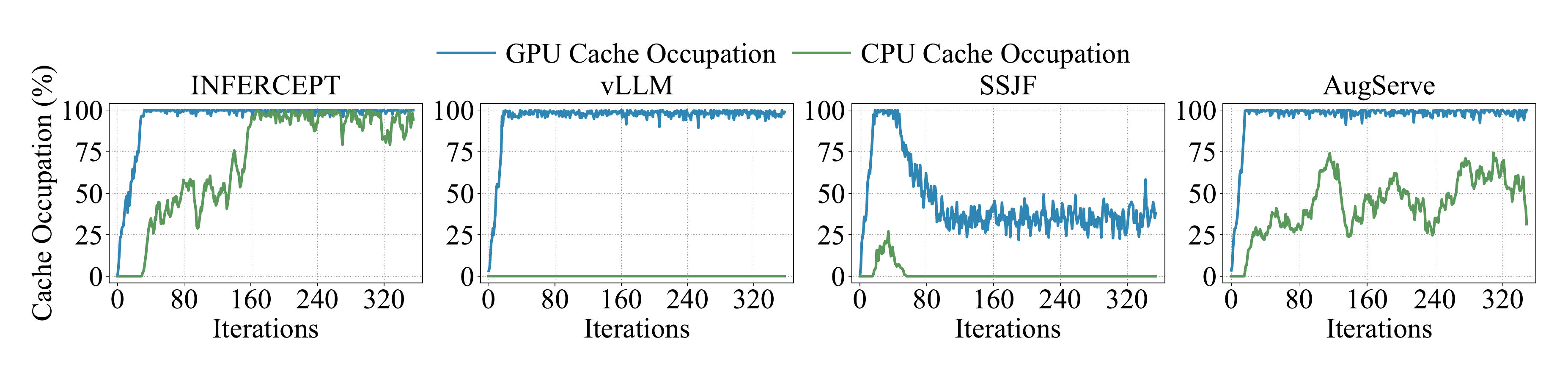}
        % \vspace{-22pt} % Add some space between image and subcaption
        \caption{GPU/CPU cache occupation comparison among vLLM, INFERCEPT, Speculative-SJF scheduling, and \name{} with OPT-13B on an H800 GPU using INFERCEPT dataset.} % Subcaption for the left group
        \label{fig:memory}
\end{figure*}

\subsection{SLO Attainment}
\label{app:slo_rate}
To better understand the source of goodput gains, we report SLO attainment under the same experimental settings as in \S\ref{sec:end-to-end}.
SLO attainment measures the fraction of requests that satisfy latency constraints and directly determines goodput.
As shown in \autoref{fig:slo_attainment}, 
\name{} consistently achieves higher SLO attainment than all baselines,
especially under higher load levels.
In contrast, FCFS-based systems quickly violate SLOs as queuing delays grow,
while SSJF provides only limited improvement due to its lack of awareness of external-call-induced execution states.
These results explain the goodput improvements observed in \S\ref{sec:end-to-end},
confirming that \name{} sustains higher goodput by maintaining high SLO attainment under contention.

\subsection{Memory Occupancy}
\label{app:memory}

\autoref{fig:memory} illustrates the GPU and CPU cache occupation across inference iterations. SSJF maintains low GPU utilization, with cache occupancy remaining below 50\% in most iterations and negligible CPU offloading.
This conservative behavior avoids memory pressure but leads to underutilized GPU resources and limited throughput.
vLLM rapidly saturates GPU memory without CPU offloading, indicating a lack of mechanisms to manage paused requests under external calls.
As a result, memory pressure accumulates on GPU, exacerbating queuing delays and HoL blocking. 
INFERCEPT actively offloads KV cache to CPU to relieve GPU pressure, but its CPU cache occupation increases steadily and remains high in later iterations,
reflecting frequent context swapping and significant offloading overhead.
Aggressive CPU offloading improves memory availability but introduces substantial resumption overhead, which is particularly harmful under frequent external-call-induced pauses.
\name{} achieves high GPU utilization while maintaining moderate and stable CPU cache occupation.
By explicitly accounting for reclaimable memory and dynamically adapting batch capacity,
\name{} controls memory pressure and avoids excessive context eviction and CPU offloading,
leading to more balanced resource utilization and stable performance.

\section{Approximation Guarantee}
\label{app:proof}
This section analyzes the value-density greedy policy under a simplified per-iteration scheduling formulation. 
Consider a fixed candidate set $U$, fixed scheduling values $v_i>0$, fixed memory costs $c_i>0$, and a fixed memory budget $B$. 
The per-iteration scheduling problem is:
\begin{equation}
\max \sum_{i\in U} v_i x_i
\quad
\mathrm{s.t.}
\quad
\sum_{i\in U} c_i x_i \le B,
\quad
x_i \in \{0,1\}.
\label{eq:simplified_knapsack}
\end{equation}
This is a standard 0--1 knapsack problem. 
% We discard infeasible singleton requests with $c_i>B$, since they cannot appear in any feasible solution. 
We discard infeasible singleton actions with $c_i>B$, since each candidate action is treated as indivisible in this simplified 0--1 formulation and thus cannot appear in any feasible solution.
Let $\rho_i=v_i/c_i$ denote the value density, and sort requests such that $\rho_1\ge\rho_2\ge\cdots\ge\rho_n$. 
Let $G$ be the feasible set obtained by density-based greedy packing, which scans requests in this order and includes a request whenever it fits. 
Let $S$ be the best feasible singleton:
\begin{equation}
S = \arg\max_{i: c_i \le B} v_i.
\end{equation}
% Let $S=\arg\max_{i:c_i\le B} v_i  $ be the best feasible singleton. 
The algorithm returns $\mathrm{ALG}=\max\{v(G),v(S)\}$.

\begin{theorem}
For the simplified per-iteration problem in \autoref{eq:simplified_knapsack}, the algorithm that returns the better of density-based greedy packing and the best feasible singleton achieves a $1/2$-approximation to the optimal 0--1 knapsack solution.
\end{theorem}

\begin{proof}
Let $\mathrm{OPT}$ be the optimal integral value.
Let $\mathrm{OPT}_{\mathrm{frac}}$ be the optimal value of the fractional relaxation, where each request can be partially selected, i.e., the binary decision $x_i \in \{0,1\}$ is relaxed to $x_i \in [0,1]$.
Clearly, $\mathrm{OPT}\le\mathrm{OPT}_{\mathrm{frac}}$. 
Since requests are sorted by non-increasing value density, the optimal fractional solution takes a density-ordered prefix and possibly a fraction of one additional request. 
That is, for some index $t$, requests $1,\ldots,t-1$ are fully selected and request $t$ is partially selected, with
\begin{equation}
\sum_{i=1}^{t-1} c_i \le B \le \sum_{i=1}^{t} c_i .
\end{equation}
Therefore,
\begin{equation}
\mathrm{OPT}_{\mathrm{frac}}
\le
\sum_{i=1}^{t-1} v_i + v_t .
\label{eq:frac_upper}
\end{equation}

By construction, the greedy packing solution $G$ includes all requests $1,\ldots,t-1$, because their total cost is at most $B$ and the greedy algorithm scans them before request $t$. 
Thus, $v(G)\ge\sum_{i=1}^{t-1}v_i$. 
Moreover, since request $t$ is feasible as a singleton after infeasible requests are removed, the best singleton satisfies $v(S)\ge v_t$. 
Combining these bounds with \autoref{eq:frac_upper}, we have
\begin{equation}
\mathrm{OPT}
\le
\mathrm{OPT}_{\mathrm{frac}}
\le
v(G)+v(S).
\end{equation}
Finally,
\begin{equation}
\mathrm{ALG}
=
\max\{v(G),v(S)\}
\ge
\frac{v(G)+v(S)}{2}
\ge
\frac{\mathrm{OPT}}{2}.
\end{equation}
Hence, the algorithm achieves a $1/2$-approximation.
\end{proof}

\paragraph{Discussion.}
This guarantee applies only to the simplified per-iteration subproblem with fixed values, fixed costs, and a fixed memory budget. 
It characterizes the greedy packing step in isolation and does not directly extend to the full augmented LLM scheduling problem, where costs evolve across service segments, memory usage depends on external call returns, context-handling policies introduce heterogeneous resumption costs, and scheduling decisions are continuously refined using runtime feedback. 
Extending formal approximation or competitive guarantees to the full online, multi-stage augmented LLM serving problem remains an important direction for future work.

\end{document}